\documentclass[journal=jacsat,manuscript=article]{achemso}

\usepackage[version=3]{mhchem} 
\usepackage{booktabs}
\usepackage{graphicx} 
\graphicspath{{Figures/}{../Figures/}} 
\usepackage{longtable}
\usepackage{multirow}
\usepackage{adjustbox}
\usepackage{tabularx}
\usepackage{caption}
\usepackage{subcaption}
\usepackage{overpic}
\usepackage{float}
\usepackage{xr}
\usepackage{hyperref}
\usepackage{cleveref}
\usepackage{lineno}
\usepackage{color}
\usepackage[normalem]{ulem}
\usepackage{cancel}
\usepackage{soul}

\makeatletter
\def\UrlAlphabet{%
      \do\a\do\b\do\c\do\d\do\e\do\f\do\g\do\h\do\i\do\j%
      \do\k\do\l\do\m\do\n\do\o\do\p\do\q\do\r\do\s\do\t%
      \do\u\do\v\do\w\do\x\do\y\do\z\do\A\do\B\do\C\do\D%
      \do\E\do\F\do\G\do\H\do\I\do\J\do\K\do\L\do\M\do\N%
      \do\O\do\P\do\Q\do\R\do\S\do\T\do\U\do\V\do\W\do\X%
      \do\Y\do\Z}
\def\UrlDigits{\do\1\do\2\do\3\do\4\do\5\do\6\do\7\do\8\do\9\do\0}
\g@addto@macro{\UrlBreaks}{\UrlOrds}
\g@addto@macro{\UrlBreaks}{\UrlAlphabet}
\g@addto@macro{\UrlBreaks}{\UrlDigits}
\makeatother

\makeatletter
\newcommand*{\addFileDependency}[1]{
  \typeout{(#1)}
  \@addtofilelist{#1}
  \IfFileExists{#1}{}{\typeout{No file #1.}}
}
\makeatother

\newcommand*{\myexternaldocument}[1]{%
    \externaldocument{#1}%
    \addFileDependency{#1.tex}%
    \addFileDependency{#1.aux}%
}

\myexternaldocument{SI}

\SectionNumbersOn

\author{Changwen Xu}
\affiliation{Department of Materials Science and Engineering, Carnegie Mellon University, Pittsburgh, PA 15213, USA}
\alsoaffiliation{Department of Mechanical Engineering, Carnegie Mellon University, Pittsburgh, PA 15213, USA}

\author{Yuyang Wang}
\affiliation{Department of Mechanical Engineering, Carnegie Mellon University, Pittsburgh, PA 15213, USA}
\alsoaffiliation{Machine Learning Department, Carnegie Mellon University, Pittsburgh, PA 15213, USA}

\author{Amir Barati Farimani}
\affiliation{Department of Materials Science and Engineering, Carnegie Mellon University, Pittsburgh, PA 15213, USA} 
\alsoaffiliation{Department of Mechanical Engineering, Carnegie Mellon University, Pittsburgh, PA 15213, USA}
\alsoaffiliation{Machine Learning Department, Carnegie Mellon University, Pittsburgh, PA 15213, USA}
\alsoaffiliation{Department of Chemical Engineering, Carnegie Mellon University, Pittsburgh, PA 15213, USA}
\email{barati@cmu.edu}

\title[An \textsf{achemso} demo]
  {TransPolymer: a Transformer-based language model for polymer property predictions}

\abbreviations{IR,NMR,UV}
\keywords{American Chemical Society, \LaTeX}

\begin{document}








\begin{abstract}
Accurate and efficient prediction of polymer properties is of great significance in polymer design. Conventionally, expensive and time-consuming experiments or simulations are required to evaluate polymer functions. Recently, Transformer models, equipped with self-attention mechanisms, have exhibited superior performance in natural language processing. However, such methods have not been investigated in polymer sciences. Herein, we report TransPolymer, a Transformer-based language model for polymer property prediction. Our proposed polymer tokenizer with chemical awareness enables learning representations from polymer sequences. Rigorous experiments on ten polymer property prediction benchmarks demonstrate the superior performance of TransPolymer. Moreover, we show that TransPolymer benefits from pretraining on large unlabeled dataset via Masked Language Modeling. Experimental results further manifest the important role of self-attention in modeling polymer sequences. We highlight this model as a promising computational tool for promoting rational polymer design and understanding structure-property relationships from a data science view.
\end{abstract}

\section{Introduction}

The accurate and efficient property prediction is essential to the design of polymers in various applications, including polymer electrolytes \cite{wang2020toward,xie2022accelerating}, organic optoelectronics \cite{st2019message, munshi2021transfer}, energy storage \cite{luo2018core, hu2020recent}, and many others \cite{bai2019accelerated, liang2021machine}. Rational representations which map polymers to continuous vector space are crucial to applying machine learning tools in polymer property prediction. Fingerprints (FPs), which have been proven to be effective in molecular machine learning models, are introduced for polymer-related tasks \cite{mannodi2018scoping}. Recently, deep neural networks (DNNs) have revolutionized polymer property prediction by directly learning expressive representations from data to generate deep fingerprints, instead of relying on manually engineered descriptors \cite{chen2021polymer}. Rahman et al. used convolutional neural networks (CNNs) for the prediction of mechanical properties of polymer-carbon nanotube surfaces \cite{rahman2021machine}, whereas CNNs suffered from failure to consider molecular structure and interactions between atoms. Graph neural networks (GNNs) \cite{scarselli2008graph}, which have outperformed many other models on several molecules and polymer benchmarks \cite{xie2018crystal, duvenaud2015convolutional, yang2019analyzing, karamad2020orbital, wang2022molecular}, are capable of learning representations from graphs and finding optimal fingerprints based on downstream tasks \cite{chen2021polymer}. For example, Park et al. \cite{park2022prediction} trained graph convolutional neural networks (GCNN) for predictions of thermal and mechanical properties of polymers and discovered that the GCNN representations for polymers resulted in comparable model performance to the popular extended-connectivity circular fingerprint (ECFP) \cite{cereto2015molecular, rogers2010extended} representation. Recently, Aldeghi et al. adapted a graph representation of molecular ensembles along with a GNN architecture to capture pivotal features and accomplish accurate predictions of electron affinity and ionization potential of conjugated polymers \cite{aldeghi2022graph}. However, GNN-based models require explicitly known structural and conformational information, which would be computationally or experimentally expensive to obtain. Plus, the degree of polymerization varies for each polymer, which makes it even harder to accurately represent polymers as graphs. Using the repeating unit only as graph is likely to result in missing structural information. Therefore, the optimal method of graph representation for polymers is still obscure.

Meanwhile, language models, like recurrent neural networks (RNNs) based models \cite{cho2014learning, schwaller2018found, tsai2020learning, flam2021language}, treat polymers as character sequences for featurization. Chemistry sequences have the same structure as a natural language like English, as suggested by Cadeddu et al., in terms of the distribution of text fragments and molecular fragments \cite{cadeddu2014organic}. This elucidates the development of sequence models similar to those in computational linguistics for extracting information from chemical sequences and realizing the intuition of understanding chemical texts just like understanding natural languages. Multiple works have investigated the development of deep language models for polymer science. Simine et al. managed to predict spectra of conjugated polymers by long short-term memory (LSTM) from coarse-grained representations of polymers \cite{simine2020predicting}. Webb et al. propose coarse-grained polymer genomes as sequences and apply LSTM to predict the properties of different polymer classes \cite{webb2020targeted}. Patel et al. further extend the coarse-grained string featurization to copolymer systems and develop GNN, CNN, as well as LSTM to model encoded copolymer sequences \cite{patel2022featurization}. Bhattacharya et al. leverage RNNs with sequence embedding to predict aggregate morphology of macromolecules \cite{bhattacharya2022predicting}. Plus, sequence models could represent molecules and polymers with Simplified Molecular-Input Line-Entry system (SMILES) \cite{weininger1988smiles} and convert the strings to embeddings for vectorization. Some works, like BigSMILES \cite{lin2019bigsmiles}, have also investigated the string-based encoding of macromolecules. Goswami et al. created encodings from polymer SMILES as input for the LSTM model for polymer glass transition temperature prediction \cite{goswami2021deep}. However, RNN-based models are generally not competitive enough to encode chemical knowledge from polymer sequences because they rely on previous hidden states for dependencies between words and tend to lose information when they reach deeper steps. In recent years, the exceptionally superior performance demonstrated by Transformer \cite{vaswani2017attention} on numerous natural language processing (NLP) tasks has shed light on studying chemistry and materials science by language models. Since proposed, Transformer and its variants have soon brought about significant changes in NLP tasks over the past few years. Transformer is featured with using attention mechanism only so that it can capture relationships between tokens in a sentence without relying on past hidden states. Many Transformer-based models like BERT \cite{devlin2018bert}, RoBERTa \cite{liu2019roberta}, GPT \cite{brown2020language}, ELMo \cite{peters2018dissecting}, and XLM \cite{conneau2019cross} have emerged as effective pretraining methods by self-supervised learning of representations from unlabeled texts, leading to performance enhancement on various downstream tasks. On this account, many works have already applied Transformer on property predictions of small organic molecules \cite{honda2019smiles, ying2021transformers, irwin2022chemformer, magar2022crystal}. SMILES-BERT was proposed to pretrain the model of BERT-like architecture through a masked SMILES recovery task and then generalize into different molecular property prediction tasks \cite{wang2019smiles}. Similarly, ChemBERTa \cite{chithrananda2020chemberta}, a RoBERTa-like model for molecular property prediction, was also introduced, following the pretrain-finetune pipeline. ChemBERTa demonstrated competitive performance on multiple downstream tasks and scaled well with the size of pretraining datasets. Transformer-based models could even be used for processing reactions. Schwaller et al. mimicked machine translation tasks and trained Transformer on reaction sequences represented by SMILES for reaction prediction with high accuracy \cite{schwaller2019molecular}. Recently, Transformer has been further proven to be effective as a structure-agnostic model in material science tasks, for example, predicting MOF properties based on a text string representation \cite{cao2023moformer}. Despite the wide investigation of Transformer for molecules and materials, such models have not yet been leveraged to learn representations of polymers. Compared with small molecules, designing Transformer-based models for polymers is more challenging because the standard SMILES encoding fails to model the polymer structure and misses fundamental factors influencing polymer properties like degree of polymerization and temperature of measurement. Moreover, the polymer sequences used as input should contain information on not only the definition of monomers but also the arrangement of monomers in polymers \cite{perry2020100th}. In addition, sequence models for polymers are confronted with an inherent scarcity of handy, well-labeled data, considering the hard work in the characterization process in the laboratory. The situation becomes even worse when some of the polymer data sources are not fully accessible \cite{le2012quantitative, persson2016silicon}.

\begin{figure}[!t]
    \centering
    \begin{overpic}[width=0.95\textwidth]{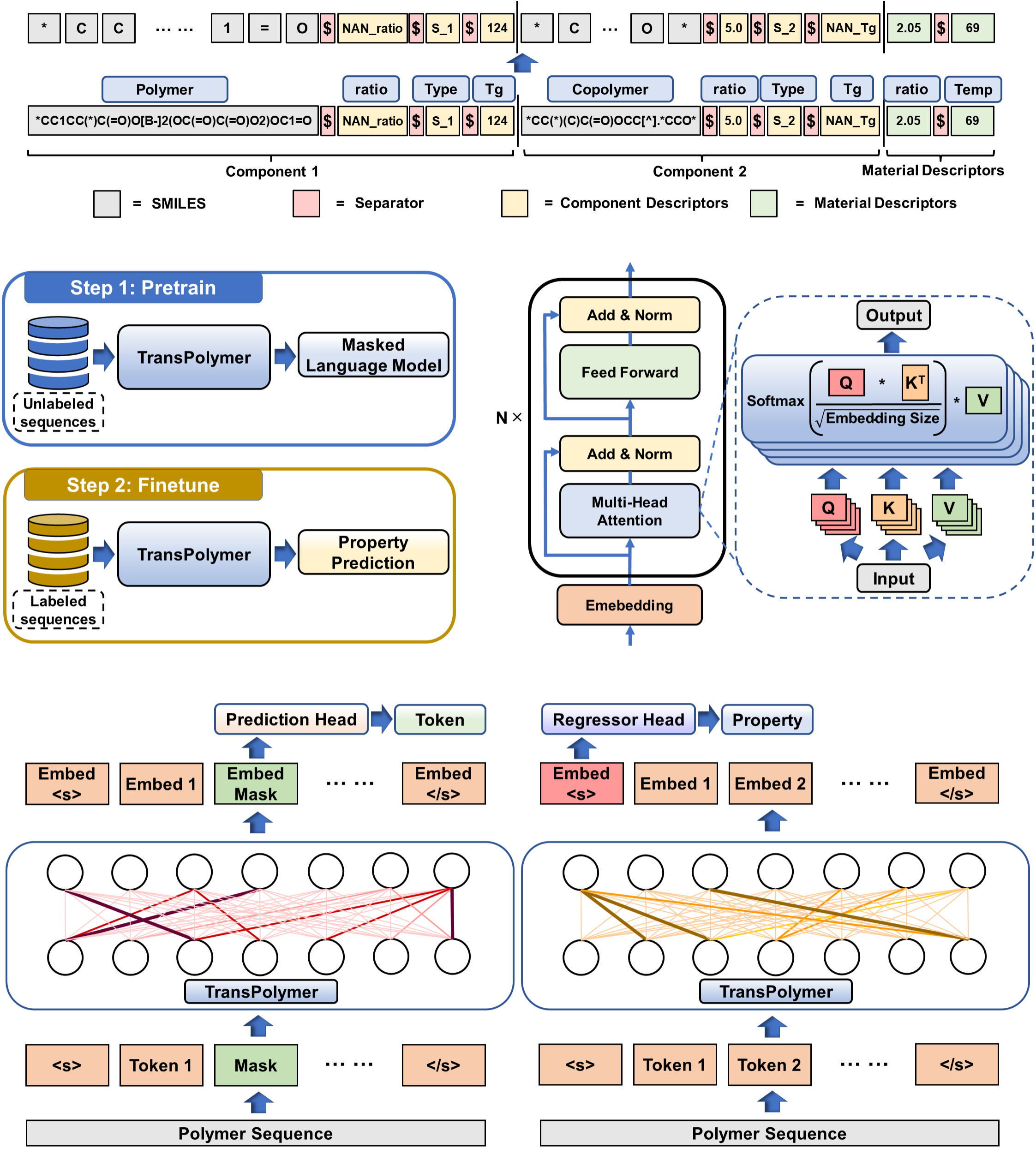}
        \put(-1, 100){\textbf{a}}
        \put(-1, 77){\textbf{b}}
        \put(42, 77){\textbf{c}}
        \put(-1, 39){\textbf{d}}
    \end{overpic}
    \caption{(a) Polymer tokenization. Illustrated by the example, the sequence which comprises components with polymer SMILES and other descriptors is tokenized with chemical awareness. (b) The whole TransPolymer framework with a pretrain-finetune pipeline.  (c) Sketch of Transformer encoder and multi-head attention. (d) Illustration of the pretraining (left) and finetuning (right) phases of TransPolymer. The model is pretrained with Masked Language Modeling to recover original tokens, while the feature vector corresponding to the special token `$\langle s \rangle$' of the last hidden layer is used for prediction when finetuning. Within the TransPolymer block, lines of deeper color and larger width stand for higher attention scores.}
    \label{F1}
\end{figure}
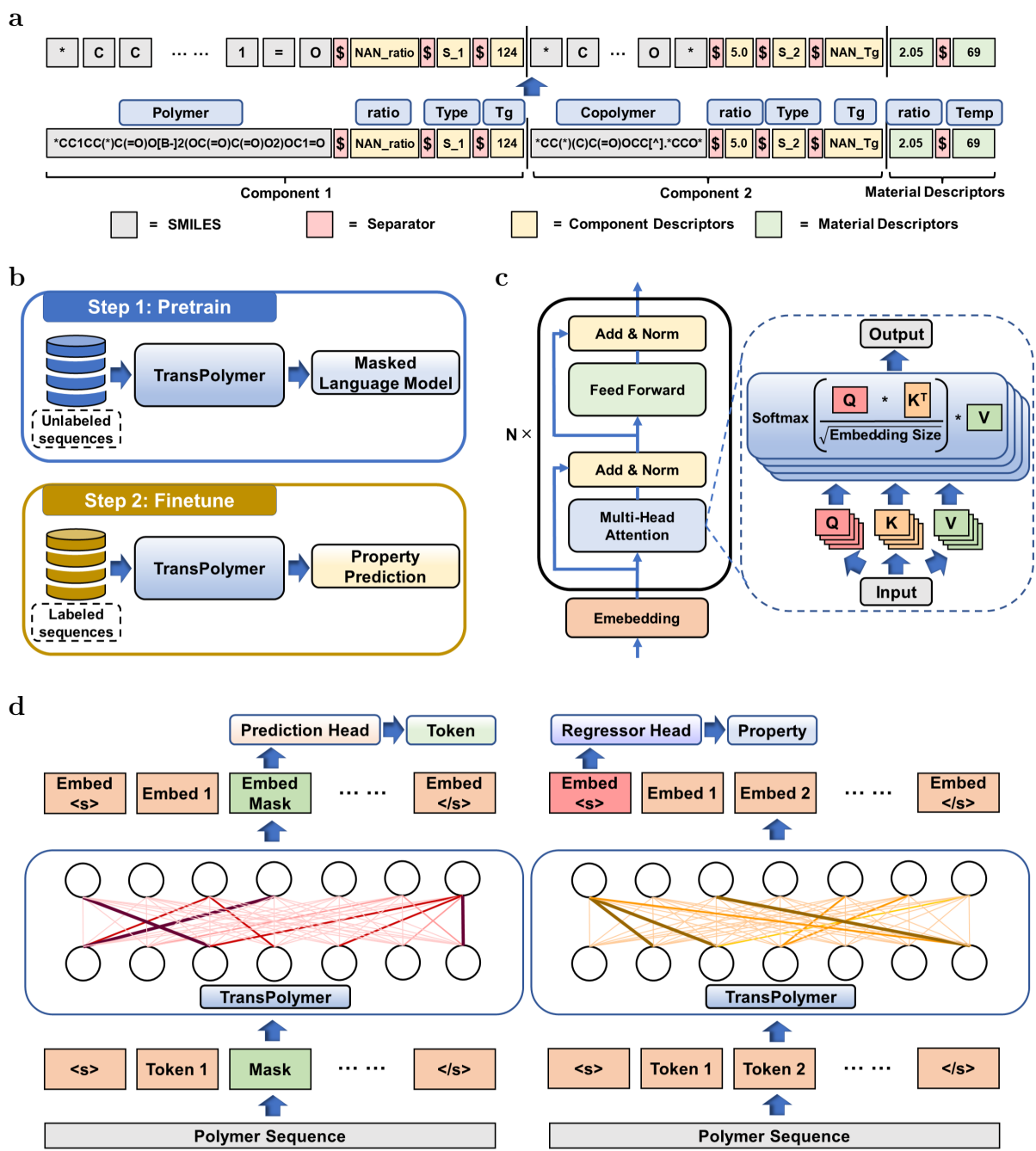

Herein, we propose TransPolymer, a Transformer-based language model for polymer property predictions. To the best of our knowledge, it is the first work to introduce the Transformer-based model to polymer sciences. Polymers are represented by sequences based on SMILES of their repeating units as well as structural descriptors and then tokenized by a chemically-aware tokenizer as the input of TransPolymer, shown in Fig. \ref{F1}(a). Even though there is still information which cannot be explicitly obtained from input sequences, like bond angles or overall polymer chain configuration, such information can still be learned implicitly by the model. TransPolymer consists of a RoBERTa architecture and a multi-layer perceptron (MLP) regressor head, for predictions of various polymer properties. In the pretraining phase, TransPolymer is trained through Masked Language Modeling (MLM) with approximately 5M augmented unlabeled polymers from the PI1M database \cite{ma2020pi1m}. In MLM, tokens in sequences are randomly masked and the objective is to recover the original tokens based on the contexts. Afterward, TransPolymer is finetuned and evaluated on ten datasets of polymers concerning various properties, covering polymer electrolyte conductivity, band gap, electron affinity, ionization energy, crystallization tendency, dielectric constant, refractive index, and p-type polymer OPV power conversion efficiency \cite{schauser2021database, hatakeyama2020ai, kuenneth2021polymer, nagasawa2018computer}. For each entry in the datasets, the corresponding polymer sequence, containing polymer SMILES as well as useful descriptors like temperature and special tokens are tokenized as input of TransPolymer. The pretraining and finetuning processes are illustrated in Fig. \ref{F1}(b) and (d). Data augmentation is also implemented for better learning of features from polymer sequences. TransPolymer achieves state-of-the-art (SOTA) results on all ten benchmarks and surpasses other baseline models by large margins in most cases. Ablation studies provide further evidence of what contributes to the superior performance of TransPolymer by investigating the roles of MLM pretraining on large unlabeled data, finetuning both Transformer encoders and the regressor head, and data augmentation. The evidence from visualization of attention scores illustrates that TransPolymer can encode chemical information about internal interactions of polymers and influential factors of polymer properties. Such a method learns generalizable features that can be transferred to property prediction of polymers, which is of great significance in polymer design.

\section{Results}

\subsection{TransPolymer framework}

Our TransPolymer framework consists of tokenization, Transformer encoder, pretraining, and finetuning. Each polymer data is first converted to a string of tokens through tokenization. Polymer sequences are more challenging to design than molecule or protein sequences as polymers contain complex hierarchical structures and compositions. For instance, two polymers that have the same repeating units can vary in terms of the degree of polymerization. Therefore, we propose a chemical-aware polymer tokenization method as shown in Fig. \ref{F1}(a). The repeating units of polymers are embedded using SMILES and additional descriptors (e.g., degree of polymerization, polydispersity, and chain conformation) are included to model the polymer system. Plus, copolymers are modeled by combining the SMILES of each constituting repeating unit along with the ratios and the arrangements of those repeating units. Moreover, materials consisting of mixtures of polymers are represented by concatenating the sequences for each component as well as the descriptors for the materials. Besides, each token represents either an element, the value of a polymer descriptor, or a special separator. Therefore, the tokenization strategy is chemical-aware and thus has an edge over the tokenizer trained for natural languages which tokenizes based on single letters. More details about the design of our chemical-aware tokenization strategy could be found in the Methods section. 

Transformer encoders are built upon stacked self-attention and point-wise, fully connected layers \cite{vaswani2017attention}, shown in Fig. \ref{F1}(c). Unlike RNN or CNN models, Transformer depends on the self-attention mechanism that relates tokens at different positions in a sequence to learn representations. Scaled dot-product attention across tokens is applied which relies on the query, key, and value matrices. More details about self-attention can be found in the Methods section. In our case, the Transformer encoder is made up of 6 hidden layers and each hidden layer contains 12 attention heads. The hyperparameters of TransPolymer are chosen by starting from the common setting of RoBERTa \cite{liu2019roberta} and then tuned according to model performance. 

To learn better representations from large unlabeled polymer data, the Transformer encoder is pretrained via Masked Language Modeling (MLM), a universal and effective pretraining method for various NLP tasks \cite{salazar2019masked, bao2020unilmv2, yang2020universal}. As shown in Fig. \ref{F1}(d) (left), 15\% of tokens of a sequence are randomly chosen for possible replacement, and the pretraining objective is to predict the original tokens by learning from the contexts. The pretrained model is then finetuned for predicting polymer properties with labeled data. Particularly, the final hidden vector of the special token `$\langle s \rangle$' at the beginning of the sequence is fed into a regressor head which is made up of one hidden layer with SiLU as the activation function for prediction as illustrated in Fig. \ref{F1}(d) (right).

\subsection{Experimental settings}

PI1M, the benchmark of polymer informatics, is used for pretraining. The benchmark, whose size is around 1M, was built by Ma et al. by training a generative model on polymer data collected from the PolyInfo database \cite{ma2020pi1m, otsuka2011polyinfo}. The generated sequences consist of monomer SMILES and `*' signs representing the polymerization points. The $\sim$1M database was demonstrated to cover similar chemical space as PolyInfo but populate space where data in PolyInfo are sparse. Therefore, the database can serve as an important benchmark for multiple tasks in polymer informatics. 

To finetune the pretrained TransPolymer, ten datasets are used in our experiments which cover various properties of different polymer materials, and the distributions of polymer sequence lengths vary from each other (shown in Supplementary Figure 1). Plus, data in all the datasets are of different types: sequences from Egc, Egb, Eea, Ei, Xc, EPS, and Nc datasets are about polymers only so that the inputs are just polymer SMILES; while PE-\uppercase\expandafter{\romannumeral1}, PE-\uppercase\expandafter{\romannumeral2}, and OPV datasets describe polymer-based materials so that the sequences contain additional descriptors. In particular, PE-\uppercase\expandafter{\romannumeral1} which is about polymer electrolytes involves mixtures of multiple components in polymer materials. Hence, these datasets provide challenging and comprehensive benchmarks to evaluate the performance of TransPolymer. A summary of the ten datasets for downstream tasks is shown in Table \ref{downstream datasets}. 

\begin{table}[t]
    \caption{Summary of datasets for downstream tasks.}
    \label{downstream datasets}
    \centering
    \begin{adjustbox}{width=\textwidth}
        \begin{tabular}{cccccc}
        \toprule
         Dataset & Property & \# Data & \# Augmented train data & \# Test data & Data split \\
         \midrule
         PE-\uppercase\expandafter{\romannumeral1} \cite{hatakeyama2020ai} & conductivity & 9185 & 34803 & 146 & train-test split by year \\ 
         PE-\uppercase\expandafter{\romannumeral2} \cite{schauser2021database} & conductivity & 271 & 8864 & 55 & 5-fold cross-validation \\ 
         Egc \cite{kuenneth2021polymer} & bandgap (chain) & 3380 & 5408 & 676 & 5-fold cross-validation\\
         Egb \cite{kuenneth2021polymer} & bandgap (bulk) & 561 & 6443 & 113 & 5-fold cross-validation \\
         Eea \cite{kuenneth2021polymer} & electron affinity & 368 & 3993 & 74 & 5-fold cross-validation \\
         Ei \cite{kuenneth2021polymer} & ionization energy & 370 & 4000 & 74 &5-fold cross-validation \\
         Xc \cite{kuenneth2021polymer} & crystallization tendency & 432 & 8837 & 87 & 5-fold cross-validation \\
         EPS \cite{kuenneth2021polymer} & dielectric constant & 382 & 4188 & 77 &5-fold cross-validation \\
         
         Nc \cite{kuenneth2021polymer} & refractive index & 382 & 4188 & 77 & 5-fold cross-validation\\
         OPV \cite{nagasawa2018computer} & power conversion efficiency & 1203 & 4810 & 241 & 5-fold cross-validation \\
        \bottomrule 
    		 
        \end{tabular}
    \end{adjustbox}
    
\end{table}

We apply data augmentation to each dataset that we use by removing canonicalization from SMILES and generating non-canonical SMILES which correspond to the same structure as the canonical ones. For PI1M database, each data entry is augmented to five so that the augmented dataset with the size of $\sim$5M is used for pretraining. For downstream datasets, we limit the numbers of augmented SMILES for large datasets with long SMILES for the following reasons: long SMILES tend to generate more non-canonical SMILES which might alter the original data distribution; we are not able to use all the augmented data for finetuning given the limited computation resources. We include the number of data points after augmentation in Table 1 and summarize the augmentation strategy for each downstream dataset in Supplementary Table 1.

\subsection{Polymer property prediction results}

The performance of our pretrained TransPolymer model on ten property prediction tasks is illustrated below. We use root mean square error (RMSE) and $R^2$ as metrics for evaluation. For each benchmark, the baseline models and data splitting are adopted from the original literature. Except for PE-\uppercase\expandafter{\romannumeral1} which is trained on data from the year 2018 and evaluated on data from the year 2019, all other datasets are split by five-fold cross-validation. When cross-validation is used, the metrics are calculated by taking the average of those by each fold. We also train Random Forest models using Extended Connectivity Fingerprint (ECFP) \cite{cereto2015molecular, rogers2010extended}, one of the state-of-the-art fingerprint approaches, to compare with TransPolymer. Besides, we develop long short-term memory (LSTM), another widely used language model, as well as unpretrained TransPolymer trained purely via supervised learning as baseline models in all the benchmarks. TransPolymer\textsubscript{unpretrained} and TransPolymer\textsubscript{pretrained} denote unpretrained and pretrained TransPolymer, respectively. 

\begin{figure*}[htbp]
    \centering
    \makebox[\textwidth][c]{
        \begin{overpic}[scale=0.125]{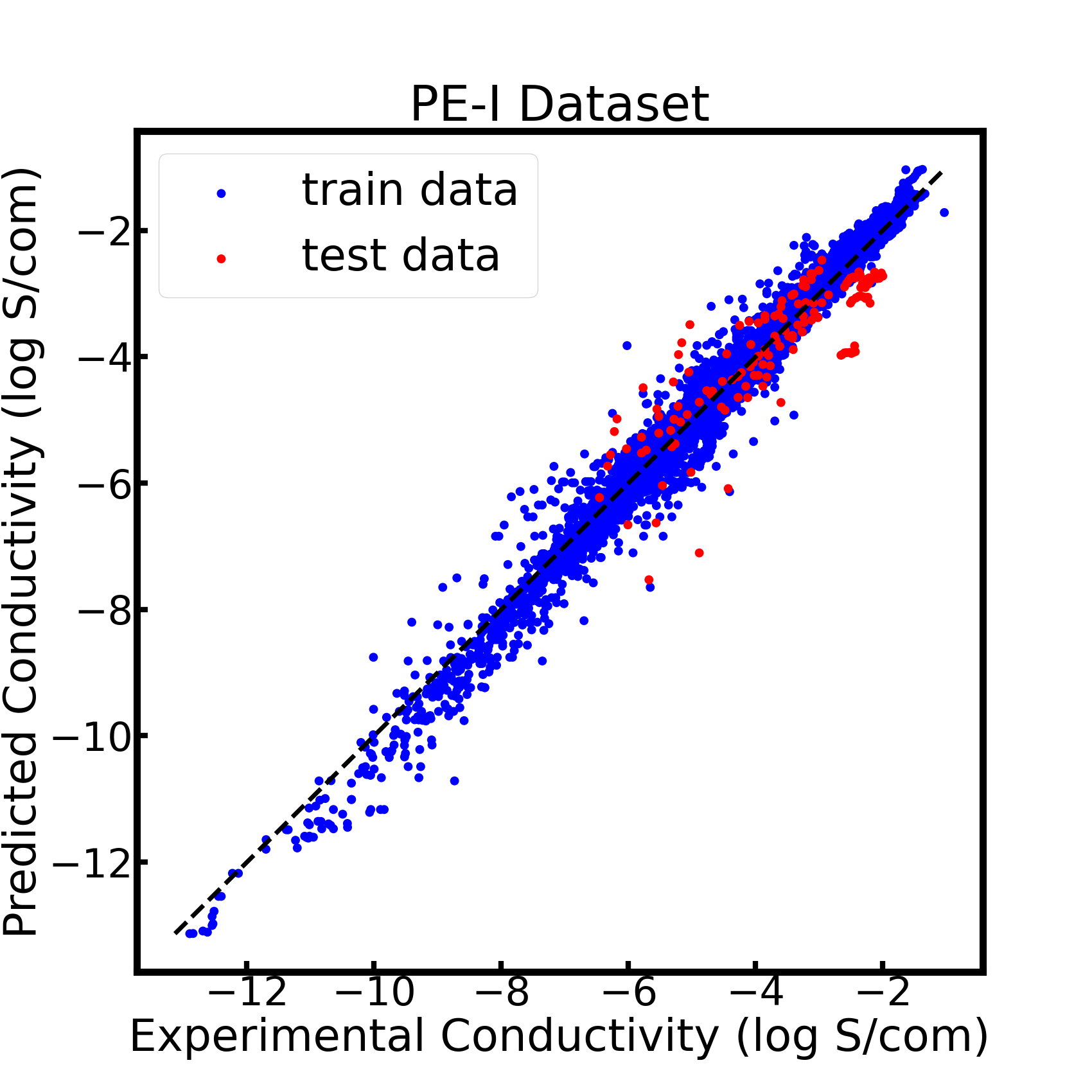}
        \put(-3,95){\textbf{a}}
    \end{overpic}
    \begin{overpic}[scale=0.125]{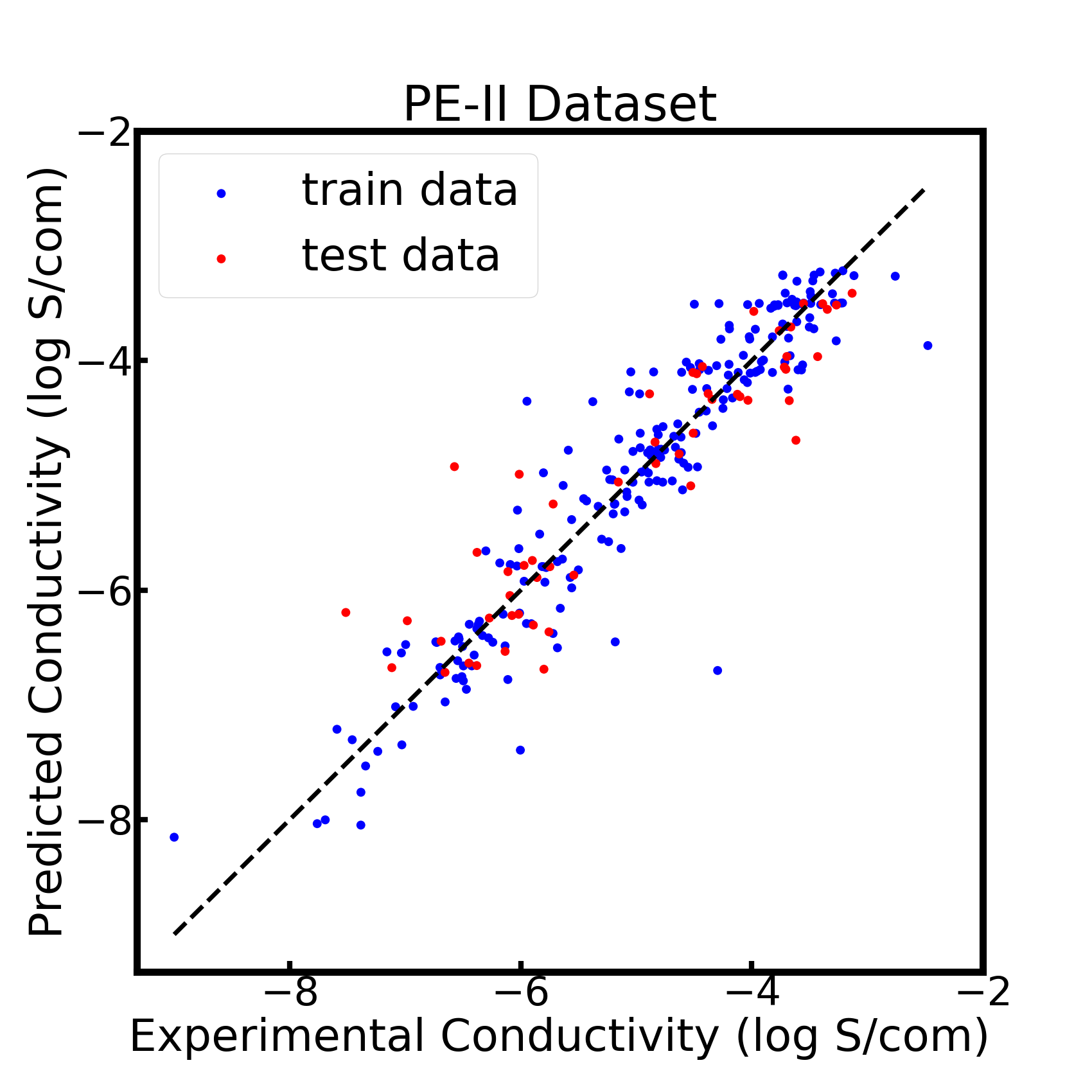}
        \put(-3,95){\textbf{b}}
    \end{overpic}
    \begin{overpic}[scale=0.125]{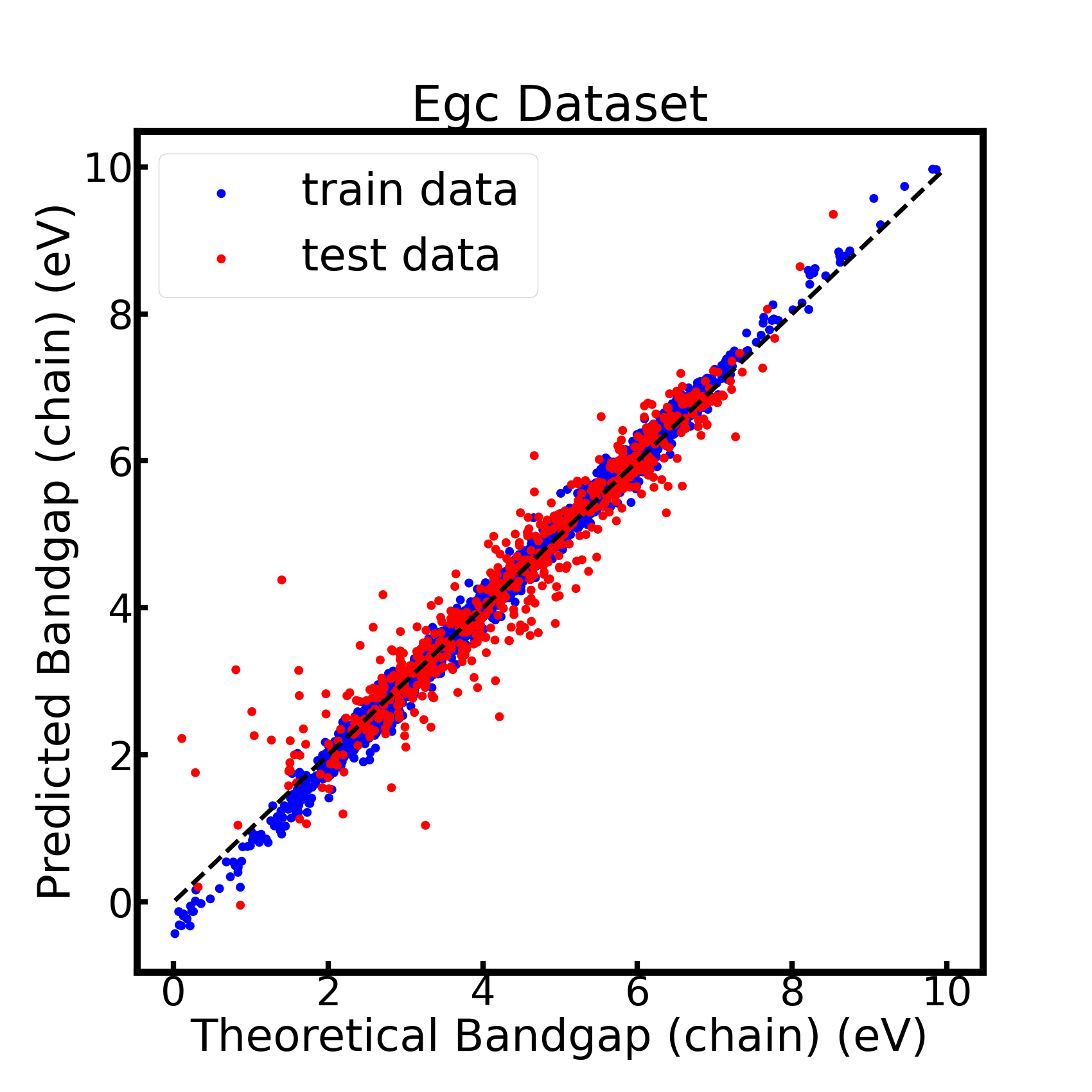}
        \put(-3,95){\textbf{c}}
    \end{overpic}
    }
    \makebox[\textwidth][c]{
    \begin{overpic}[scale=0.125]{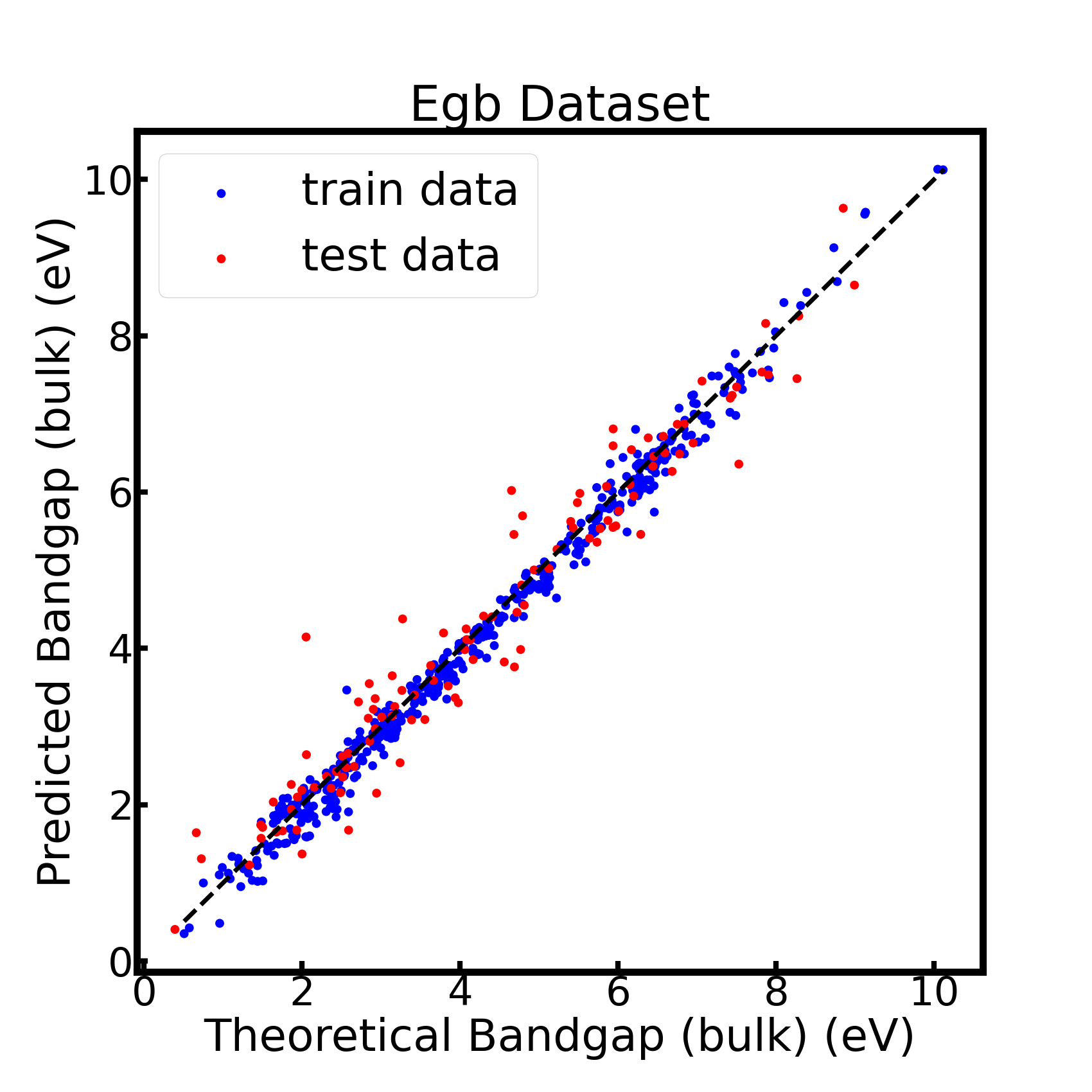}
        \put(-3,95){\textbf{d}}
    \end{overpic}
    \begin{overpic}[scale=0.125]{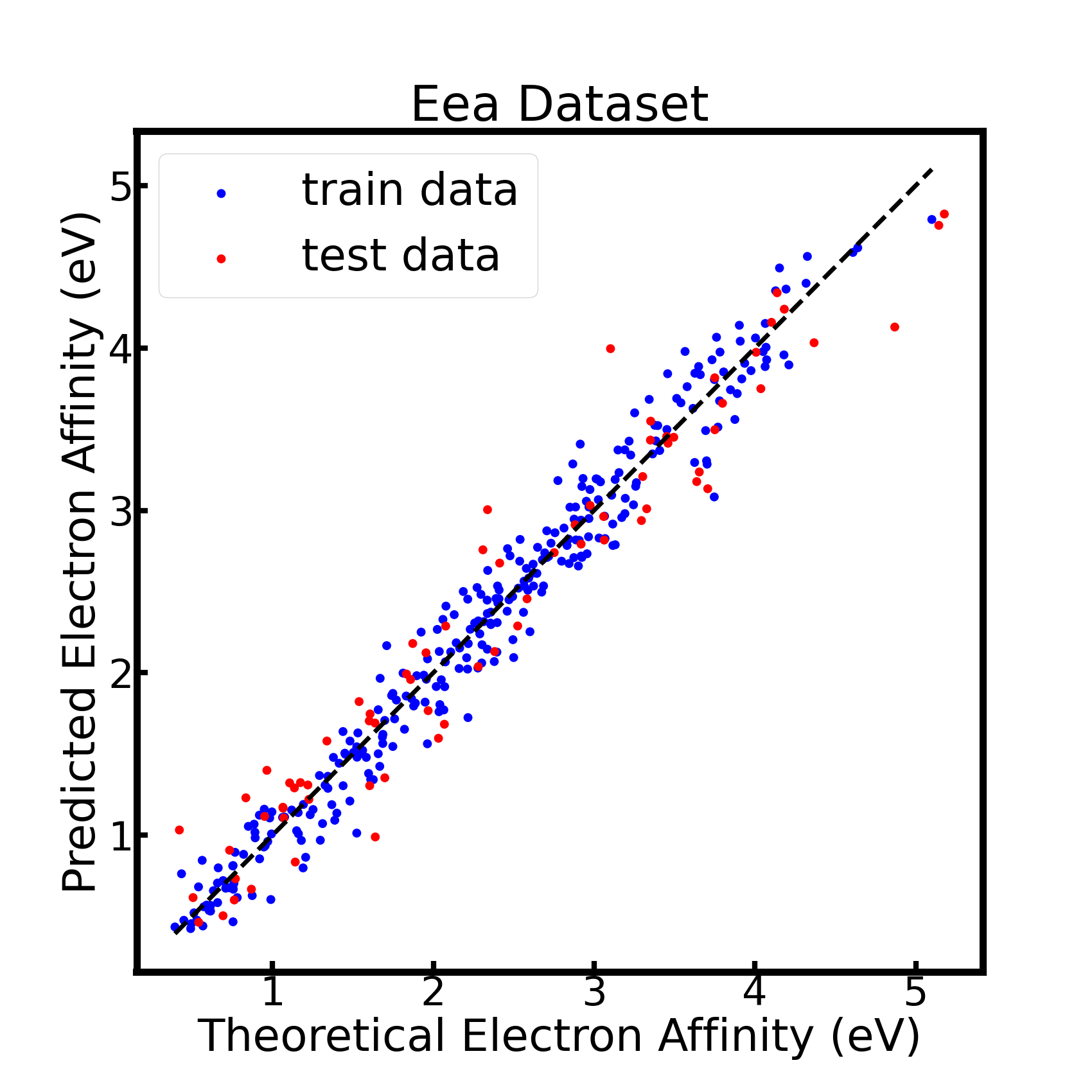}
        \put(-3,95){\textbf{e}}
    \end{overpic}
    \begin{overpic}[scale=0.125]{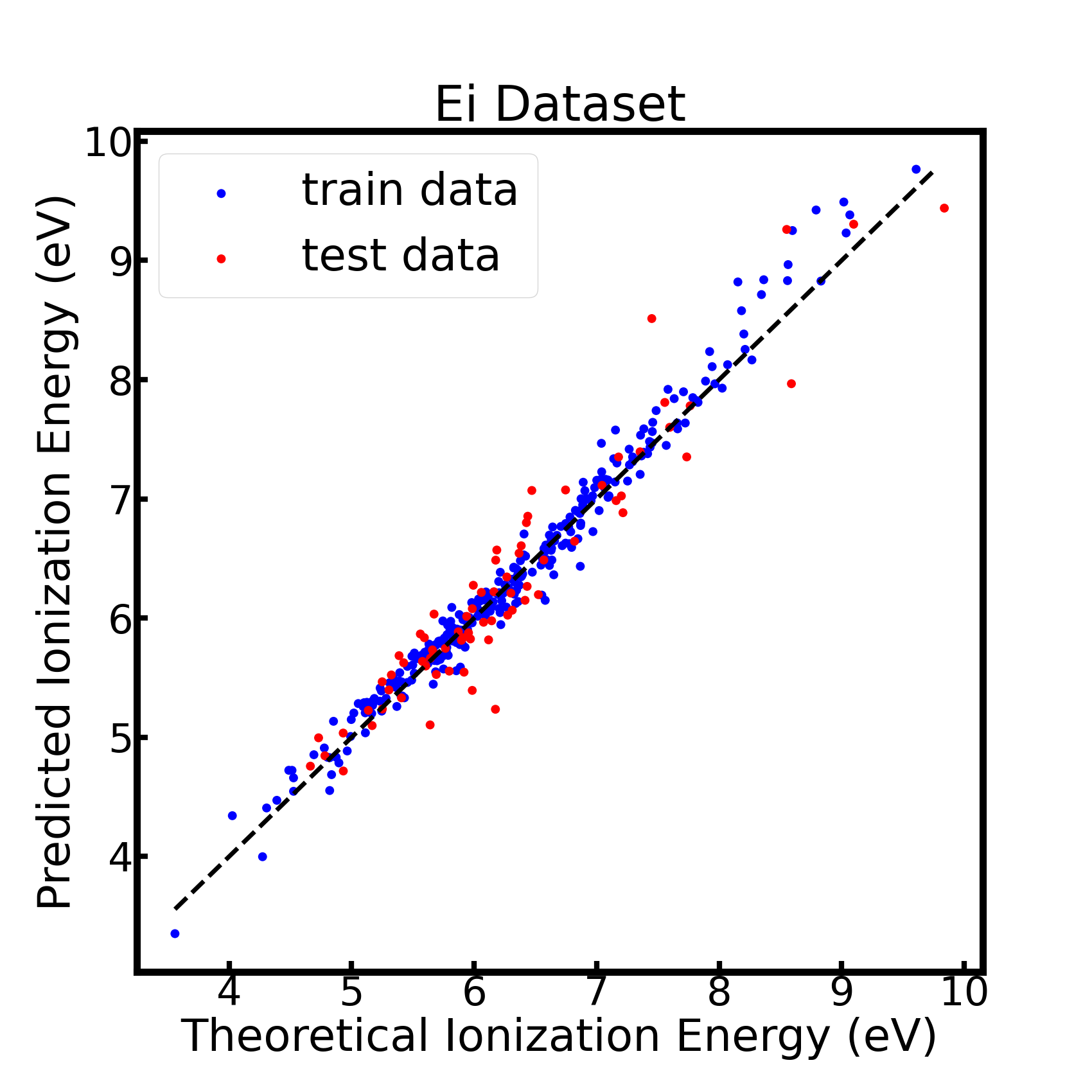}
        \put(-3,95){\textbf{f}}
    \end{overpic}
    }
    \makebox[\textwidth][c]{
    \begin{overpic}[scale=0.125]{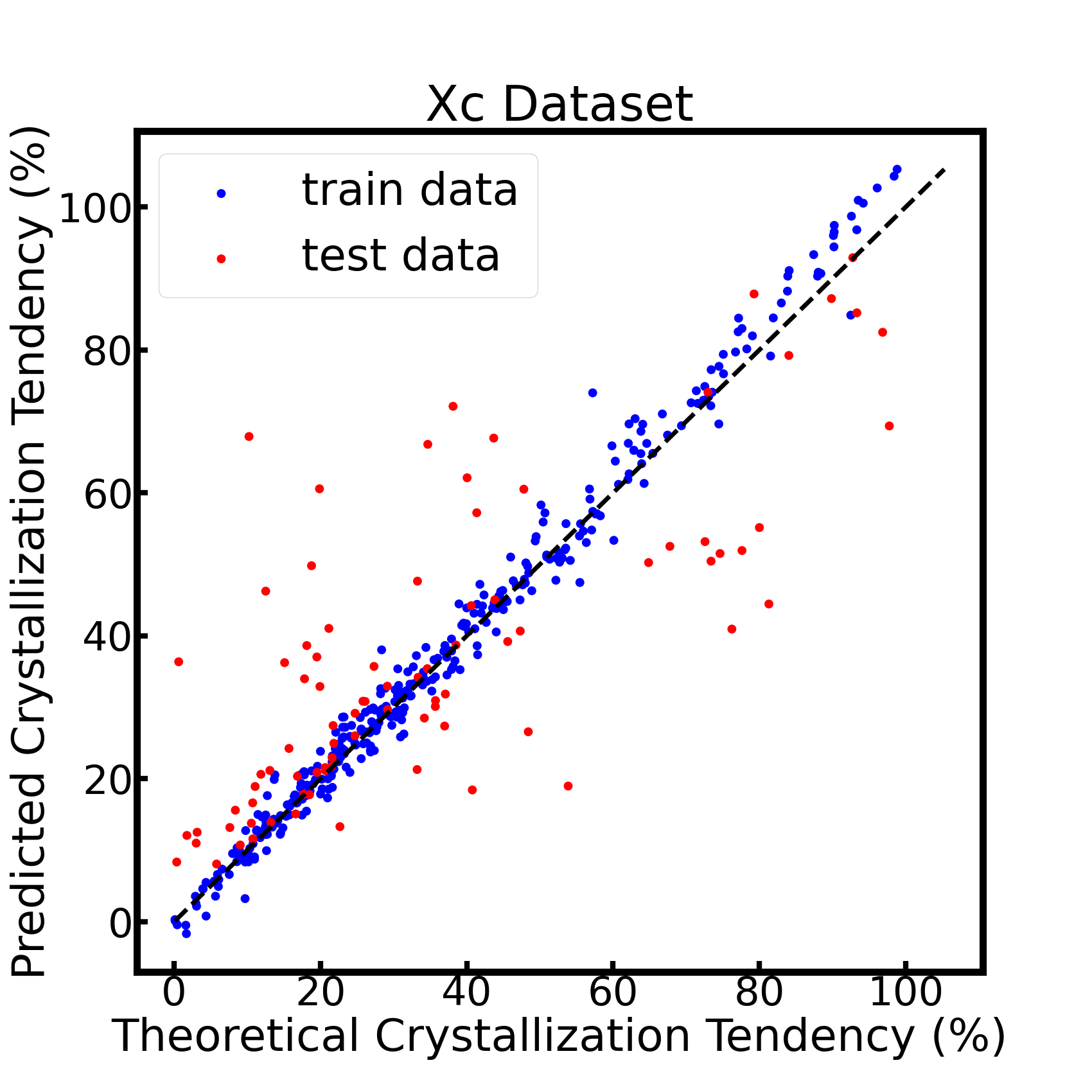}
        \put(-3,95){\textbf{g}}
    \end{overpic}
    \begin{overpic}[scale=0.125]{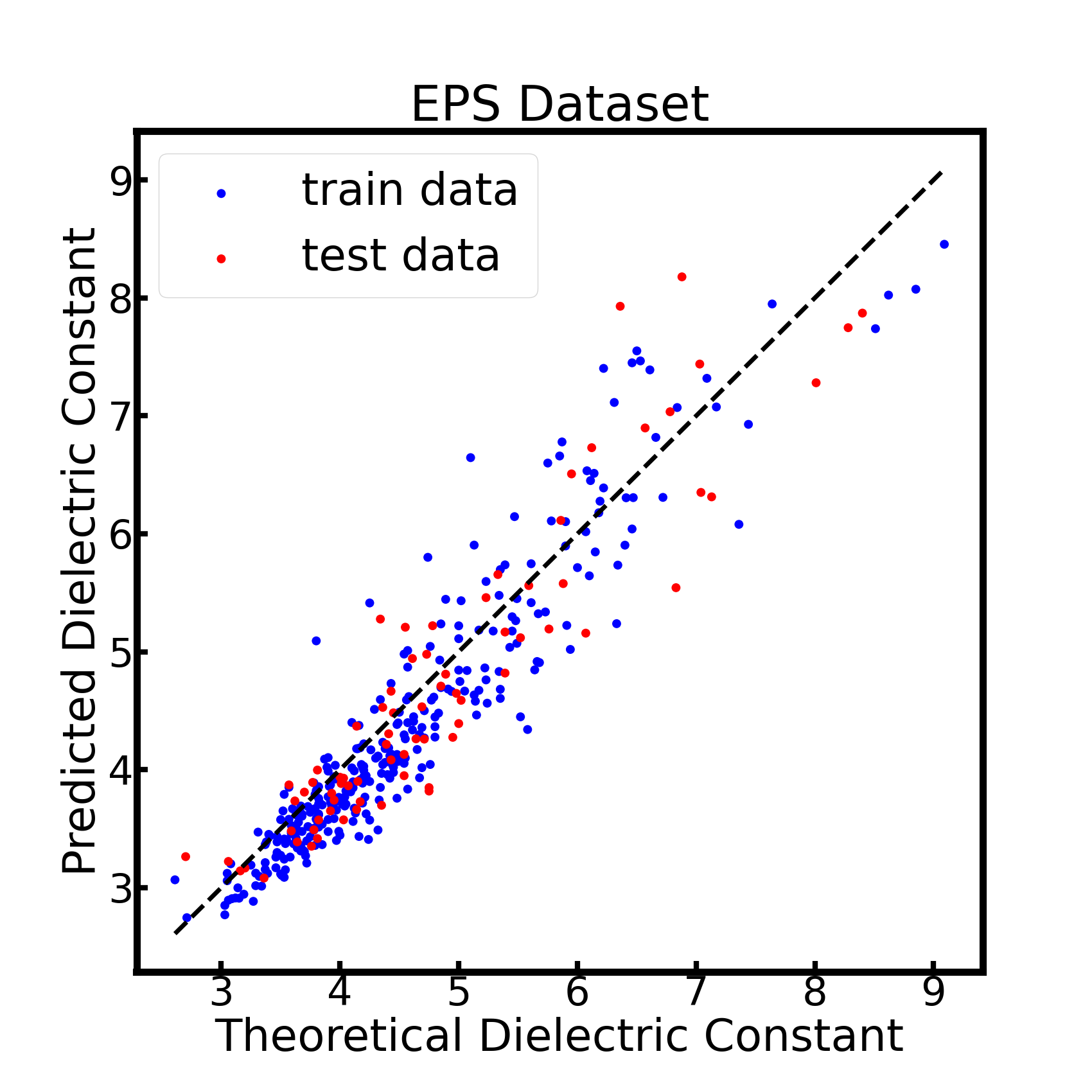}
        \put(-3,95){\textbf{h}}
    \end{overpic}
    \begin{overpic}[scale=0.125]{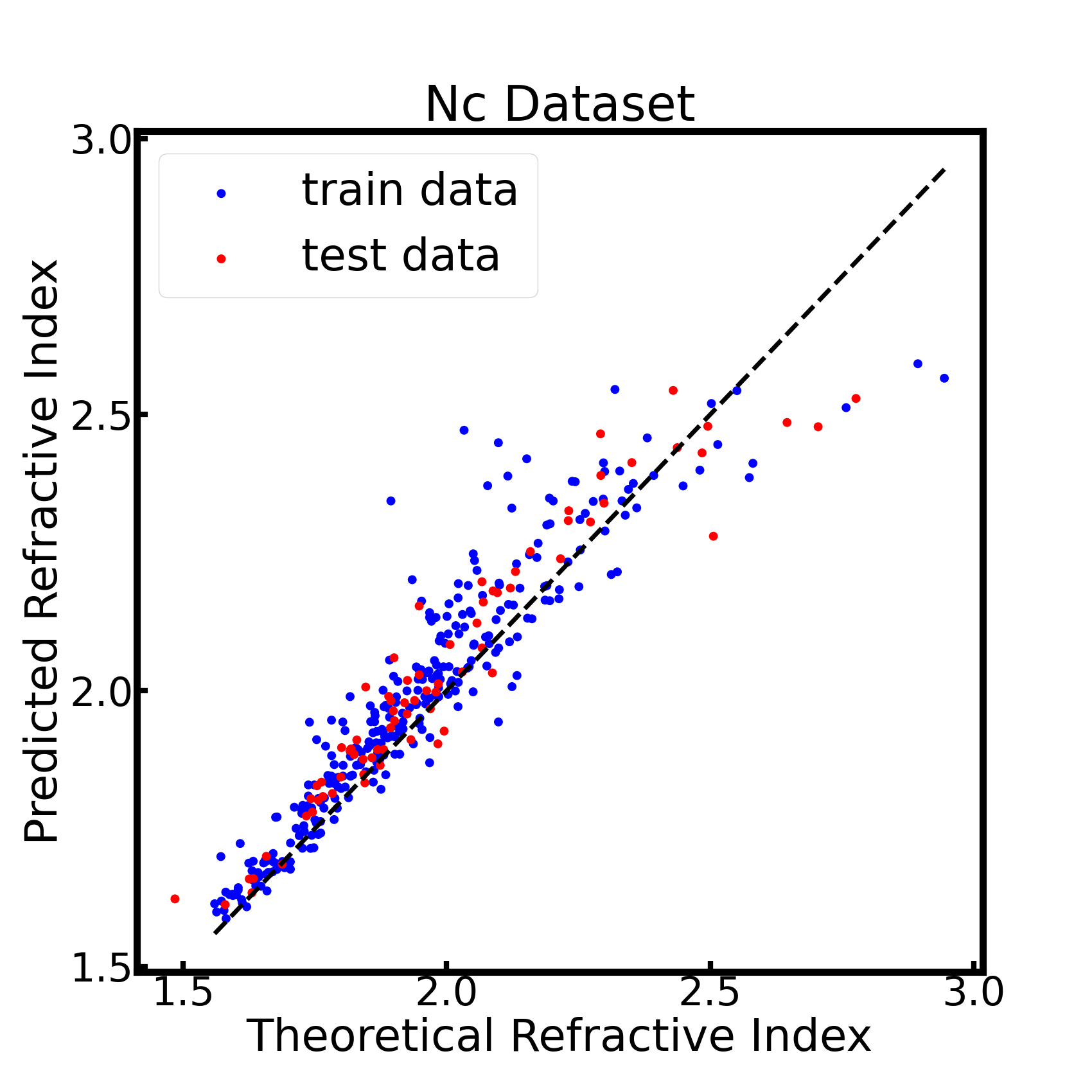}
        \put(-3,95){\textbf{i}}
    \end{overpic}
    
    }
    \makebox[\textwidth][c]{
    \begin{overpic}[scale=0.125]{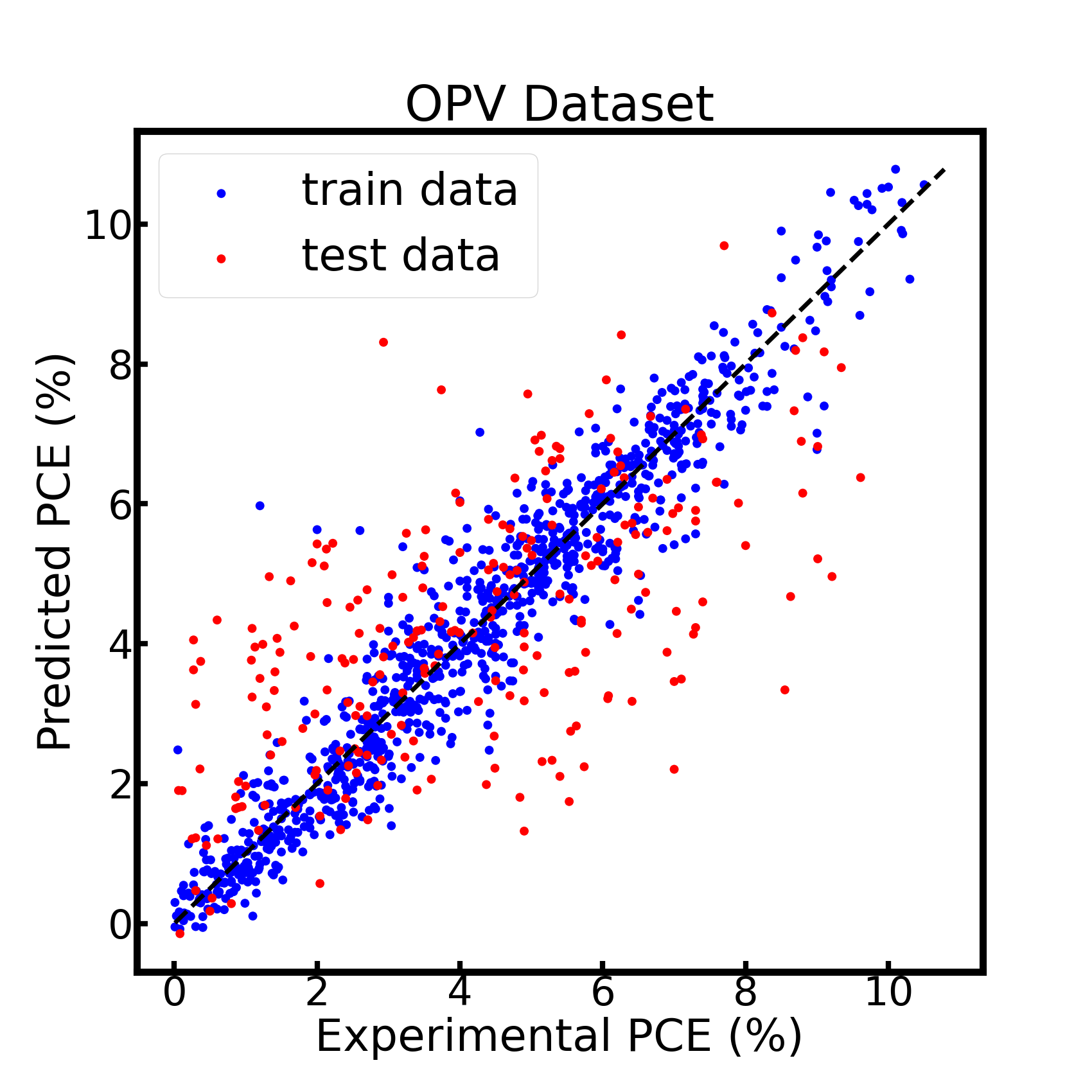}
        \put(-3,95){\textbf{j}}
    \end{overpic}
    }
    \caption{Scatter plots of ground truth vs. predicted values by TransPolymer\textsubscript{pretrained} for downstream datasets: (a) PE-\uppercase\expandafter{\romannumeral1}, (b) PE-\uppercase\expandafter{\romannumeral2}, (c) Egc, (d) Egb, (e) Eea, (f) Ei, (g) Xc, (h) EPS, (i) Nc, and (j) OPV. The dashed lines on diagonals stand for perfect regression.}
    \label{Predict_vs_true}
    
\end{figure*}

The results of TransPolymer and baselines on PE-\uppercase\expandafter{\romannumeral1} are illustrated in Table \ref{T1}. The original literature used gated GNN to generate fingerprints for the prediction of polymer electrolyte conductivity by Gaussian process \cite{hatakeyama2020ai}. The fingerprints are also passed to random forest and supporting vector machine (SVM) for comparison. Another random forest is trained based on ECFP fingerprints. The results of most baseline models indicate strong overfitting which is attributed to the introduction of unconventional conductors consisting of conjugated polybenzimidazole and ionic liquid. For instance, Gaussian Process trained on GNN fingerprints achieves a $R^2$ of 0.90 on the training set but only 0.16 on the test set, and GNN + Random Forest gets a negative test $R^2$ even the train $R^2$ is 0.91. Random Forest trained on ECFP fingerprints stands out among all the baseline models, whereas its performance on test dataset is still poor. However, TransPolymer\textsubscript{pretrained} not only achieves the highest scores on the training set but also improves the performance on the test set significantly, which is illustrated by the $R^2$ of 0.69 on the test set. Such information demonstrates that TransPolymer is capable of learning the intrinsic relationship between polymers and their properties and suffers less from overfitting. Notably, TransPolymer\textsubscript{unpretrained} also achieves competitive results and shows mild overfitting compared with other baseline models. This indicates the effectiveness of the attention mechanism of Transformer-based models. The scatter plots of ground truth vs. predicted values for PE-\uppercase\expandafter{\romannumeral1} by TransPolymer\textsubscript{pretrained} are illustrated in Fig. \ref{Predict_vs_true}(a) and Supplementary Figure 2(a).

\begin{table}[t]
    \caption{Performance of TransPolymer and baseline models on PE-\uppercase\expandafter{\romannumeral1}.}
    \label{T1}
    \centering
    \begin{adjustbox}{width=\textwidth}
        \begin{tabular}{ccccc}
        \toprule
         Model & Train RMSE (S/cm$^*$) ($\downarrow$) & Test RMSE (S/cm$^*$) ($\downarrow$) & Train $R^2$ ($\uparrow$) & Test $R^2$ ($\uparrow$)\\ \midrule
         Gaussian Process (GNN FP) & 0.55 & 0.97 & 0.90 & 0.16 \\ 
         Random Forest (GNN FP) & 0.50 & 2.23 & 0.91 & -2.64 \\ 
         SVM (GNN FP) & 1.34 & 2.12 & 0.04 & -1.94 \\
         Random Forest (ECFP) & 0.15 & 1.00 & 0.99 & 0.32\\
         LSTM & 1.03 & 1.36 & 0.67 & -0.25  \\           $\text{TransPolymer}_{\text{unpretrained}}$ & 0.88 & 1.02 & 0.70 & 0.30 \\
         $\text{TransPolymer}_{\text{pretrained}}$ & \textbf{0.20} & \textbf{0.67} & \textbf{0.98} & \textbf{0.69}  \\

        \bottomrule
        * The units are in logarithm scale
    		 
        \end{tabular}
    \end{adjustbox}
    
\end{table}

As is shown in Table \ref{T5}, the results of TransPolymer and baselines including Ridge, random forest, gradient boosting, and extra trees which were trained on chemical descriptors generated from polymers from PE-\uppercase\expandafter{\romannumeral2} in the original paper \cite{schauser2021database} are listed, as well as random forest trained on ECFP fingerprints. Although gradient boosting surpasses other models on training sets by obtaining nearly perfect regression outcomes, its performance on test sets drops significantly. In contrast, TransPolymer\textsubscript{pretrained}, which achieves the lowest RMSE of 0.61 and highest $R^2$ of 0.73 on the average of cross-validation sets, exhibits better generalization. The scatter plots of ground truth vs. predicted values for PE-\uppercase\expandafter{\romannumeral2} by TransPolymer\textsubscript{pretrained} are illustrated in Fig. \ref{Predict_vs_true}(b) and Supplementary Figure 2(b).

\begin{table}[t]
    \caption{Performance of TransPolymer and baseline models on PE-\uppercase\expandafter{\romannumeral2}.}
    \label{T5}
    \centering
    \begin{adjustbox}{width=\textwidth}
    \begin{tabular}{ccccc}
        \toprule
         Model & Train RMSE (S/cm$^*$) ($\downarrow$) & Test RMSE (S/cm$^*$) ($\downarrow$) & Train $R^2$ ($\uparrow$) & Test $R^2$ ($\uparrow$)\\ \midrule
         Ridge (Chemical descriptors) & 0.58 & 0.67 & 0.77 & 0.58 \\
         Random Forest (Chemical descriptors) & 0.26 & 0.64 & 0.96 & 0.71 \\ 
         Gradient Boosting (Chemical descriptors) & \textbf{0.00}& 0.66 & \textbf{0.99} & 0.68 \\ 
         Extra Trees (Chemical descriptors) & 0.10 & 0.63 & 0.98 & 0.72 \\
         Random Forest (ECFP) & 0.22 & 0.94 & 0.96 & 0.27\\
         LSTM & 1.16 & 1.18 & 0.05 & 0.00 \\
         $\text{TransPolymer}_{\text{unpretrained}}$ & 0.18 & 0.80 & 0.97 & 0.54  \\
         $\text{TransPolymer}_{\text{pretrained}}$ & 0.18 & \textbf{0.61} & 0.96 & \textbf{0.73} \\

        \bottomrule
        * The units are in logarithm scale
    		 
    \end{tabular}
    \end{adjustbox}
\end{table}

Table \ref{T4} summarizes the performance of TransPolymer and baselines on Egc, Egb, Eea, Ei, Xc, EPS, and Nc datasets from Kuenneth et al \cite{kuenneth2021polymer}. In the original literature, both Gaussian process and neural networks were trained on each dataset with polymer genome (PG) fingerprints \cite{kim2018polymer} as input, some of which resulted in desirable performance while some of which did not. Meanwhile, PG fingerprints are demonstrated to surpass ECFP fingerprints on the datasets used by Kuenneth et al. For Egc, Egb, and Eea, despite the high scores by other models, TransPolymer\textsubscript{pretrained} is still able to enhance the performance, lowering RMSE and enhancing $R^2$. In contrast, baseline models perform poorly on Xc whose test $R^2$ scores are less than 0. However, TransPolymer\textsubscript{pretrained} significantly lowers test RMSE and increases $R^2$ to 0.50. Notably, The authors of the original paper used multi-task learning to enhance model performance and achieved higher scores than TransPolymer\textsubscript{pretrained} on some of the datasets, like Egb, EPS, and Nc (the average test RMSE and $R^2$ are 0.43 and 0.95 for Egb, 0.39 and 0.86 for EPS, and 0.07 and 0.91 for Nc, respectively). Access to multiple properties of one polymer, however, may not be available from time to time, which limits the application of multi-task learning. In addition, the TransPolymer\textsubscript{pretrained} still outperforms multi-task learning models on four out of the seven chosen datasets. Hence the improvement by TransPolymer compared with single-task baselines should still be highly valued. The scatter plots of ground truth vs. predicted values for Egc, Egb, Eea, Ei, Xc, EPS, and Nc datasets by TransPolymer\textsubscript{pretrained} are depicted in Fig. \ref{Predict_vs_true}(c)-(i) and Supplementary Figure 2(c)-(i), respectively.

\begin{table}[t]
    \caption{Performance of TransPolymer and baseline models on datasets from literature by Kuenneth et al.\cite{kuenneth2021polymer}}
    \label{T4}
    \centering
    \begin{adjustbox}{width=\textwidth}
    \begin{tabular}{cccccccc|ccccccc}
        \toprule
          \multirow{2}*{Model} & \multicolumn{7}{c}{Test RMSE ($\downarrow$)} & \multicolumn{7}{c}{Test $R^2$ ($\uparrow$)} \\
          ~ & Egc (eV) & Egb (eV) & Eea (eV) & Ei (eV) & Xc (\%) & EPS & Nc & Egc & Egb & Eea & Ei & Xc & EPS & Nc \\ \midrule
          Gaussian Process (PG) & 0.48 & 0.55 & \textbf{0.32} & 0.42 & 24.42 & 0.53 & \textbf{0.10} & 0.90 & 0.91 & 0.90 & 0.77 & \textless 0 & 0.68 & 0.79 \\
          Neural Network (PG) & 0.49 & 0.57 & \textbf{0.32} & 0.45 & 20.74 & 0.54 & \textbf{0.10} & 0.89 & 0.89 & 0.87 & 0.74 & \textless 0 & 0.71 & 0.78 \\
          Random Forest (ECFP) & 0.81 & 0.88 & 0.56 & 0.58 & 25.61 & 0.75 & 0.14 & 0.65 & 0.66 & 0.70 & 0.57 & -0.29 & 0.50 & 0.56\\
          LSTM & 0.58 & 1.94 & 1.04 & 0.94 & 23.67 & 1.11 & 0.23 & 0.86 & 0.00
 & 0.06 & 0.10 & 0.00 & -0.02 & 0.02 \\
          $\text{TransPolymer}_{\text{unpretrained}}$ & 0.63 & 0.61 & 0.36 & 0.46 & 20.11 & 0.59 & \textbf{0.10} & 0.84 & 0.90 & 0.89 & 0.78 & 0.27 & 0.70 & 0.80 \\
          $\text{TransPolymer}_{\text{pretrained}}$ & \textbf{0.44} &\textbf{0.52} & \textbf{0.32} & \textbf{0.39} & \textbf{16.57} & \textbf{0.52} & \textbf{0.10} & \textbf{0.92} & \textbf{0.93} & \textbf{0.91} & \textbf{0.84} & \textbf{0.50} & \textbf{0.76} & \textbf{0.82} \\

        \bottomrule
    		 
    \end{tabular}
    \end{adjustbox}
\end{table}

TransPolymer and baselines are trained on p-type polymer OPV dataset whose results are shown in Table \ref{T3}. The original paper trained random forest and artificial neural network (ANN) on the dataset using ECFP fingerprints \cite{nagasawa2018computer}. TransPolymer\textsubscript{pretrained}, in comparison with baselines, gives a slightly better performance as the average RMSE is the same as that of random forest, and the average test $R^2$ is increased by 0.05. Although all the model performance is not satisfying enough, possibly attributed to the noise in data, TransPolymer\textsubscript{pretrained} still outperforms baselines. The scatter plots of ground truth vs. predicted values for OPV by TransPolymer\textsubscript{pretrained} are depicted in Fig. \ref{Predict_vs_true}(j) and Supplementary Figure 2(j).

\begin{table}[t]
    \caption{Performance of TransPolymer and baseline models on p-type polymer OPV.}
    \label{T3}
    \centering
    \begin{adjustbox}{width=\textwidth}
    \begin{tabular}{ccccc}
        \toprule
         Model & Train RMSE (\%) ($\downarrow$) & Test RMSE (\%) ($\downarrow$) & Train $R^2$ ($\uparrow$) & Test $R^2$ ($\uparrow$)\\ \midrule
         Random Forest (ECFP) & 0.66 & \textbf{1.92} & 0.92 & 0.27 \\ 
         ANN (ECFP) & 1.58 & 2.03 & 0.55 & 0.20 \\ 
         LSTM & 2.35 & 2.34 & -0.01 & 0.00 \\
         $\text{TransPolymer}_{\text{unpretrained}}$ & 1.91 & 2.10 & 0.33 & 0.19 \\
         $\text{TransPolymer}_{\text{pretrained}}$ & \textbf{1.19} & \textbf{1.92} & \textbf{0.74} & \textbf{0.32}  \\

        \bottomrule 
    		 
    \end{tabular}
    \end{adjustbox}
\end{table}

Table \ref{performance comparison} summarizes the improvement of TransPolymer\textsubscript{pretrained} over the previous best baseline models as well as TransPolymer\textsubscript{unpretrained} on each dataset. TransPolymer\textsubscript{pretrained} has outperformed all other models on all ten datasets, further providing evidence for the generalization of TransPolymer. TransPolymer\textsubscript{pretrained} exhibits an average decrease of evaluation RMSE by 7.70\% (in percentage) and an increase of evaluation $R^2$ by 0.11 (in absolute value) compared with the best baseline models, and the two values become 18.5\% and 0.12, respectively, when it comes to comparison with TransPolymer\textsubscript{unpretrained}. Therefore, the pretrained TransPolymer could hopefully be a universal pretrained model for polymer property prediction tasks and applied to other tasks by finetuning. Besides, TransPolymer equipped with MLM pretraining technique shows significant advantages over other models in dealing with complicated polymer systems. Specifically, on PE-\uppercase\expandafter{\romannumeral1} benchmark, TransPolymer\textsubscript{pretrained} improves $R^2$ by 0.37 comparing with the previous best baseline model and by 0.39 comparing with TransPolymer\textsubscript{unpretrained}. PE-\uppercase\expandafter{\romannumeral1} contains not only polymer SMILES but also key descriptors of the materials like temperature and component ratios within the materials. The data in PE-\uppercase\expandafter{\romannumeral1} is noisy due to the existence of different types of components in the polymer materials, for instance, copolymers, anions, and ionic liquids. Also, models are trained on data from the year 2018 and evaluated on data from the year 2019, which gives a more challenging setting. Therefore it is reasonable to infer that TransPolymer is better at learning features out of noisy data and giving a robust performance. It is noticeable that LSTM becomes the least competitive model in almost every downstream task, such evidence demonstrates the significance of attention mechanisms in understanding chemical knowledge from polymer sequences.

\begin{table}[t]
    \caption{Improvement of performance of TransPolymer\textsubscript{pretrained} compared with baselines and TransPolymer\textsubscript{unpretrained} in terms of decrease of test RMSE (in percentage) and increase of test $R^2$ (in absolute value).}
    \label{performance comparison}
    \centering
        \begin{tabular}{ccc|cc}
        \toprule
         \multirow{2}*{Dataset} & \multicolumn{2}{c}{vs. best baselines} & \multicolumn{2}{c}{vs. TransPolymer\textsubscript{unpretrained}} \\ 
         ~ & RMSE ($\downarrow$) & $R^2$ ($\uparrow$) & RMSE ($\downarrow$) & $R^2$ ($\uparrow$) \\
         \midrule
         PE-\uppercase\expandafter{\romannumeral1} & -30.9\% & +0.37 & -52.2\% & +0.39 \\ 
         PE-\uppercase\expandafter{\romannumeral2} & -3.17\% & +0.01 & -23.8\% & +0.19 \\ 
         Egc & -8.33\% & +0.02 & -30.2\% & +0.08 \\
         Egb & -5.45\% & +0.02 & -14.8\% & +0.03 \\
         Eea & 0.00\% & +0.01 & -11.1\% & +0.02 \\
         Ei & -7.14\% & +0.07 & -15.2\% & +0.06 \\
         Xc & -20.1\% & +0.50 & -17.6\% & +0.23 \\
         EPS & -1.89\% & +0.05 & -11.9\% & +0.06 \\
         
         Nc & 0.00\% & +0.03 & 0.00\% & +0.02 \\
         OPV & 0.00\% & +0.05 & -8.57\% & +0.13 \\
         \midrule
         Average & -7.70\% & +0.11 & -18.5\% & +0.12 \\
         
        \bottomrule 
    		 
        \end{tabular}
    
\end{table}

\subsection{Abaltion Studies}

\begin{figure}[!t]
    \centering
    \makebox[\textwidth][c]{
        \begin{overpic}[scale=0.28]{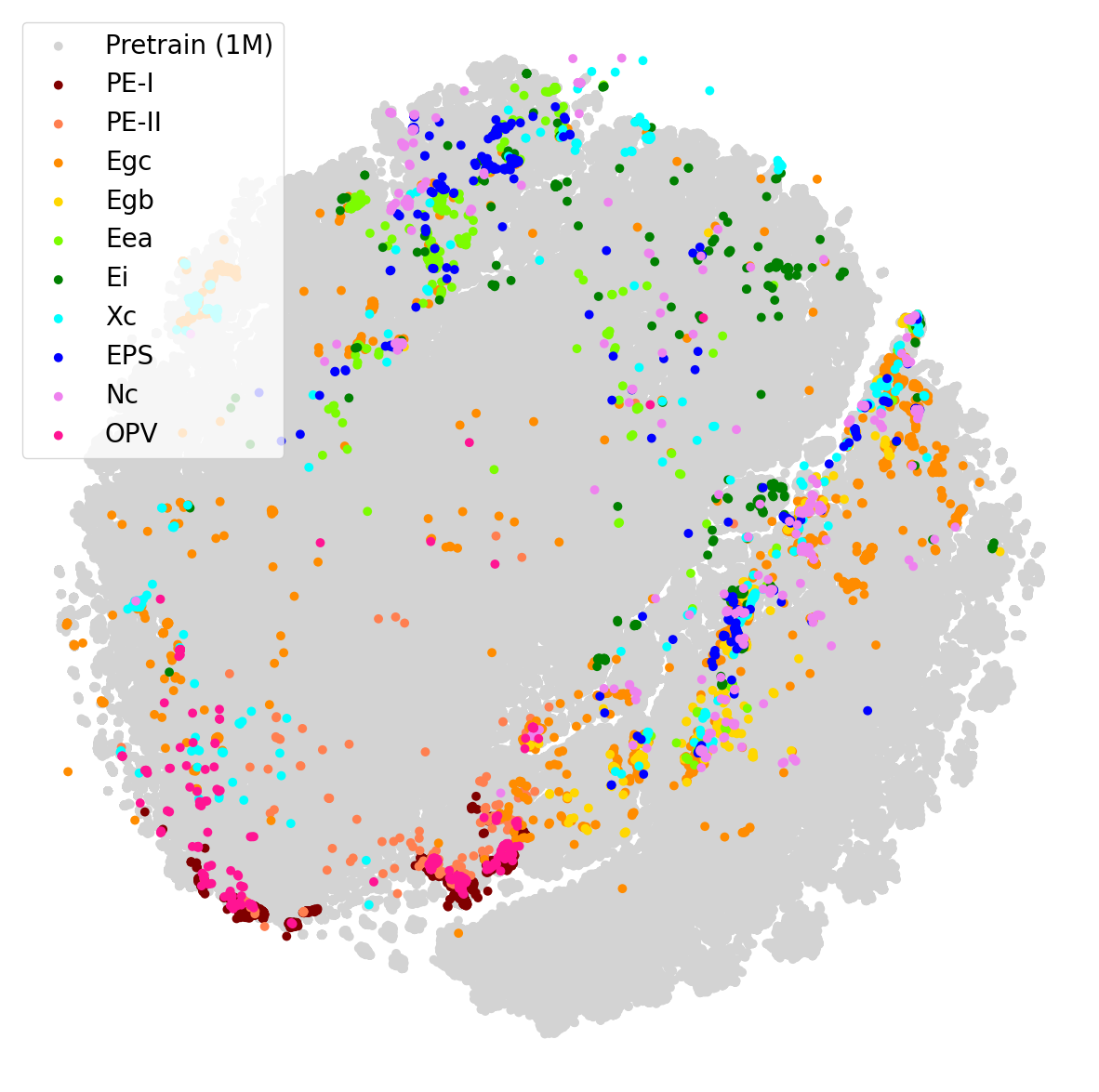}
        \put(-3,95){\textbf{a}}
    \end{overpic}
    \begin{overpic}[scale=0.28]{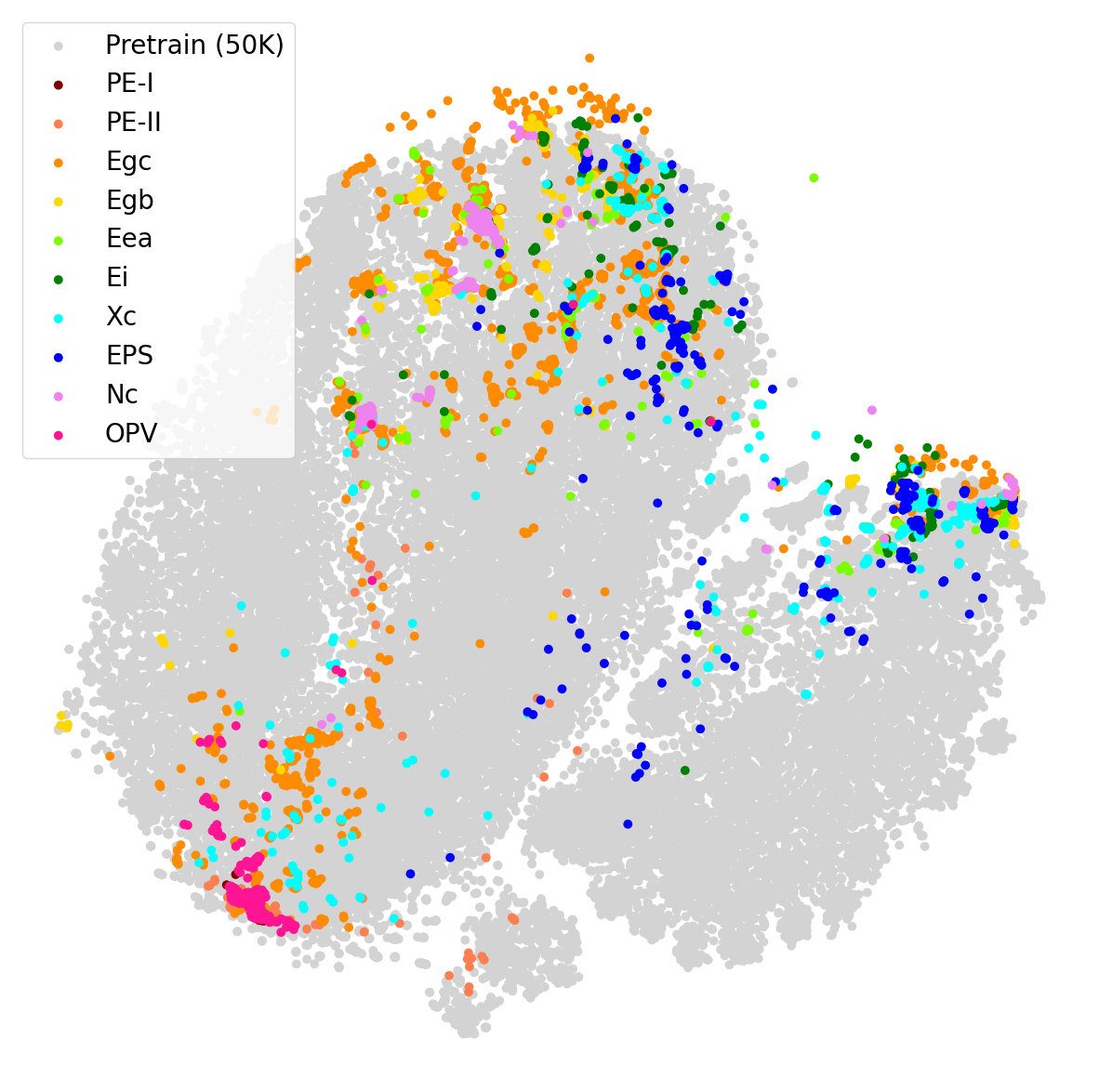}
        \put(-3,95){\textbf{b}}
    \end{overpic}
    }
    \makebox[\textwidth][c]{
        \begin{overpic}[scale=0.28]{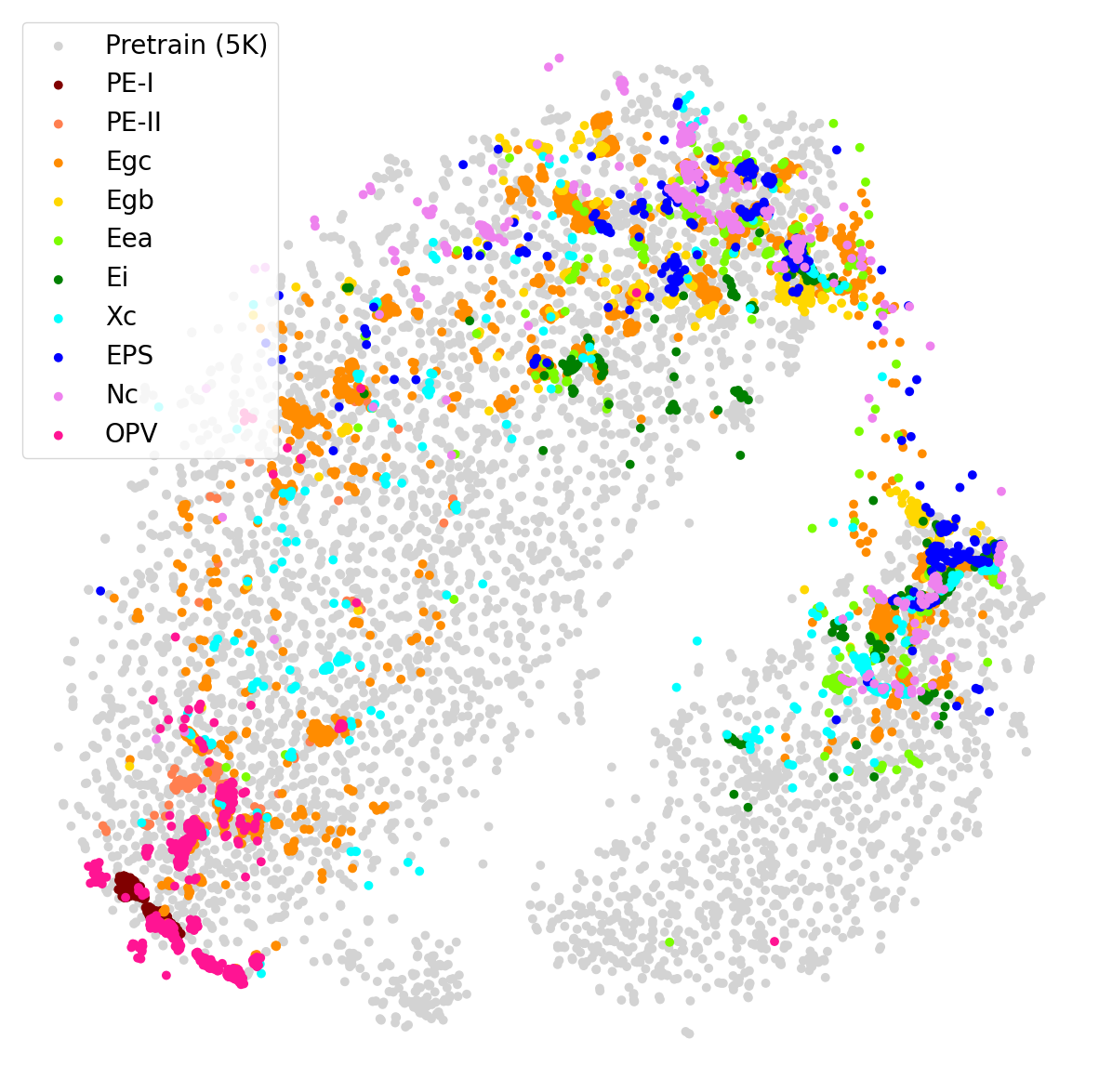}
        \put(-3,95){\textbf{c}}
    \end{overpic}
    
    }
    \caption{t-SNE visualization of pretraining and downstream data. The embeddings are obtained by first fitting on the (a) 1M (original), (b) 50K, and (c) 5K pretraining data and then transforming downstream data to the corresponding data space.}
    \label{tSNE}
\end{figure}

The effects of pretraining could be further demonstrated by the chemical space taken up by polymer SMILES from the pretraining and downstream datasets visualized by t-SNE \cite{van2008visualizing}, shown in Fig. \ref{tSNE}. Each polymer SMILES is converted to TransPolymer embedding with the size of $\text{sequence length}\times \text{embedding size}$. Max pooling is implemented to convert the embedding matrices to vectors so that the strong characteristics in embeddings could be preserved in the input of t-SNE. We use openTSNE library \cite{polivcar2019opentsne} to create 2D embeddings via pretraining data and map downstream data to the same 2D space. As illustrated in Fig. \ref{tSNE}(a), almost every downstream data point lies in the space covered by the original $\sim$1M pretraining data points, indicating the effectiveness of pretraining in better representation learning of TransPolymer. Data points from datasets like Xc which exhibit minor evidence of clustering in the chemical space cover a wide range of polymers, explaining the phenomenon that other models struggle on Xc while pretrained TransPolymer learns reasonable representations. Meanwhile, for datasets that cluster in the chemical space, other models can obtain reasonable results whereas TransPolymer achieves better results. Additionally, it should be pointed out that the numbers of unique polymer SMILES in PE-\uppercase\expandafter{\romannumeral1} and PE-\uppercase\expandafter{\romannumeral2} are much smaller than the sizes of the datasets as many instances share the same polymer SMILES while differing in descriptors like molecular weight and temperature, hence the visualization of polymer SMILES cannot fully reflect the chemical space taken up by the polymers from these datasets.

Besides, we have also investigated how the size of the pretraining dataset affects the downstream performance. We randomly pick up 5K, 50K, 500K, and 1M (original size) data points from the initial pretraining dataset without augmentation, and pretrain TransPolymer with them and compare the results with those by TransPolymer trained with 5M augmented data. The results are summarized in Supplementary Table 5. Plus, Fig. \ref{bar} presents the bar plot of $R^2$ for each experiment we have performed. Error bars are included in the figure if cross-validation is implemented in experiments. As shown in the table and the figure, the results demonstrate a clear trend of enhanced downstream performance (decreasing RMSE and increasing $R^2$) with increasing pretraining size. In particular, the model performance on some datasets, for example, PE-I, Nc, and OPV, are even worse than training TransPolymer from scratch (the results by TransPolymer\textsubscript{unpretrained} in Table \ref{T1}-\ref{T3}). A possible explanation is that the small amount of pretraining size results in the limited data space covered by pretraining data, thus making some downstream data points out of the distribution of pretraining data. Fig. \ref{tSNE}(b) and (c) visualize the data space by fitting on 50K and 5K pretraining data, respectively, in which a lot of space taken up downstream data points is not covered by pretraining data. Therefore, the results emphasize the effects of pretraining with a large number of unlabeled sequences.

\begin{figure}[!t]
    \centering
    \includegraphics[scale=0.35]{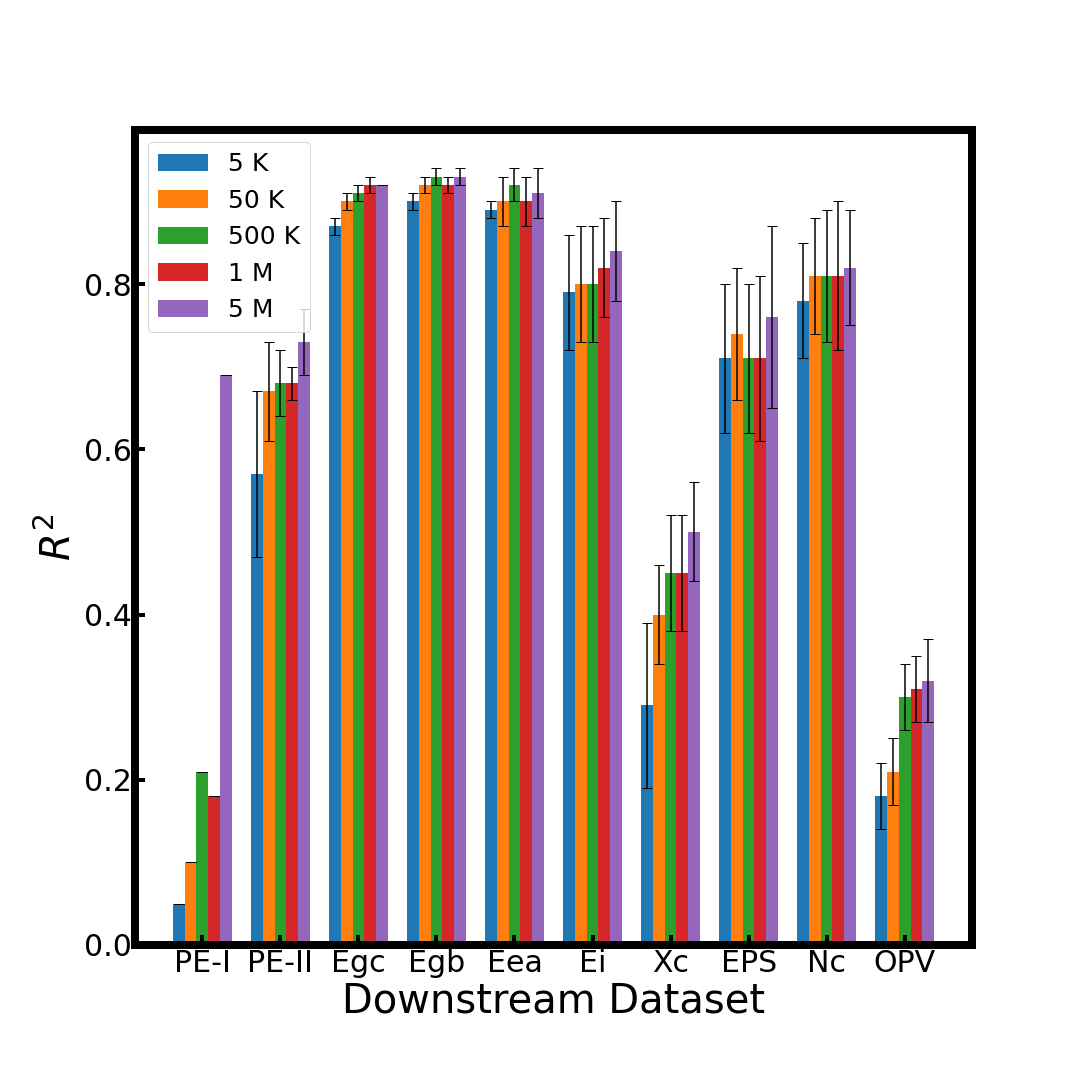}
    \caption{Bar plot of $R^2$ for each downstream task with different pretraining data sizes.}
    \label{bar}
\end{figure}

The results from TransPolymer\textsubscript{pretrained} so far are all derived by pretraining first and then finetuning the whole model on the downstream datasets. Besides, we also consider another setting where in downstream tasks only the regressor head is finetuned while the pretrained Transformer encoder is frozen. The comparison of the performance of TransPolymer\textsubscript{pretrained} between finetuning the regressor head only and finetuning the whole model is presented in Table \ref{regressor}. Standard deviation is included in the results if cross-validation is applied for downstream tasks. Reasonable results could be obtained by freezing the pretrained encoders and training the regressor head only. For instance, the model performance on Xc dataset already surpasses the baseline models, and the model performance on Ei, Nc, and OPV datasets is slightly worse than the corresponding best baselines. However, the performance on all the downstream tasks increases significantly if both the Transformer encoders and the regressor head are finetuned, which indicates that the regressor head only is not enough to learn task-specific information. In fact, the attention mechanism plays a key role in learning not only generalizable but also task-specific information. Even though the pretrained TransPolymer is transferable to various downstream tasks and more efficient, it is necessary to finetune the Transformer encoders with task-related data points for better performance.

\begin{table}[t]
    \caption{Comparison of performance of TransPolymer\textsubscript{pretrained} between finetuning the regressor head only and finetuning the whole model in terms of test RMSE and $R^2$.}
    \label{regressor}
    \centering
        \begin{tabular}{ccc|cc}
        \toprule
         \multirow{2}*{Dataset} & \multicolumn{2}{c}{Finetuning the regressor head} & \multicolumn{2}{c}{Finetuning the whole model} \\
         ~ & RMSE & $R^2$ & RMSE & $R^2$ \\ \midrule
         PE-I & 1.14 & 0.12 & 0.67 & 0.69 \\  
         PE-II & 0.91 $\pm$ 0.12 & 0.40 $\pm$ 0.10 & 0.61 $\pm$ 0.07 & 0.73 $\pm$ 0.04 \\
         Egc & 0.69 $\pm$ 0.03 & 0.81 $\pm$ 0.02 & 0.44 $\pm$ 0.01 & 0.92 $\pm$ 0.00 \\
         Egb & 0.83 $\pm$ 0.05 & 0.81 $\pm$ 0.02 & 0.52 $\pm$ 0.05 & 0.93 $\pm$ 0.01 \\
         Eea & 0.52 $\pm$ 0.04 & 0.76 $\pm$ 0.04 & 0.32 $\pm$ 0.02 & 0.91 $\pm$ 0.03 \\
         Ei & 0.51 $\pm$ 0.05 & 0.73 $\pm$ 0.05 & 0.39 $\pm$ 0.07 & 0.84 $\pm$ 0.06 \\
         Xc & 19.18 $\pm$ 2.15 & 0.34 $\pm$ 0.10 & 16.57 $\pm$ 0.68 & 0.50 $\pm$ 0.06 \\
         EPS & 0.72 $\pm$ 0.09 & 0.58 $\pm$ 0.07 & 0.52 $\pm$ 0.07 & 0.76 $\pm$ 0.11 \\
         
         Nc & 0.13 $\pm$ 0.02 & 0.70 $\pm$ 0.06 & 0.10 $\pm$ 0.02 & 0.82 $\pm$ 0.07 \\
         OPV & 2.04 $\pm$ 0.06 & 0.24 $\pm$ 0.03 & 1.92 $\pm$ 0.06 & 0.32 $\pm$ 0.05 \\
        \bottomrule 
    		 
        \end{tabular}
    
\end{table}

Data augmentation is implemented not only in pretraining but also in finetuning. The comparison between the model performance on downstream tasks with pretraining on the original $\sim$1M dataset and the augmented $\sim$5M dataset (shown in Supplementary Table 5) has already demonstrated the significance of data augmentation in model performance enhancement. In this part, we use the model pretrained on the $\sim$5M augmented pretraining dataset but finetune TransPolymer without augmenting the downstream datasets to investigate to what extent the TransPolymer model can improve the best baseline models for downstream tasks. The model performance enhancement with or without data augmentation compared with best baseline models is summarized in Table \ref{performance comparison augmentation}. For most downstream tasks, TransPolymer\textsubscript{pretrained} can improve model performance without data augmentation, while such improvement would become more significant if data augmentation is applied. For PE-II dataset, however, TransPolymer\textsubscript{pretrained} is not comparable to the best baseline model without data augmentation since the original dataset contains only 271 data points in total. Because of the data-greedy characteristics of Transformer, data augmentation could be a crucial factor in finetuning, especially when data are scarce (which is very common in chemical and materials science regimes). Therefore, data augmentation can help generalize the model to sequences unseen in training data.

\begin{table}[t]
    \caption{Improvement of performance of TransPolymer\textsubscript{pretrained} without and with data augmentation in finetuning compared with best baselines in terms of decrease of test RMSE (in percentage) and increase of test $R^2$ (in absolute value).}
    \label{performance comparison augmentation}
    \centering
        \begin{tabular}{ccc|cc}
        \toprule
         \multirow{2}*{Dataset} & \multicolumn{2}{c}{No Augmentation} & \multicolumn{2}{c}{Augmentation} \\ 
         ~ & RMSE ($\downarrow$) & $R^2$ ($\uparrow$) & RMSE ($\downarrow$) & $R^2$ ($\uparrow$) \\
         \midrule
         PE-\uppercase\expandafter{\romannumeral1} & -15.5\% & +0.22 & -30.9\% & +0.37  \\ 
         PE-\uppercase\expandafter{\romannumeral2} & +22.2\% & -0.15 & -3.17\% & +0.01 \\ 
         Egc & -4.17\% & +0.01 & -8.33\% & +0.02 \\
         Egb & +9.09\% & -0.02 & -5.45\% & +0.02 \\
         Eea & +9.38\% & 0.00 & 0.00\% & +0.01 \\
         Ei & 0.00\% & +0.03 & -7.14\% & +0.07 \\
         Xc & -14.5\% & +0.43 & -20.1\% & +0.50 \\
         EPS & +7.55\% & 0.00 & -1.89\% & +0.05 \\
         
         Nc & 0.00\% & +0.02 & 0.00\% & +0.03 \\
         OPV & +0.52\% & +0.03 & 0.00\% & +0.05 \\
         \midrule
         Average & +1.46\% & +0.06 & -7.70\% & +0.11 \\
         
        \bottomrule 
    		 
        \end{tabular}
    
\end{table}

\subsection{Self-attention Visualization}

\begin{figure}[!t]
    \centering
    
    \begin{overpic}[width=0.9\textwidth]{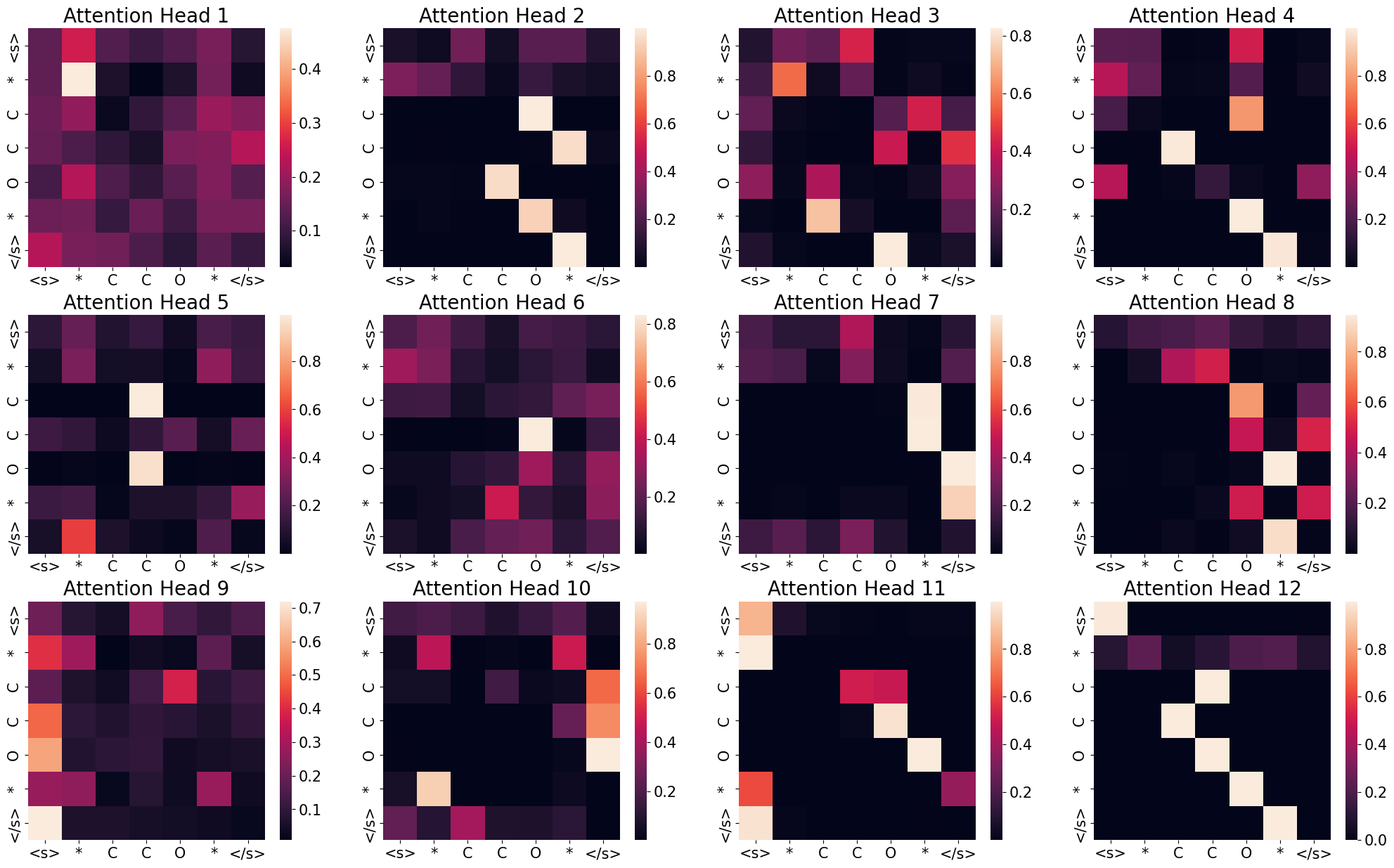}
        \put(-1, 62){\textbf{a}}
    \end{overpic}
    \vspace{10mm}

    \begin{overpic}[width=0.9\textwidth]{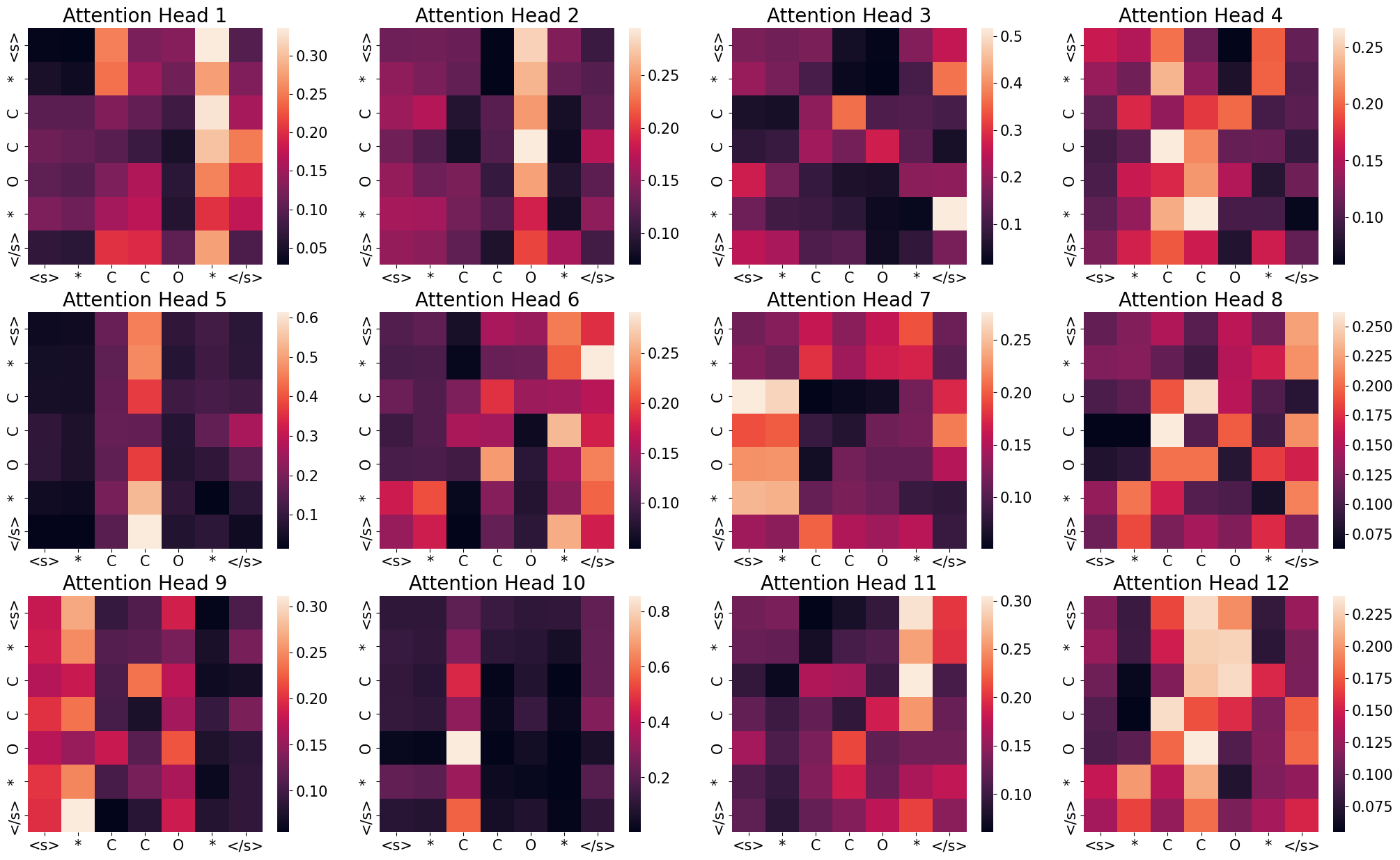}
        \put(-1,62){\textbf{b}}
    \end{overpic}
    
    \caption{Visualization of attention scores in (a) the first and (b) the last hidden layer of pretrained TransPolymer.}
    \label{Attention}
\end{figure}

AAttention scores, serving as an indicator of how closely two tokens align with each other, could be used for understanding how much chemical knowledge TransPolymer learns from pretraining and how each token contributes to the prediction results. Take poly(ethylene oxide) (*CCO*), which is one of the most prevailing polymer electrolytes, as an example. The attention scores between each token in the first and last hidden layer are shown in Fig. \ref{Attention}(a) and (b), respectively. The attention score matrices of 12 attention heads generated from the first hidden layer indicate strong relationships between tokens in the neighborhood, which could be inferred from the emergence of high attention scores around the diagonals of matrices. This trend makes sense because the nearby tokens in polymer SMILES usually represent atoms bonded to each other in the polymer, and atoms are most significantly affected by their local environments. Therefore, the first hidden layer, which is the closest layer to inputs, could capture such chemical information. In contrast, the attention scores from the last hidden layer tend to be more uniform, thus lacking an interpretable pattern. Such phenomenon has also been observed by Abnar et al. who discovered that the embeddings of tokens would become contextualized for deeper hidden layers and might carry similar information \cite{abnar2020quantifying}.

\begin{figure}[!t]
    \centering
    \begin{overpic}[width=\textwidth]
        {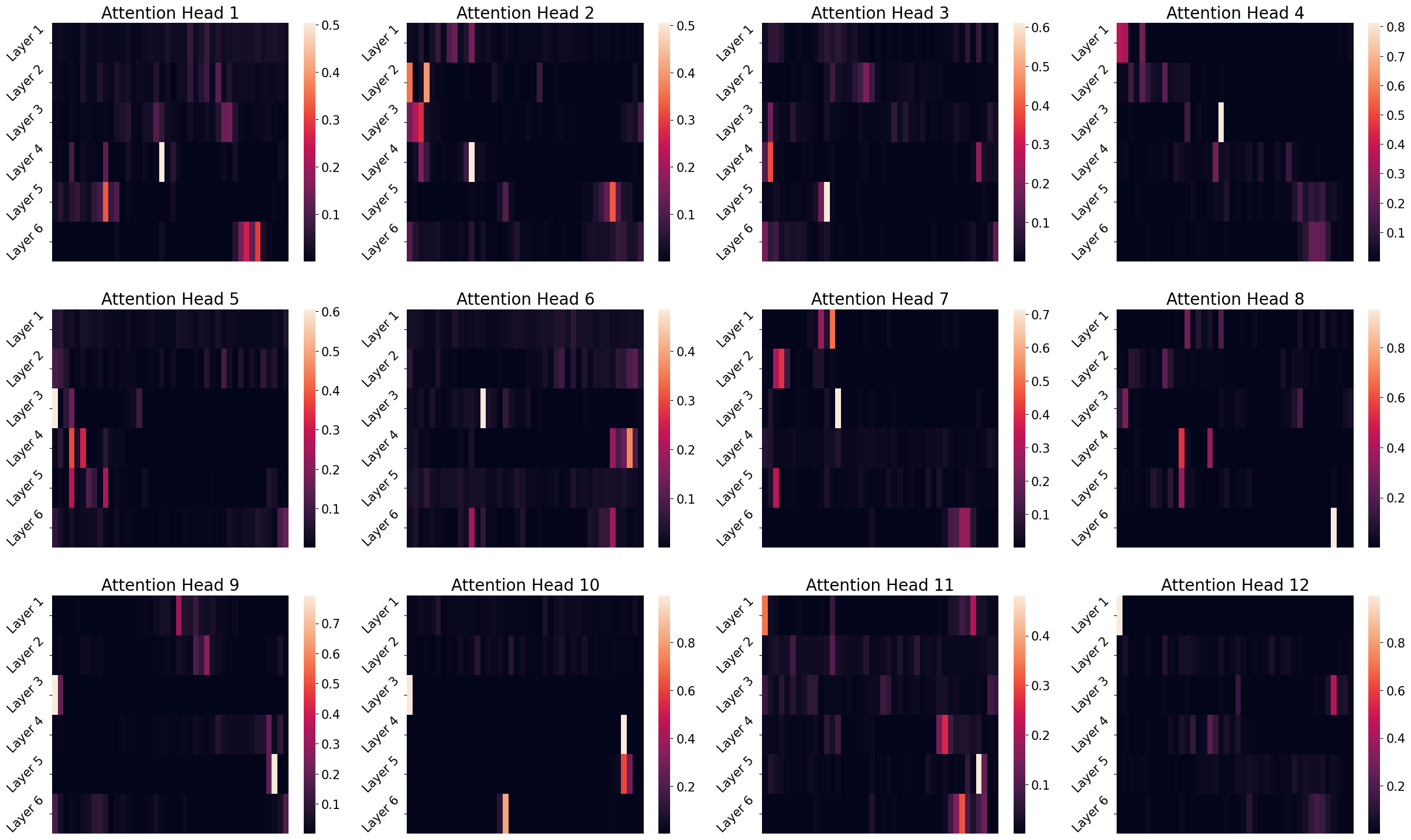}
    \end{overpic}
    \begin{overpic}[scale=0.5]
        {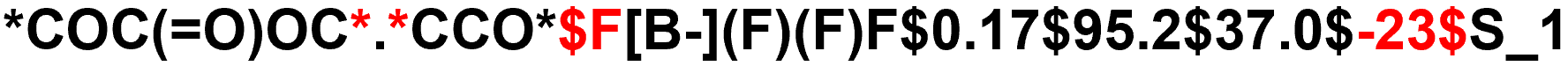}
    \end{overpic}
    \caption{Visualization of attention scores between `$\langle s \rangle$' token and other tokens at different hidden layers in each attention head. At the bottom is the sequence used for visualization in which the tokens having high attention scores with `$\langle s \rangle$' are marked in red.}
    \label{Attention CLS}
\end{figure}

When finetuning TransPolymer, the vector of the special token `$\langle s \rangle$' from the last hidden state is used for prediction. Hence, to check the impacts of tokens on prediction results, the attention scores between `$\langle s \rangle$' and other tokens from all 6 hidden layers in each attention head are illustrated with the example of the PEC-PEO blend electrolyte coming from PE-\uppercase\expandafter{\romannumeral2} whose polymer SMILES is `*COC(=O)OC*.*CCO*'. In addition to polymer SMILES, the sequence also includes `F[B-](F)(F)F', `0.17', `95.2', `37.0', `-23', and `S\_1' which stand for the anion in the electrolyte, the ratio between lithium ions and functional groups in the polymer, comonomer percentage, molecular weight (kDa), glass transition temperature ($\text{T}_{\text{g}}$), and linear chain structure, respectively. As is illustrated in Fig. \ref{Attention CLS}, the `$\langle s \rangle$' token tends to focus on certain tokens, like the `*', `\$', and `-23', which are marked in red in the example sequence in Fig. \ref{Attention CLS}. Since $\text{T}_{\text{g}}$ usually plays an important role in determining the conductivity of polymers \cite{schauser2020glass}, the finetuned Transpolyemr could understand the influential parts on properties in a polymer sequence. However, it is also widely argued that the attention weights cannot fully depict the relationship between tokens and prediction results because a high attention score does not necessarily guarantee that the pair of tokens is important to the prediction results given that attention scores do not consider Value matrices \cite{hao2021self}. More related work is needed to fully address the attention interpretation problem.

\section{Discussion}

In summary, we have proposed TransPolymer, a Transformer-based model with MLM pretraining, for accurate and efficient polymer property prediction. By rationally designing a polymer tokenization strategy, we can map a polymer instance to a sequence of tokens. Data augmentation is implemented to enlarge the available data for representation learning. TransPolymer is first pretrained on approximately 5M unlabeled polymer sequences by MLM, then finetuned on different downstream datasets, outperforming all the baselines and unpretrained TransPolymer. The superior model performance could be further explained by the impact of pretraining with large unlabeled data, finetuning Transformer encoders, and data augmentation for data space enlargement. The attention scores from hidden layers in TransPolymer provide evidence of the efficacy of learning representations with chemical awareness and suggest the influential tokens on final prediction results. 

Given the desirable model performance and outstanding generalization ability out of a small number of labeled downstream data, we anticipate that TransPolymer would serve as a potential solution to predicting newly designed polymer properties and guiding polymer design. For example, the pretrained TransPolymer could be applied in the active-learning-guided polymer discovery framework \cite{reis2021machine, tamasi2022machine}, in which TransPolymer serves to virtually screen the polymer space, recommend the potential candidates with desirable properties based on model predictions, and get updated by learning on data from experimental evaluation. In addition, the outstanding performance of TransPolymer on copolymer datasets compared with existing baseline models has shed light on the exploration of copolymers. In a nutshell, even though the main focus of this paper is placed on regression, TransPolymer can pave the way for several promising (co)polymer discovery frameworks.

\section{Methods}

\subsection{Polymer tokenization}

Unlike small molecules which are easily represented by SMILES, polymers are more complex to be converted to sequences since SMILES fails to incorporate pivotal information like connectivity between repeating units and degree of polymerization. As a result, we need to design the polymer sequences to take account of that information. To design the polymer sequences, each repeating unit of the polymer is first recognized and converted to SMILES, then `*' signs are added at the places which represent the ends of the repeating unit to indicate the connectivity between repeating units. Such a strategy to indicate repeating units has been widely used in string-based polymer representations \cite{batra2020polymers, chen2021predicting}. For the cases of copolymers, `.' is used to separate different constituents, and `\^{}' is used to indicate branches in copolymers. Other information like the degree of polymerization and molecular weight, if accessible, will be put after the polymer SMILES separated by special tokens. Take the example of the sequence given in Fig. \ref{F1}(a), the sequence describes a polymer electrolyte system including two components separated by the special token `\textbar'. Descriptors like the ratio between repeating units in the copolymer, component type, and glass transition temperature ($\text{T}_{\text{g}}$ for short) are added for each component separated by `\$', and the ratio between components and temperature are put at the end of the sequence. Adding these descriptors can improve the performance of property predictions as suggested by Patel et al. \cite{patel2022featurization}. Unique `NAN' tokens are assigned for missing values of each descriptor in the dataset. For example, `NAN\_Tg' indicates the missing value of glass transition temperature, and `NAN\_MW’ indicates the missing molecular weight at that place. These unique NAN tokens are added during finetuning to include available chemical descriptors in the datasets. Therefore, different datasets can contain different NAN tokens. Notably, other descriptors like molecular weight and degree of polymerization are omitted in this example because their values for each component are missing. However, for practical usage, these values should also be included with unique `NAN' characters. Besides, considering the varying constituents in copolymers as well as components in composites, the `NAN' tokens for ratios are padded to the maximum possible numbers. 

When tokenizing the polymer sequences, the regular expression in the tokenizer adapted from the RoBERTa tokenizer is transformed to search for all the possible elements in polymers as well as the vocabulary for descriptors and special tokens. Consequently, the polymer tokenizer can correctly slice polymers into constituting atoms. For example, `Si' which represents a silicon atom in polymer sequences would be recognized as a single token by our polymer tokenizer whereas `S' and `i' are likely to be separated into different tokens when using the RoBERTa tokenizer. Values for descriptors and special tokens are converted to single tokens as well, where all the non-text values, e.g., temperature, are discretized and treated as one token by the tokenizer.

\subsection{Data augmentation}

To enlarge the available polymer data for better representation learning, data augmentation is applied to the polymer SMILES within polymer sequences from each dataset we use. The augmentation technique is borrowed from Lambard et al. \cite{lambard2020smiles}. First, canonicalization is removed from SMILES representations; then, atoms in SMILES are renumbered by rotation of their indices; finally, for each renumbering case, grammatically correct SMILES which preserve isomerism of original polymers or molecules and prevent Kekulisation are reconstructed \cite{weininger1988smiles, eyben2010opensmile}. Also, duplicate SMILES are removed from the expanded list. SMILES augmentation is implemented by RDKit library \cite{landrum2006rdkit}. In particular, data augmentation is only applied to training sets after the train-test split to avoid information leakage.

\subsection{Transformer-based encoder}
Our TransPolymer model is based on Transformer encoder architecture \cite{vaswani2017attention}. Unlike RNN-based models which encoded temporal information by recurrence, Transformer uses self-attention layers instead. The attention mechanism used in Transformer is named Scaled Dot-Product Attention, which maps input data into three vectors: queries (Q), keys (K), and values (V). The attention is computed by first computing the dot product of the query with all keys, dividing each by $\sqrt{d_k}$ for scaling where $d_k$ is the dimension of keys, applying softmax function to obtain the weights of values, and finally deriving the attention. The dot product between queries and keys computes how closely aligned the keys are with the queries. Therefore, the attention score is able to reflect how closely related the two embeddings of tokens are. The formula of Scaled Dot-Product Attention can be written as:
\begin{equation}
    \text{Attention}(Q,K,V) = \text{softmax}\left(\dfrac{QK^{\text{T}}}{\sqrt{d_{\text{k}}}}\right)V
\end{equation}
Multi-head attention is performed instead of single attention by linearly projecting Q, K, and V with different projections and applying the attention function in parallel. The outputs are concatenated and projected again to obtain the final results. In this way, information from different subspaces could be learned by the model. 

The input of Transformer model, namely embeddings, maps tokens in sequences to vectors. Due to the absence of recurrence, word embeddings only are not sufficient to encode sequence order. Therefore, positional encodings are introduced so that the model can know the relative or absolute position of the token in the sequence. In Transformer, position encodings are represented by trigonometric functions:
\begin{equation}
    \text{PE}_{\text{pos},2\text{i}} = \sin{(\text{pos}/10000^{2\text{i}/d_{\text{model}}})}
\end{equation}
\begin{equation}
    \text{PE}_{\text{pos},2\text{i}+1} = \cos{(\text{pos}/10000^{2\text{i}/d_{\text{model}}})}
\end{equation}
where $\text{pos}$ is the position of the token and $\text{i}$ is the dimension. By this means, the relative positions of tokens could be learned by the model.

\subsection{Pretraining with MLM}
To pretrain TransPolymer with Masked Language Modeling (MLM), 15\% of tokens of a sequence are chosen for possible replacement. Among the chosen tokens, 80\% of which are masked, 10\% of which are replaced by randomly selected vocabulary tokens, and 10\% are left unchanged, in order to generate proper contextual embeddings for all tokens and bias the representation towards the actual observed words \cite{devlin2018bert}. Such a pretraining strategy enables TransPolymer to learn the ``chemical grammar" of polymer sequences by recovering the original tokens so that chemical knowledge is encoded by the model.

The pretraining database is split into training and validation sets by a ratio of 80/20. We use AdamW as the optimizer, where the learning rate is $5\times 10^{-5}$, betas parameters are (0.9, 0.999), epsilon is $1\times 10^{-6}$, and weight decay is 0. A linear scheduler with a warm-up ratio of 0.05 is set up so that the learning rate increases from 0 to the learning rate set in the optimizer in the first 5\% training steps then decreases linearly to zero. The batch size is set to 200, and the hidden layer dropout and attention dropout are set to 0.1. The model is pretrained for 30 epochs during which the binary cross entropy loss decreases steadily from over 1 to around 0.07, and the one with the best performance on the validation set is used for finetuning. The whole pretraining process takes approximately 3 days on two RTX 6000 GPUs.

\subsection{Finetuning for polymer property prediction}

The finetuning process involves the pretrained Transformer encoder and a one-layer MLP regressor head so that representations of polymer sequences could be used for property predictions.

For the experimental settings of finetuning, AdamW is set to be the optimizer whose betas parameters are (0.9, 0.999), epsilon is $1\times 10^{-6}$, and weight decay is 0.01. Different learning rates are used for the pretrained TransPolymer and regressor head. Particularly, for some experiments a strategy of layer-wise learning rate the decay (LLRD), suggested by Zhang et al. \cite{zhang2020revisiting}, is applied. Specifically, in LLRD, the learning rate is decreased layer-by-layer from top to bottom with a multiplicative decay rate. The strategy is based on the observation that different layers learn different information from sequences. Top layers near the output learn more local and specific information, thus requiring larger learning rates; while bottom layers near inputs learn more general and common information. The specific choices of learning rates for each dataset as well as other hyperparameters of the optimizer and scheduler are exhibited in Supplementary Table 2. For each downstream dataset, the model is trained for 20 epochs and the best model is determined in terms of the RMSE and $R^2$ on the test set for evaluation. 

\section*{Data Availability}

All data used in this work are publicly available. Original datasets could be found in corresponding literature \cite{ma2020pi1m, hatakeyama2020ai, schauser2021database, kuenneth2021polymer, nagasawa2018computer}. Besides, the original and processed datasets used in this work are also available at \url{https://github.com/ChangwenXu98/TransPolymer.git}.

\section*{Code Availability}
The codes developed for this work are available at \url{https://github.com/ChangwenXu98/TransPolymer.git}.

\section*{Acknowledgments}

We thank the Advanced Research Projects Agency—Energy (ARPA-E), U.S. Department of Energy, under Award No. DE-AR0001221 and the start-up fund provided by the Department of Mechanical Engineering at Carnegie Mellon University.

\section*{Author Contributions}
A.B.F., Y.W., and C.X. conceived the idea; C.X. trained and evaluated the TransPolymer model; C.X. wrote the manuscript; A.B.F. supervised the work; all authors modified and approved the manuscript.

\section*{Competing Interests}
The authors declare no competing financial or non-financial interests.

\begin{suppinfo}

Downstream data augmentation strategies. Finetuning details. Summary of baseline. Sequence length distributions of downstream data. Ground truth vs. predictions with augmented data. Model performance with varying pretraining size. LSTM Gradient Diminishing.

\end{suppinfo}

\bibliography{achemso-demo}


\end{document}


\newpage

\beginsupplement


\section*{Supplementary Methods}

\subsection*{Downstream Data Augmentation Strategies}

The data augmentation strategies for downstream datasets are summarized in Supplementary Table 1. Copolymer data are augmented by generating equivalent SMILES of each repeating unit respectively. Different limitations on the amount of augmented data are applied considering the capacity of the computational resources we have. For example, each repeating unit in the PE-I dataset is allowed to be augmented to two, and each SMILES in the OPV dataset are allowed to generate five in total. For datasets like Ei which consist of a small amount of data (usually hundreds of data) with limited sequence length (around or even smaller than 100), each polymer sequence is allowed to generate as many augmented ones as possible.

\begin{table}[H]
    \caption{Summary of data augmentation strategies for downstream datasets.}
    \label{augmentation}
    \centering
        \begin{tabular}{cc}
        \toprule
         Dataset & Augmentation strategy \\ \midrule
         PE-\uppercase\expandafter{\romannumeral1}\cite{ref_S1} & augmenting each repeating unit of (co)polymer to twice \\ 
         PE-\uppercase\expandafter{\romannumeral2}\cite{ref_S2} & augmenting each repeating unit of (co)polymer \\ 
         Egc\cite{ref_S3} & augmenting each polymer to twice \\
         Egb\cite{ref_S3} & augmenting each polymer without upper bound limit \\
         Eea\cite{ref_S3} & augmenting each polymer without upper bound limit \\
         Ei\cite{ref_S3} & augmenting each polymer without upper bound limit \\
         Xc\cite{ref_S3} & augmenting each polymer without upper bound limit \\
         EPS\cite{ref_S3} & augmenting each polymer without upper bound limit \\
         Nc\cite{ref_S3}  & augmenting each polymer without upper bound limit \\
         OPV\cite{ref_S4} & augmenting each polymer to five times \\
        \bottomrule 
    		 
        \end{tabular}
    
\end{table}

\subsection*{Finetuning Details}

When finetuning TransPolymer, we search the hyperparameters so that the model can give the best performance on the validation set. We include the set of hyperparameters for optimizers and schedulers for each downstream task which gives the best performance in Supplementary Table 2. The LR annealing strategy we use is cosine-annealing. Besides, Supplementary Table 3 summarizes the hyperparameters we use in experiments.


\begin{table}[H]
    \caption{Summary of implementation details of optimizers and schedulers for downstream tasks.}
    \label{optimizers}
    \centering
    \begin{adjustbox}{width=\textwidth}
        \begin{tabular}{cccccc}
        \toprule
         Dataset & Regressor LR & Last Hidden State LR & LR Decay Factor & Weight Decay & Warm-up Ratio \\ \midrule
         PE-\uppercase\expandafter{\romannumeral1}\cite{ref_S1} & $1\times 10^{-4}$ & $5\times 10^{-5}$ & 1.0 & 0.01 & 0.05 \\ 
         PE-\uppercase\expandafter{\romannumeral2}\cite{ref_S2} & $5\times 10^{-5}$ & $1\times 10^{-4}$ & 1.0 & 0.00001 & 0.1 \\ 
         Egc\cite{ref_S3} & $1\times 10^{-4}$ & $1\times 10^{-4}$ & 0.9 & 0.01 & 0.1 \\
         Egb\cite{ref_S3} & $5\times 10^{-5}$ & $5\times 10^{-5}$ & 0.9 & 0.01 & 0.1 \\
         Eea\cite{ref_S3} & $5\times 10^{-5}$ & $5\times 10^{-5}$ & 0.9 & 0.01 & 0.1 \\
         Ei\cite{ref_S3} & $5\times 10^{-5}$ & $5\times 10^{-5}$ & 0.9 & 0.01 & 0.1 \\
         Xc\cite{ref_S3} & $5\times 10^{-5}$ & $5\times 10^{-5}$ & 0.9 & 0.01 & 0.1 \\
         EPS\cite{ref_S3} & $5\times 10^{-5}$ & $5\times 10^{-5}$ & 0.9 & 0.01 & 0.1 \\
         Nc\cite{ref_S3}  & $5\times 10^{-5}$ & $5\times 10^{-5}$ & 0.9 & 0.01 & 0.1 \\
         OPV\cite{ref_S4} & $1\times 10^{-4}$ & $1\times 10^{-4}$ & 0.9 & 0.01 & 0.1 \\
        \bottomrule 
    		 
        \end{tabular}
    \end{adjustbox}
    
\end{table}

\begin{table}[ht]
    \caption{Finetuning hyperparameters for TransPolymer\textsubscript{pretrained}.}
    \label{finetune}
    \centering
        \begin{tabular}{cc}
        \toprule
         Hyperparameter & Range \\ \midrule
         Batch size & \{32,64\} \\
         Last Hidden State LR & \{$5 \times 10^{-5}$,$1 \times 10^{-4}$\} \\
         Regressor LR & \{$5 \times 10^{-5}$,$1 \times 10^{-4}$\} \\
         LR Decay Factor & \{0.9,1.0\} \\
         Weight Decay & \{0.01,0.00001\} \\
         Warm-up Ratio & \{0.05,0.1\} \\
         Hidden State Dropout & \{0.1, 0.5\} \\
         Attention Dropout & \{0.1, 0.5\}  \\
         Regressor Dropout & \{0.1, 0.5\}  \\
        \bottomrule 
    		 
        \end{tabular}
    
\end{table}

\subsection*{Summary of Baseline}

The implementations of baseline models follow the original architectures used in the original paper.\cite{ref_S1, ref_S2, ref_S3, ref_S4}. The codes for all the baseline models except those for the OPV dataset are available on \url{https://github.com/KanHatakeyama/ion_predictor.git},\cite{ref_S1} \url{https://github.com/nschauser/PolymerElectrolyte.git},\cite{ref_S2} and \url{https://github.com/Ramprasad-Group/multi-task-learning.git},\cite{ref_S3} respectively. We cloned the repositories and reproduced the results reported in the paper using the same architecture described in the original paper. The baseline models for the OPV dataset are realized by ourselves by first grid-searching the hyperparameters for random forest and neural network then performing 5-fold cross-validation during training. In addition, we have trained bi-directional LSTM as an additional baseline for each downstream task to provide a direct comparison between Transformer with other language models. We search the hyperparameters to obtain the best performance, which is summarized in Supplementary Table 4. Besides, we have also trained random forest with ECFP6 fingerprints\cite{ECFP} to provide a direct comparison with SOTA fingerprinting strategies. The implementation of this baseline is similar to that for the OPV dataset: we first grid-searched the hyperparameters and then performed 5-fold cross-validation during training.

\begin{table}[H]
    \caption{Hyperparameters for LSTM.}
    \label{lstm}
    \centering
        \begin{tabular}{cc}
        \toprule
         Hyperparameter & Range \\ \midrule
         \# of Layers & \{2,3,6\} \\
         Learning Rate & \{$5 \times 10^{-5}$, $1 \times 10^{-4}$, $5 \times 10^{-4}$\} \\
         Hidden Size & \{256,768\} \\
         LSTM Dropout & \{0, 0.1, 0.3\} \\
         Freq Mask Param & \{0, 6, 15\} \\
         Time Mask Ratio & \{0, 0.25\} \\
         Add Convolution Layer & \{True, False\} \\
         Input of Regressor & \{Last hidden states, All hidden states\} \\
        \bottomrule 
    		 
        \end{tabular}
    
\end{table}

\clearpage

\section*{Supplementary Discussion}

\subsection*{Sequence Length Distributions of Downstream Data}

The distributions of polymer sequences for the datasets we use for downstream tasks are illustrated in Supplementary Figure 1. The varying distributions provide evidence that TransPolymer is transferable to various polymer datasets and can learn meaningful representations from both long and short sequences.

\begin{figure*}[ht]
    \centering
    \makebox[\textwidth][c]{
        \begin{overpic}[scale=0.115]{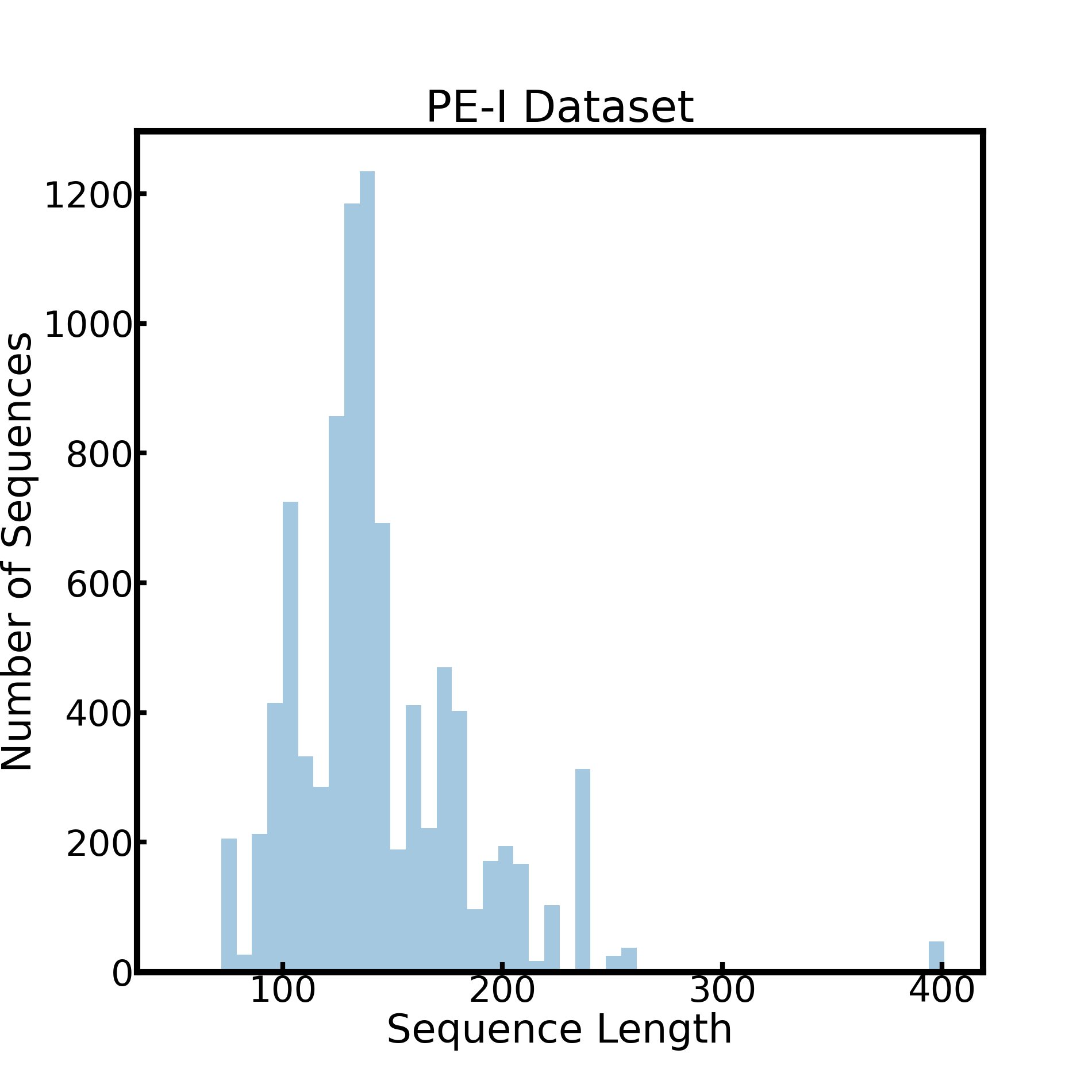}
        \put(-3,95){\textbf{a}}
    \end{overpic}
    \begin{overpic}[scale=0.115]{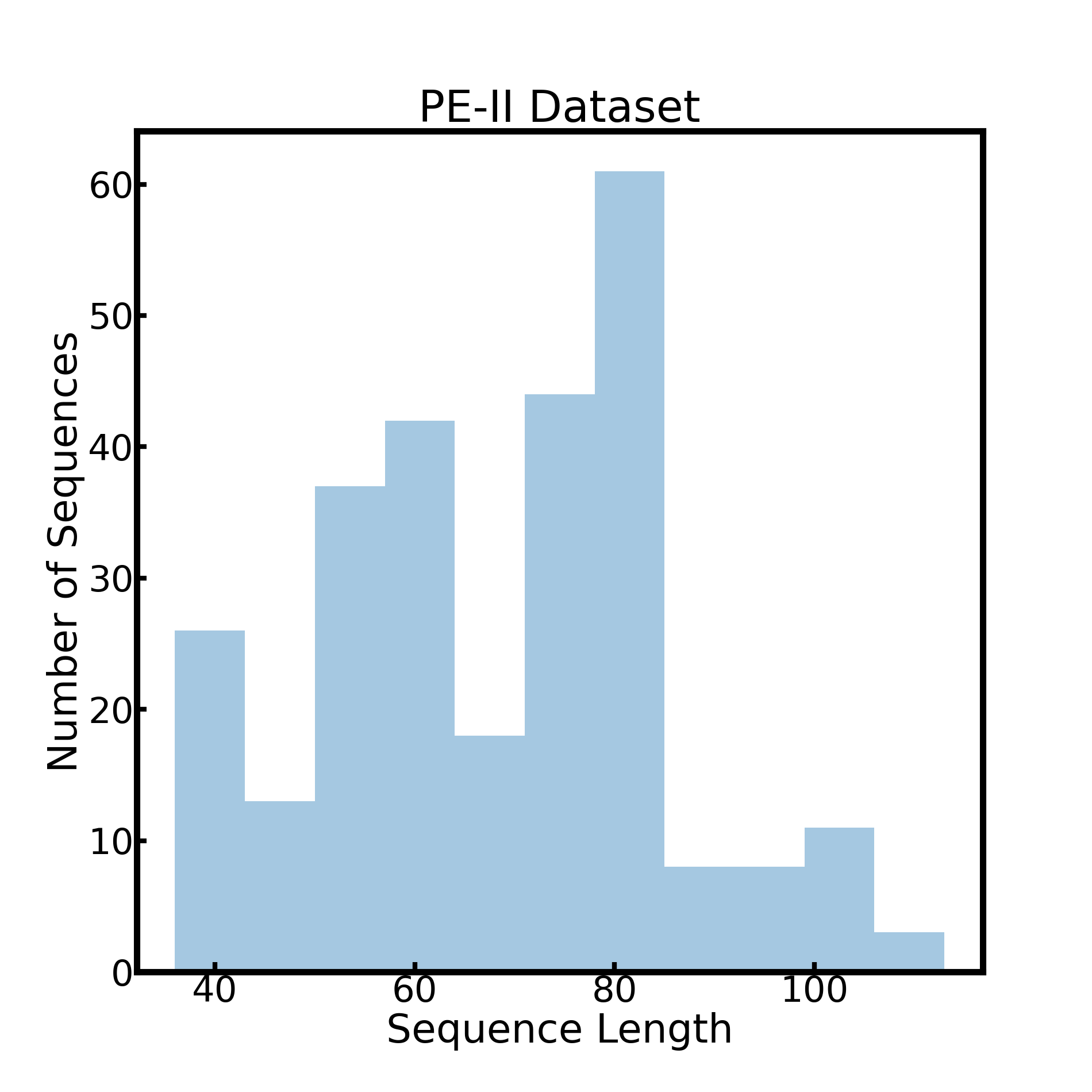}
        \put(-3,95){\textbf{b}}
    \end{overpic}
    \begin{overpic}[scale=0.115]{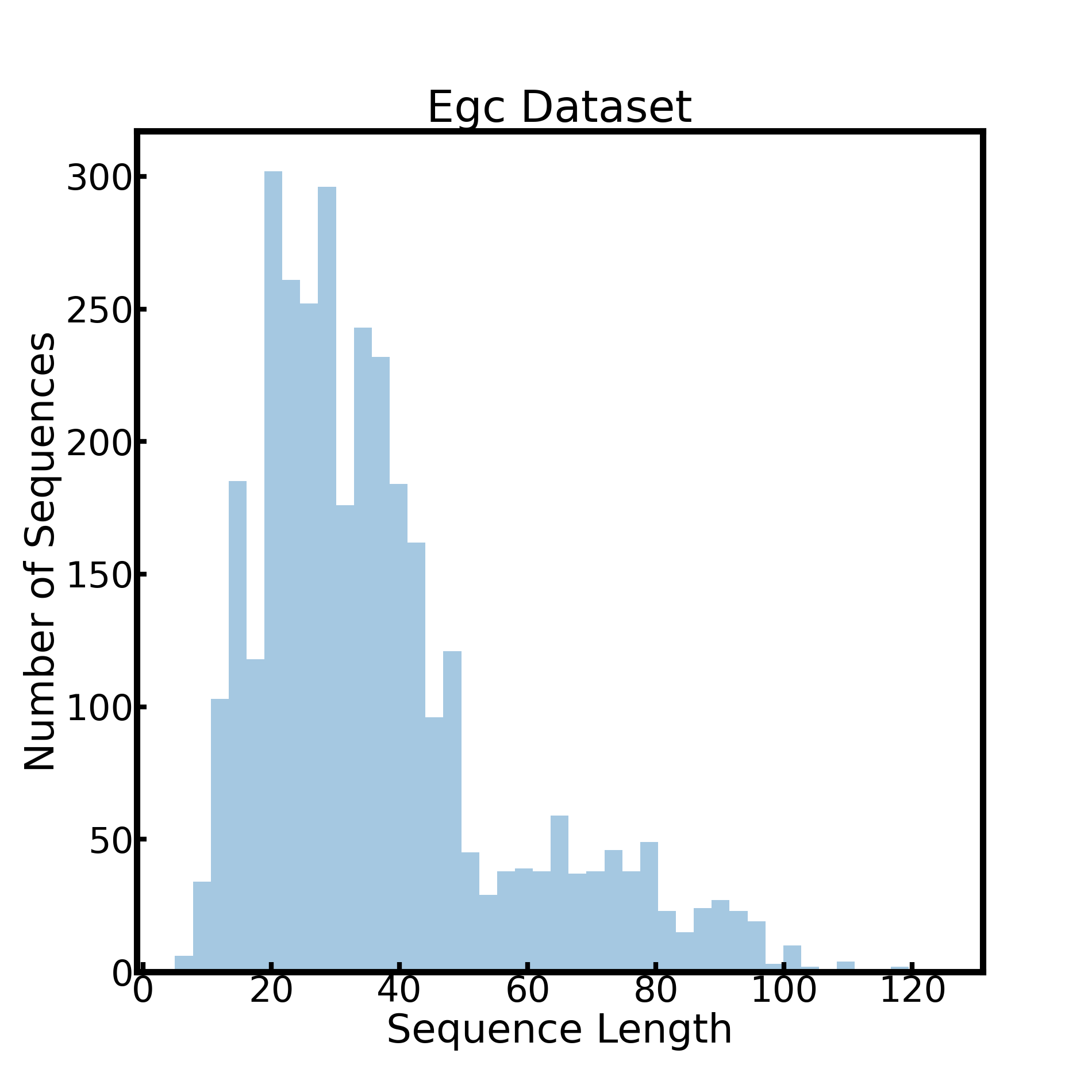}
        \put(-3,95){\textbf{c}}
    \end{overpic}
    }
    \makebox[\textwidth][c]{
    \begin{overpic}[scale=0.115]{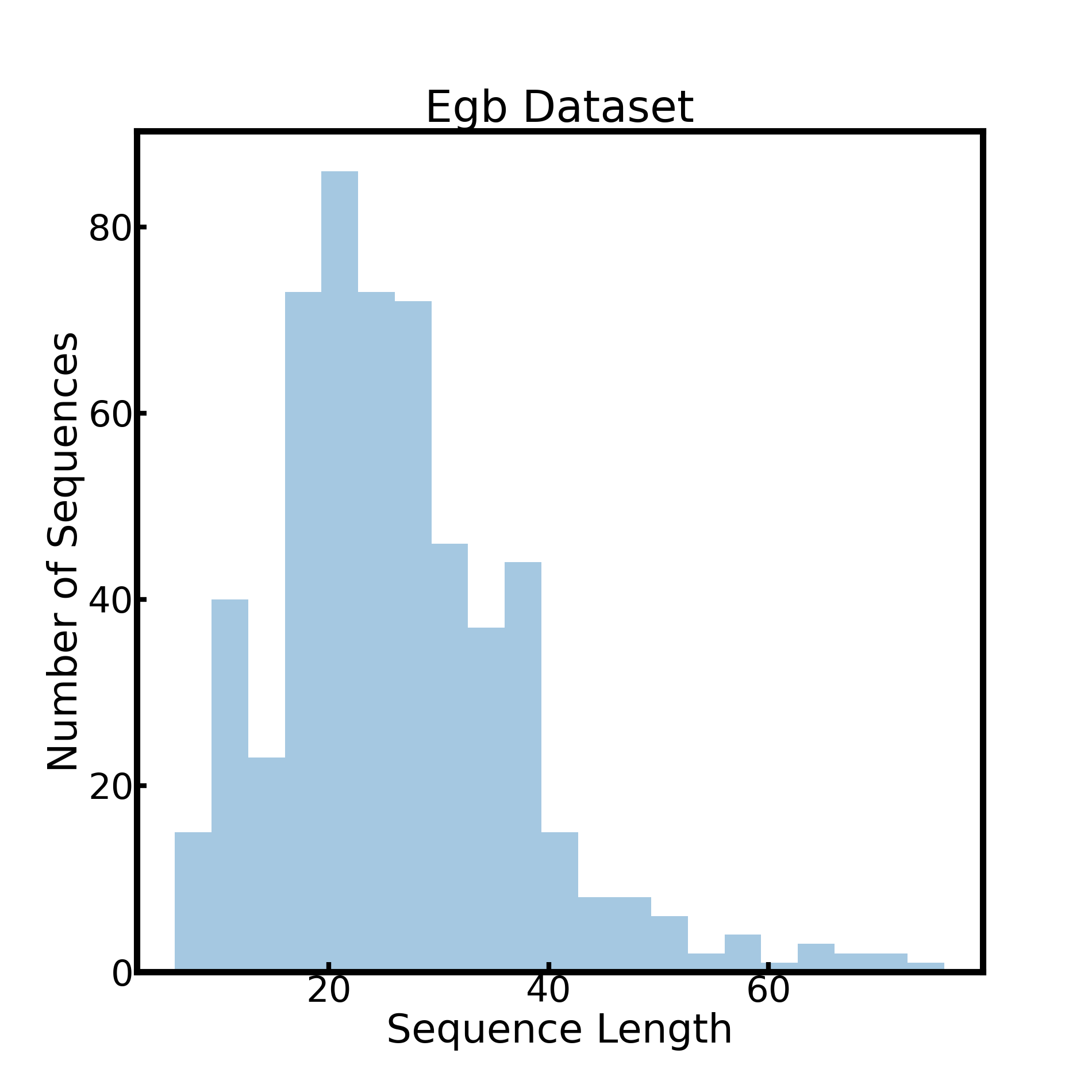}
        \put(-3,95){\textbf{d}}
    \end{overpic}
    \begin{overpic}[scale=0.115]{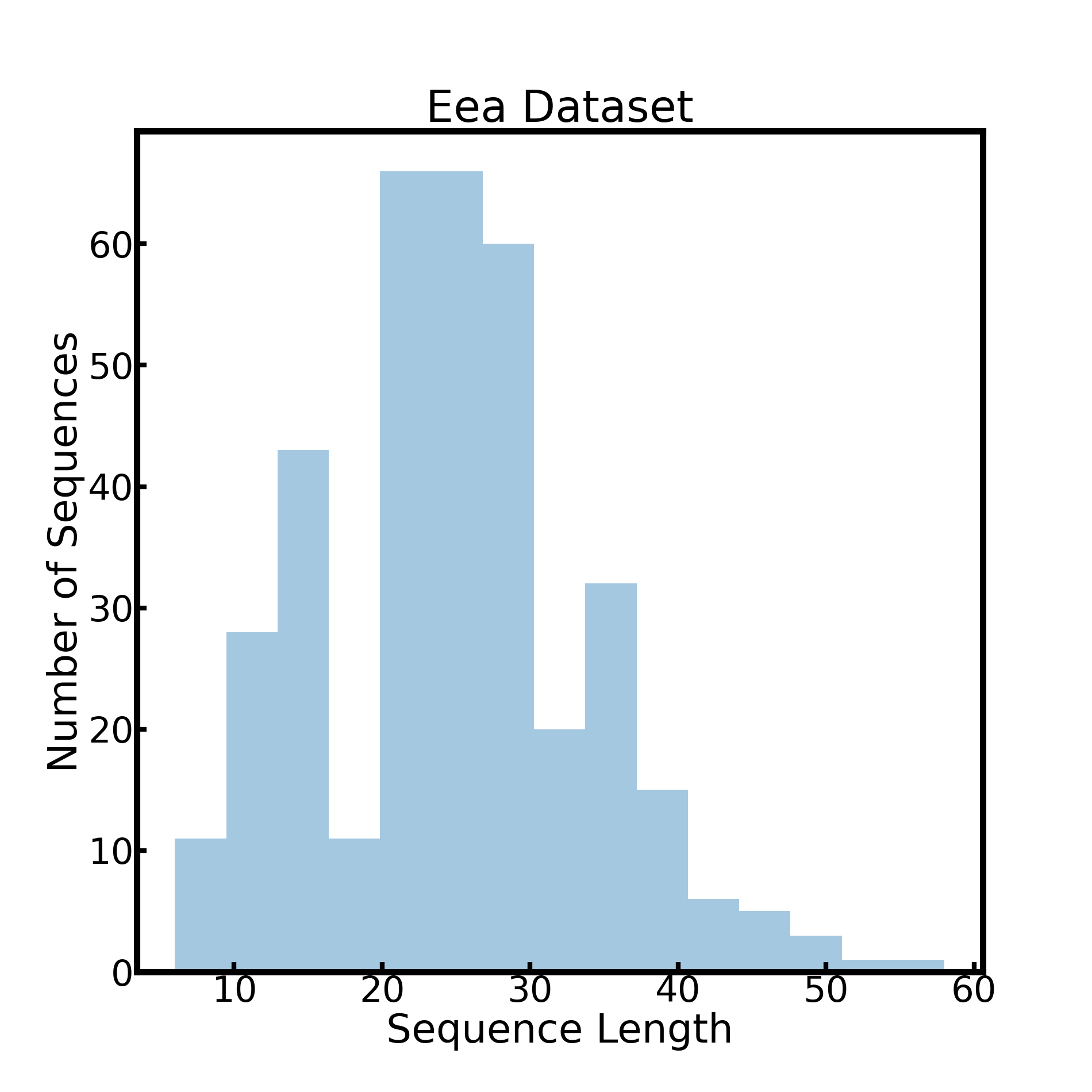}
        \put(-3,95){\textbf{e}}
    \end{overpic}
    \begin{overpic}[scale=0.115]{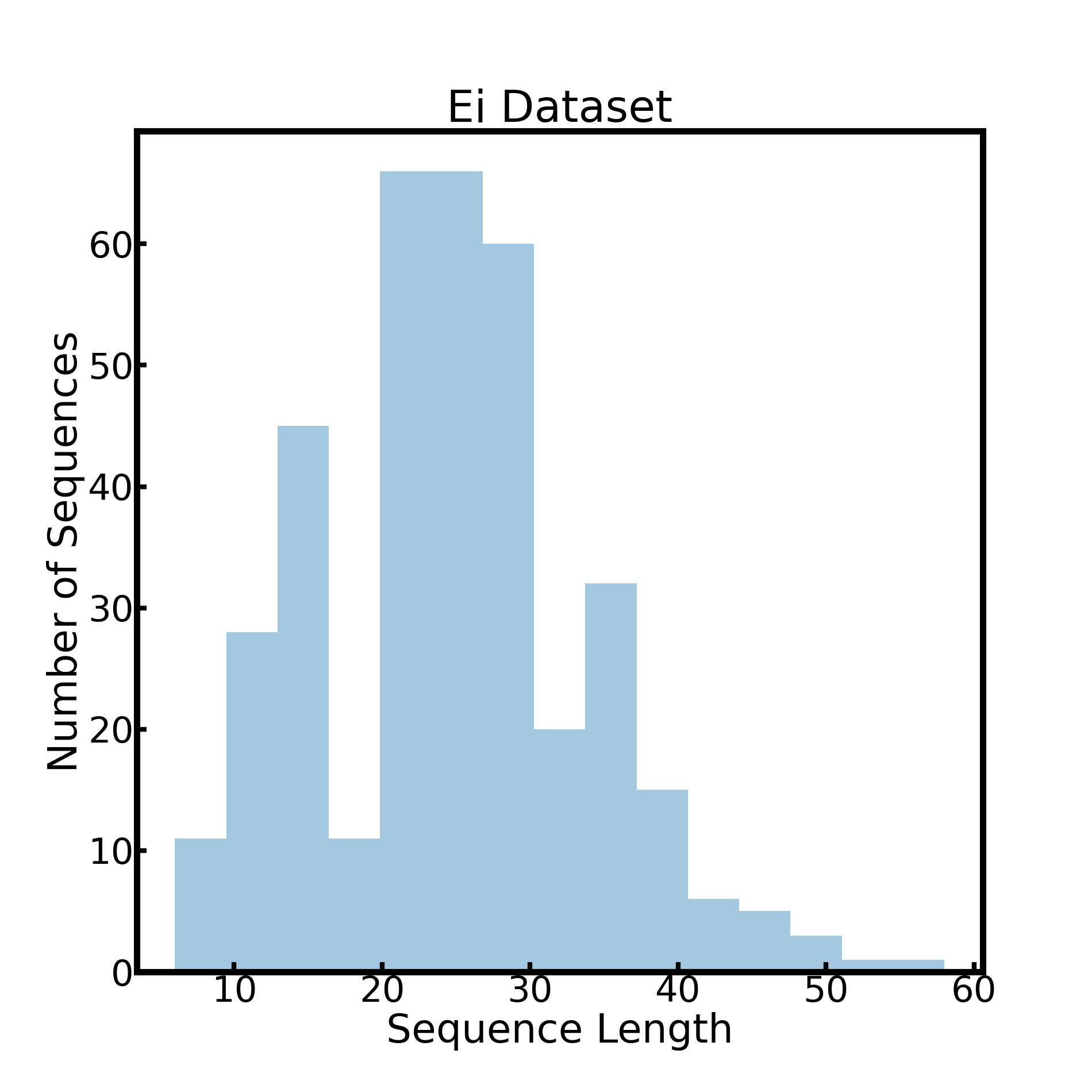}
        \put(-3,95){\textbf{f}}
    \end{overpic}
    }
    \makebox[\textwidth][c]{
    \begin{overpic}[scale=0.115]{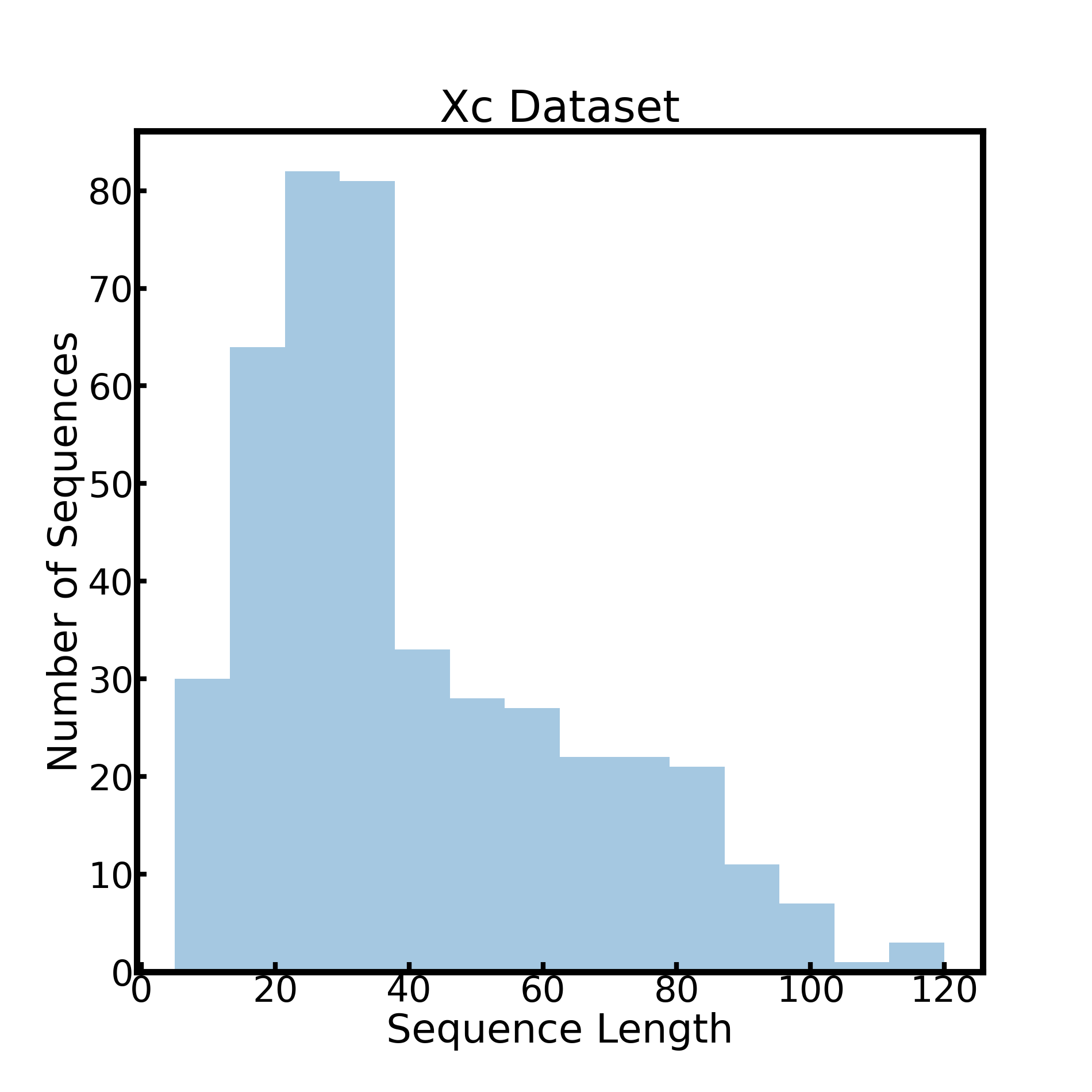}
        \put(-3,95){\textbf{g}}
    \end{overpic}
    \begin{overpic}[scale=0.115]{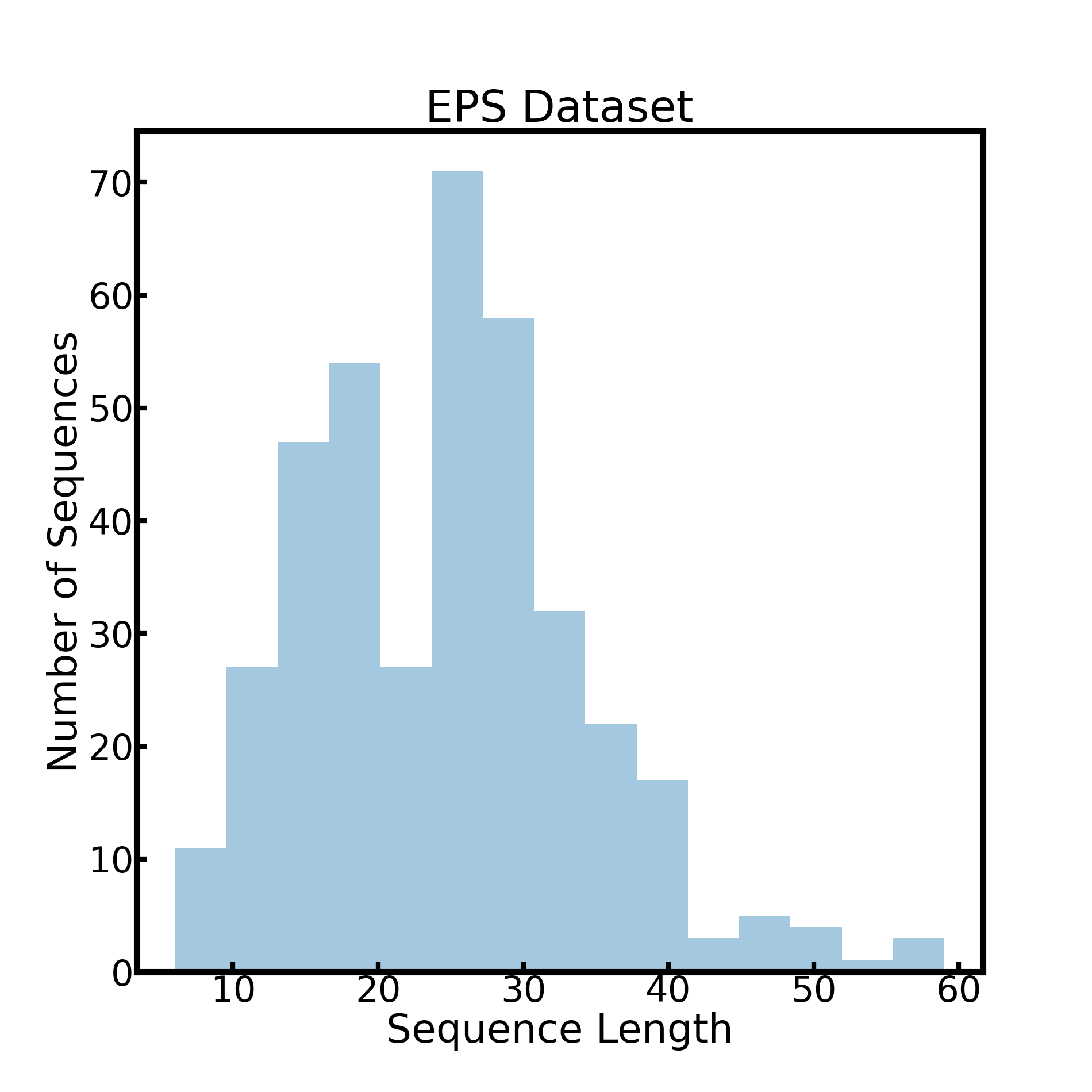}
        \put(-3,95){\textbf{h}}
    \end{overpic}
    \begin{overpic}[scale=0.115]{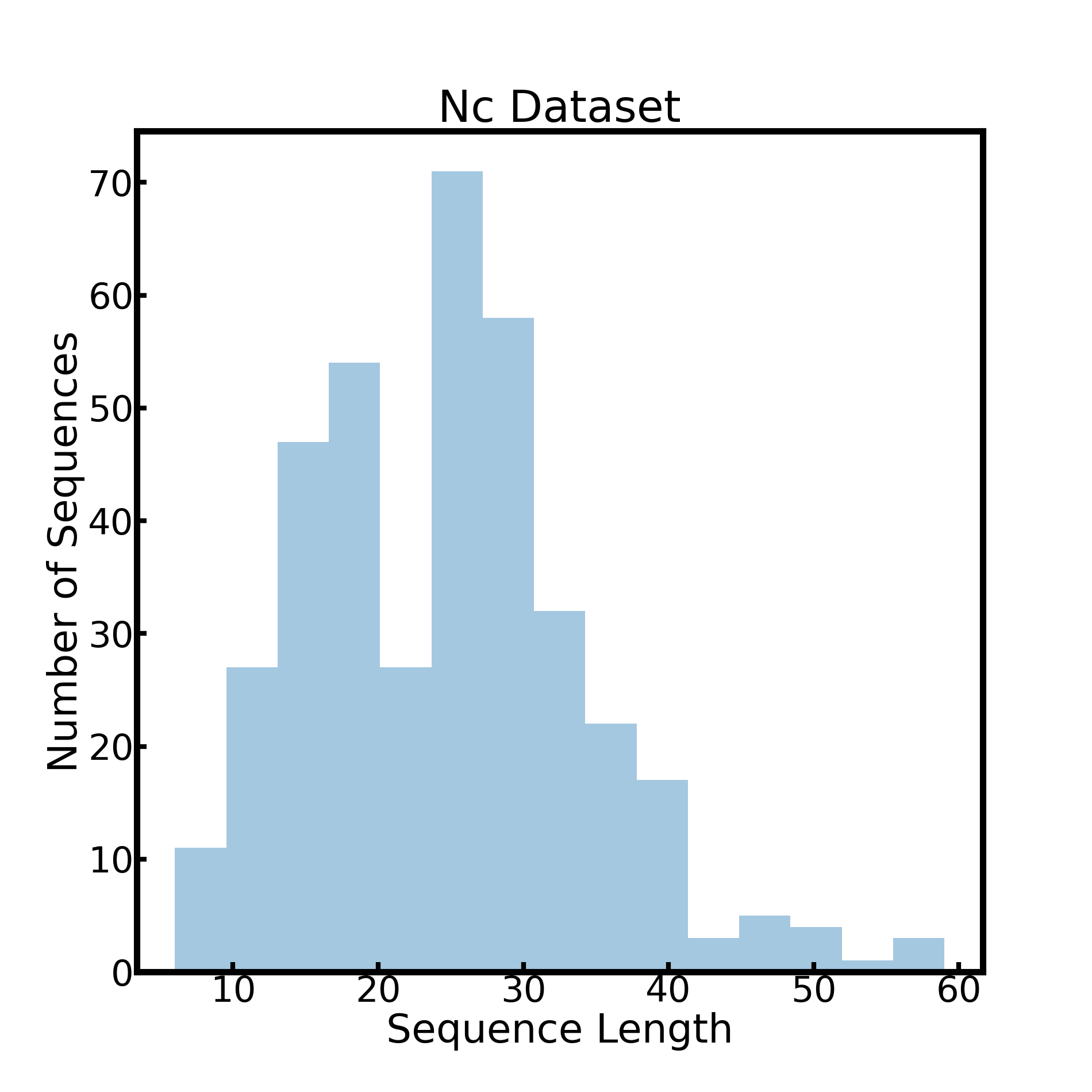}
        \put(-3,95){\textbf{i}}
    \end{overpic}
    
    }
    \makebox[\textwidth][c]{
    \begin{overpic}[scale=0.115]{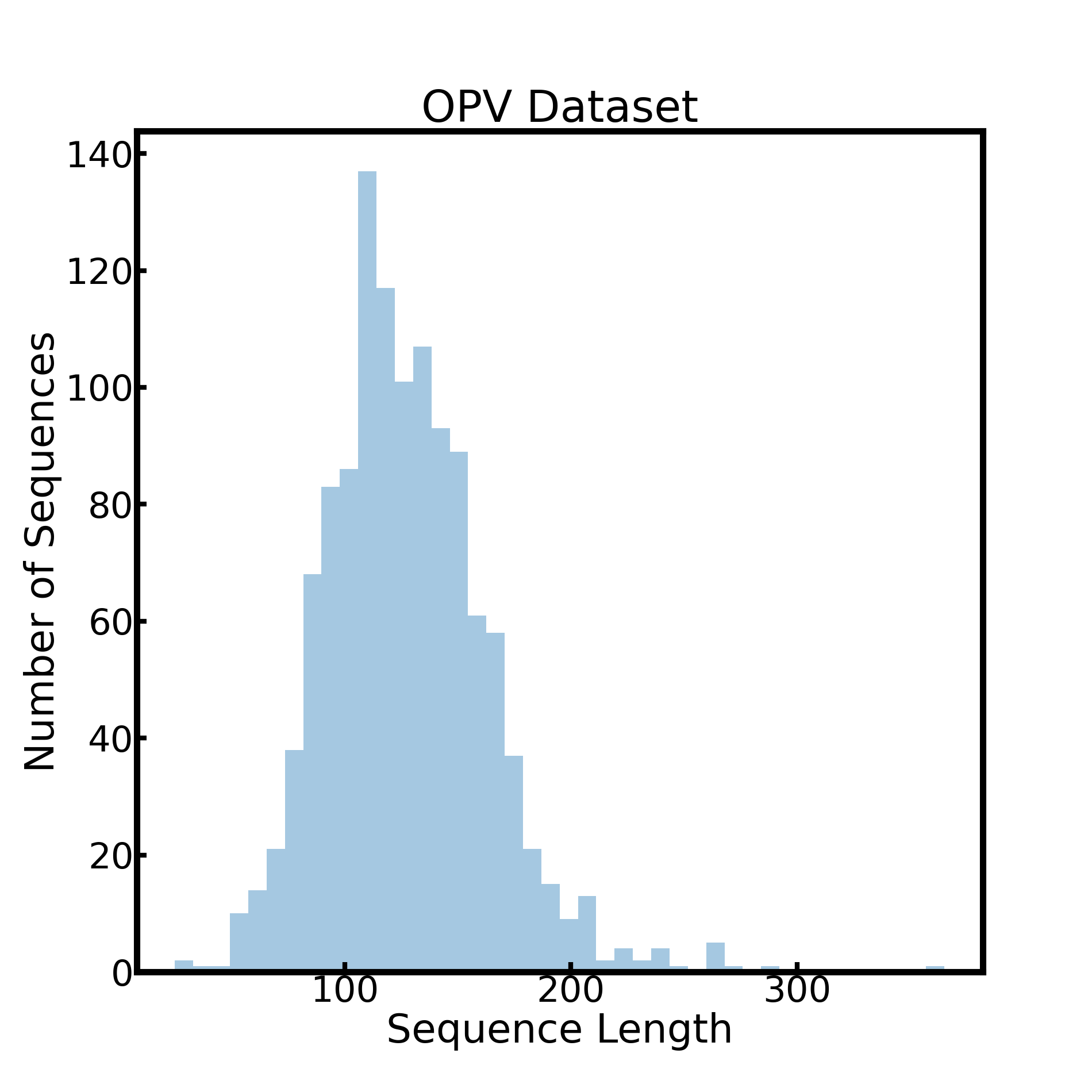}
        \put(-3,95){\textbf{j}}
    \end{overpic}
    }
    \caption{Sequence length distributions of downstream datasets before data augmentation: (a) PE-\uppercase\expandafter{\romannumeral1}, (b) PE-\uppercase\expandafter{\romannumeral2}, (c) Egc, (d) Egb, (e) Eea, (f) Ei, (g) Xc, (h) EPS, (i) Nc, and (j) OPV.}
    \label{length}
    
\end{figure*}

\subsection*{Ground Truth vs. Predictions with Augmented Data}

Supplementary Figure 2 presents the scatter plots of ground truth vs. predicted values for the downstream tasks. In the plots, we not only include training and test data contained in the original datasets provided by the literature for reference\cite{ref_S1, ref_S2, ref_S3, ref_S4} but also the augmented data generated from training data. The true value of the augmented data is the same as that of the corresponding original training data point, while the predicted property values of augmented data are clustered around the predicted value of the original training data point. 

\begin{figure*}[htbp]
    \centering
    \makebox[\textwidth][c]{
        \begin{overpic}[scale=0.125]{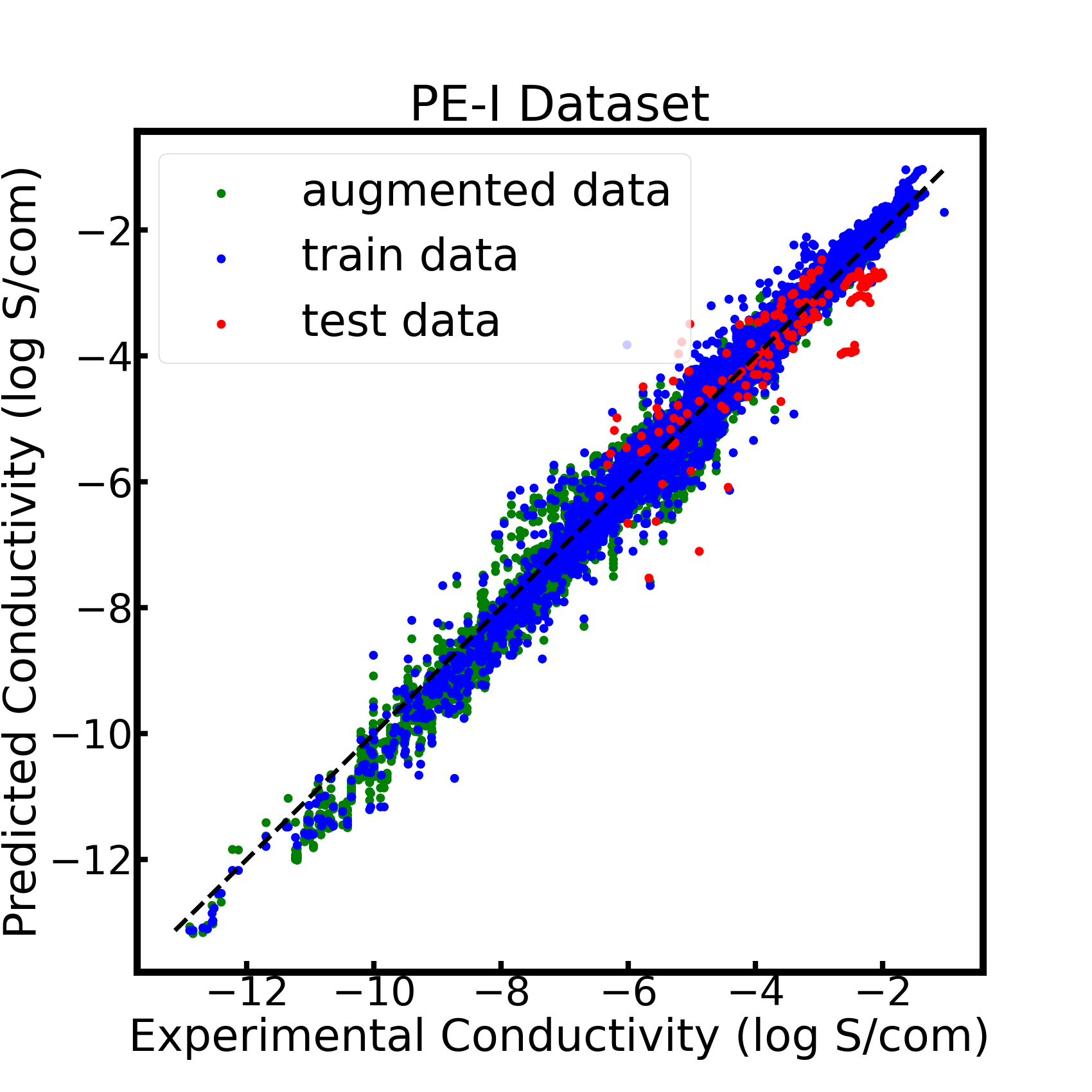}
        \put(-3,95){\textbf{a}}
    \end{overpic}
    \begin{overpic}[scale=0.125]{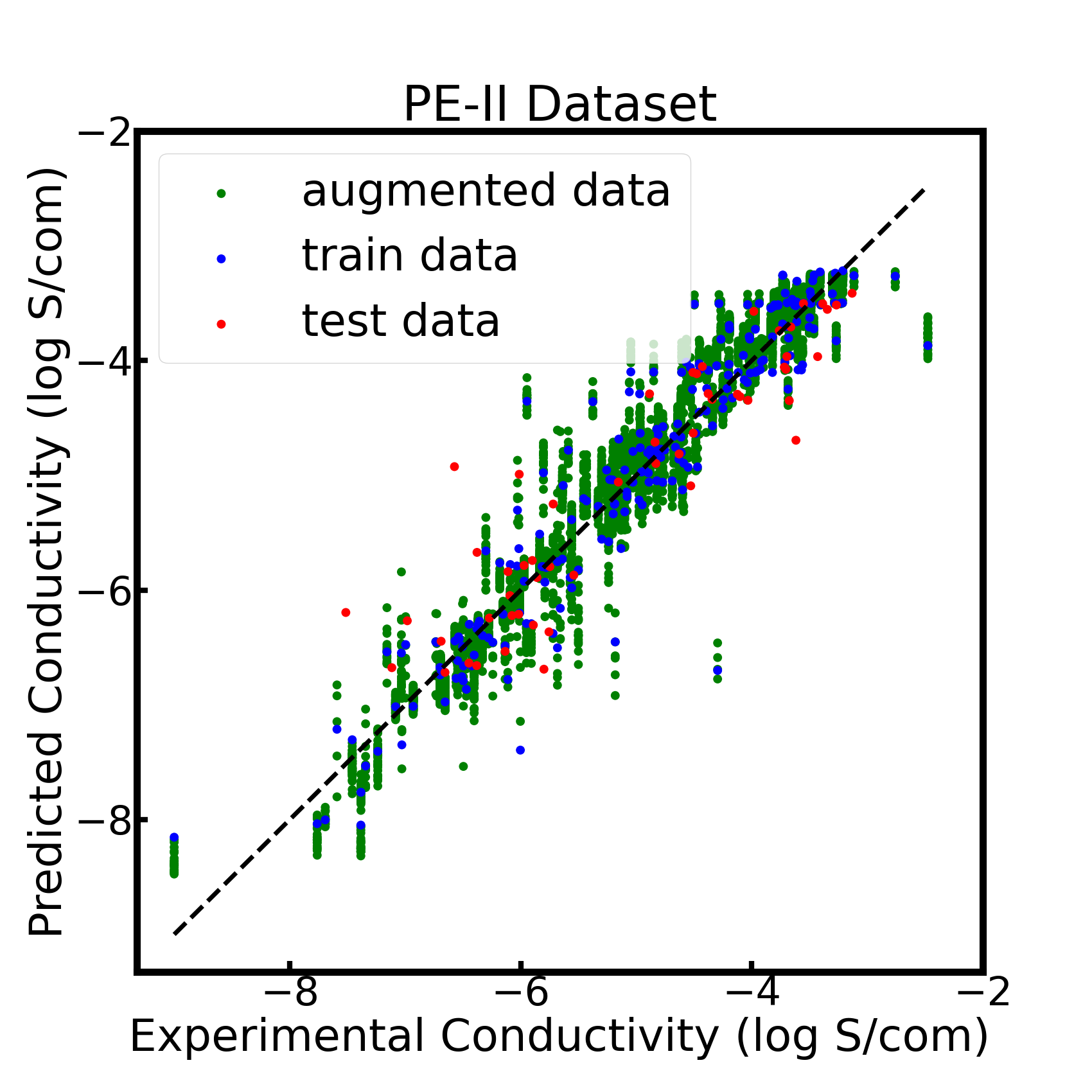}
        \put(-3,95){\textbf{b}}
    \end{overpic}
    \begin{overpic}[scale=0.125]{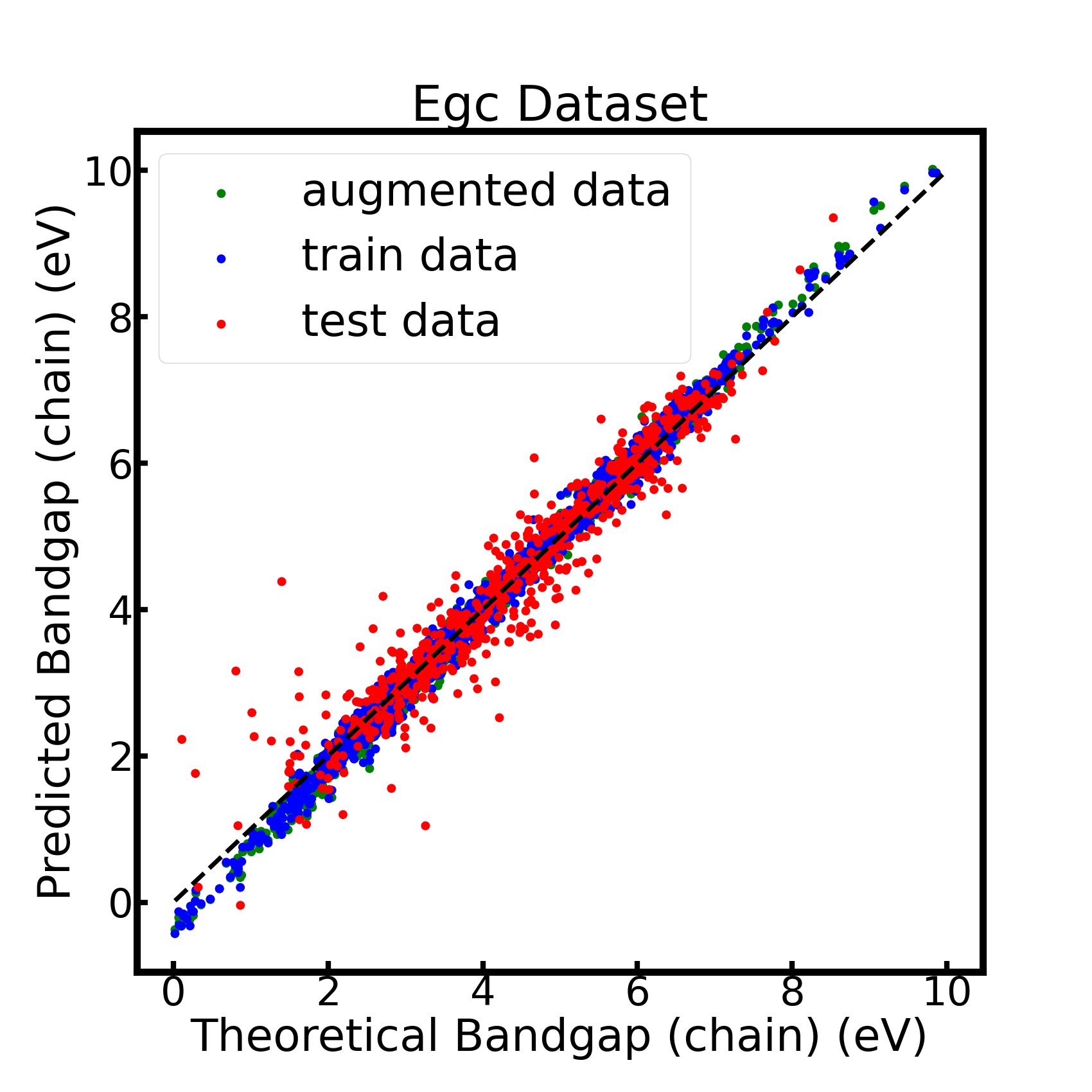}
        \put(-3,95){\textbf{c}}
    \end{overpic}
    }
    \makebox[\textwidth][c]{
    \begin{overpic}[scale=0.125]{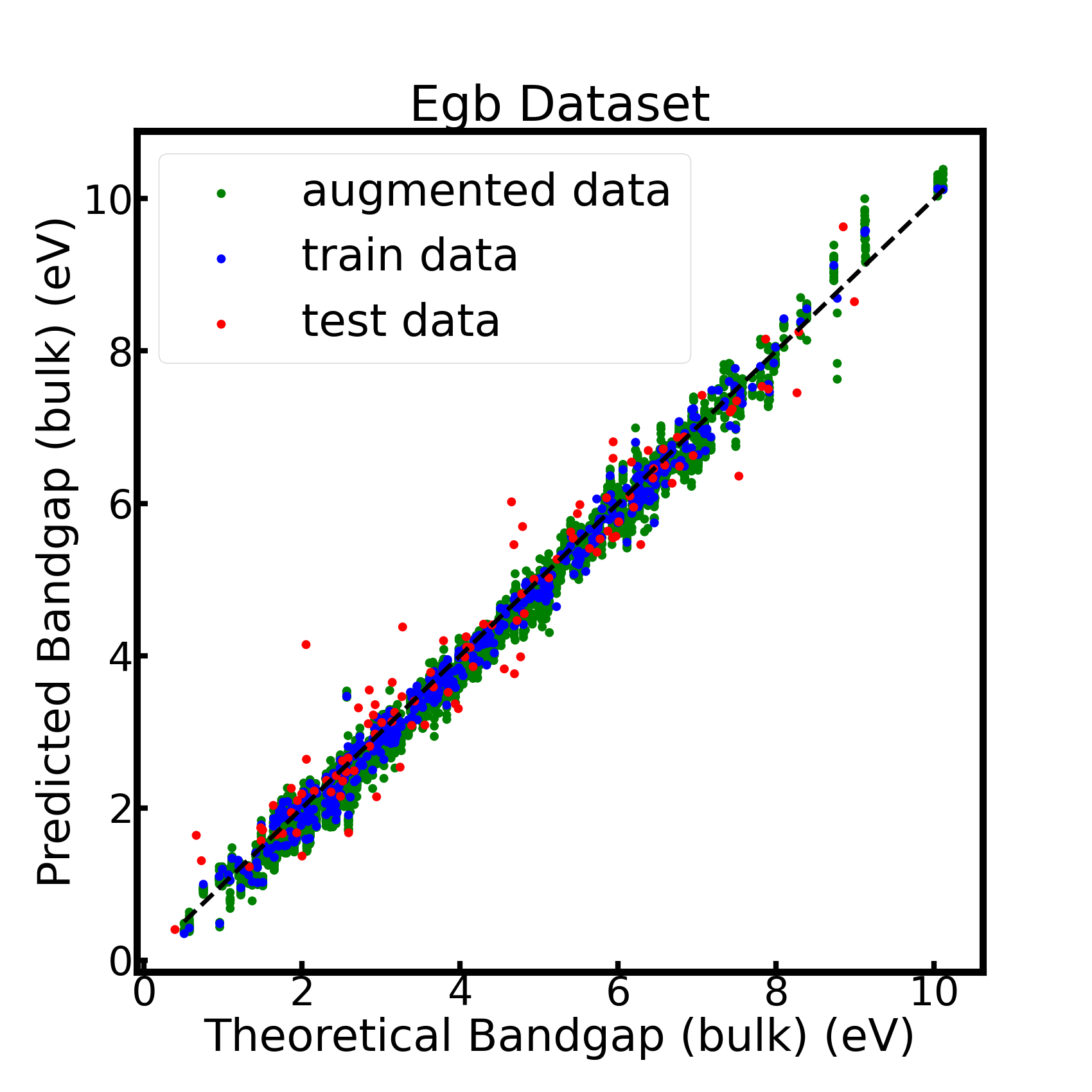}
        \put(-3,95){\textbf{d}}
    \end{overpic}
    \begin{overpic}[scale=0.125]{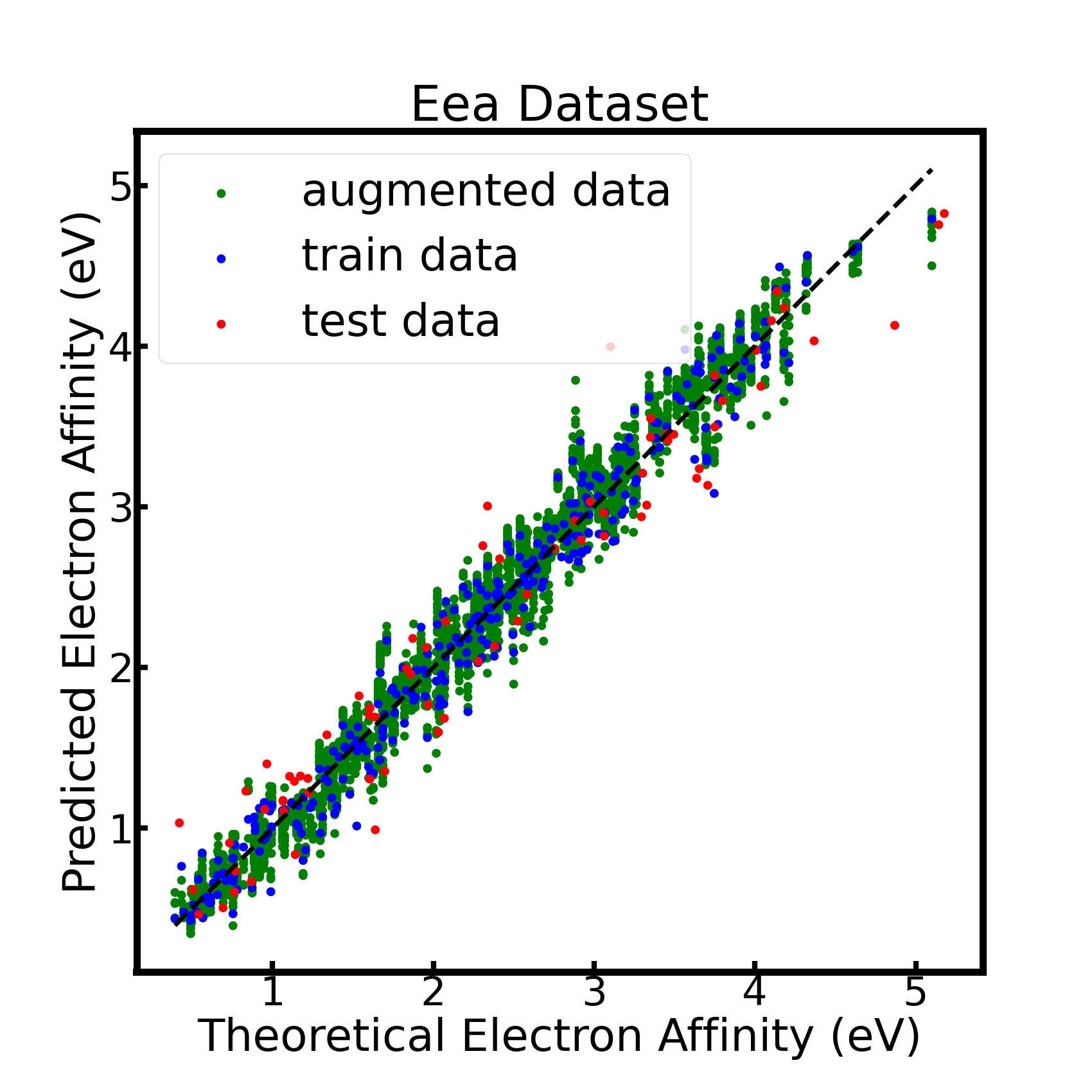}
        \put(-3,95){\textbf{e}}
    \end{overpic}
    \begin{overpic}[scale=0.125]{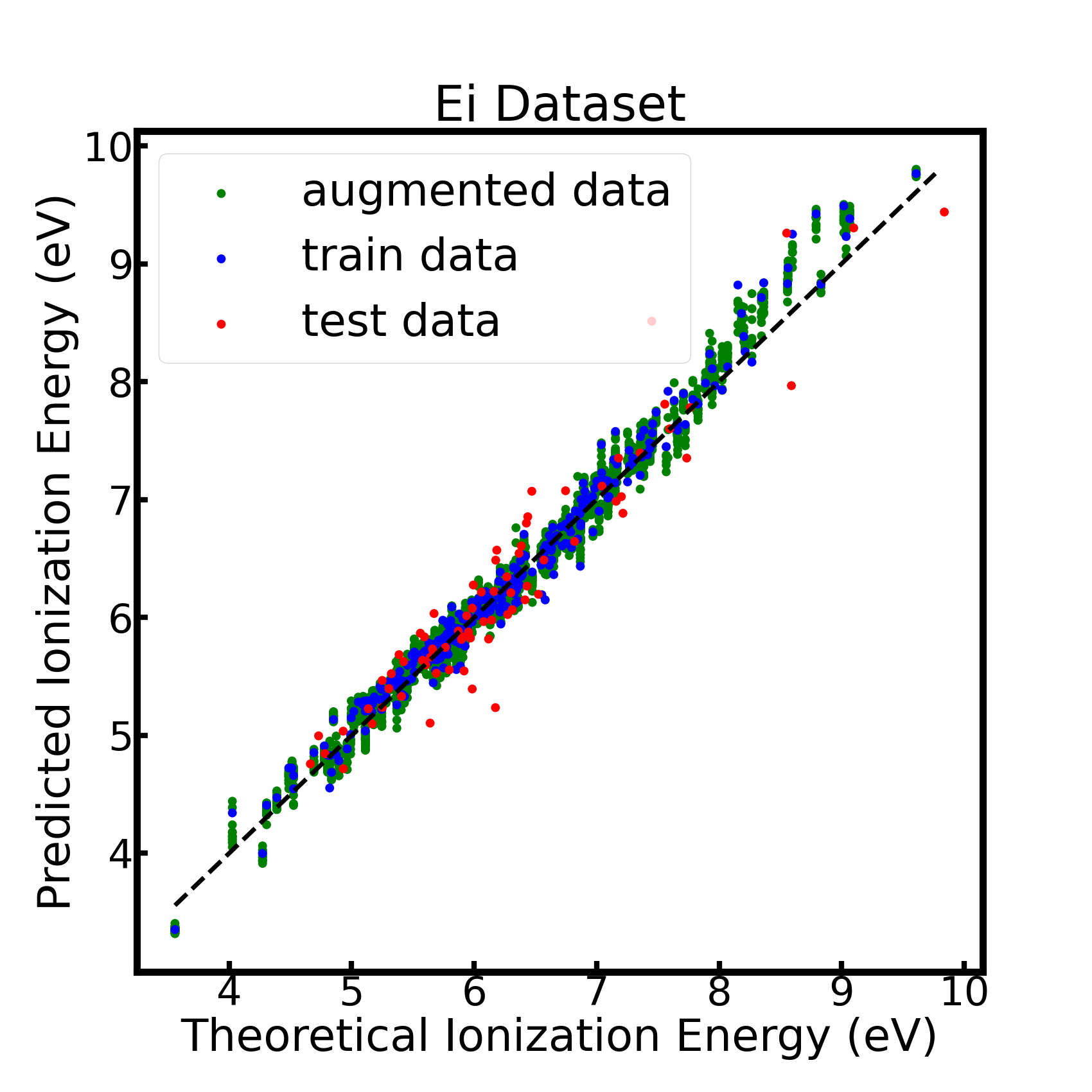}
        \put(-3,95){\textbf{f}}
    \end{overpic}
    }
    \makebox[\textwidth][c]{
    \begin{overpic}[scale=0.125]{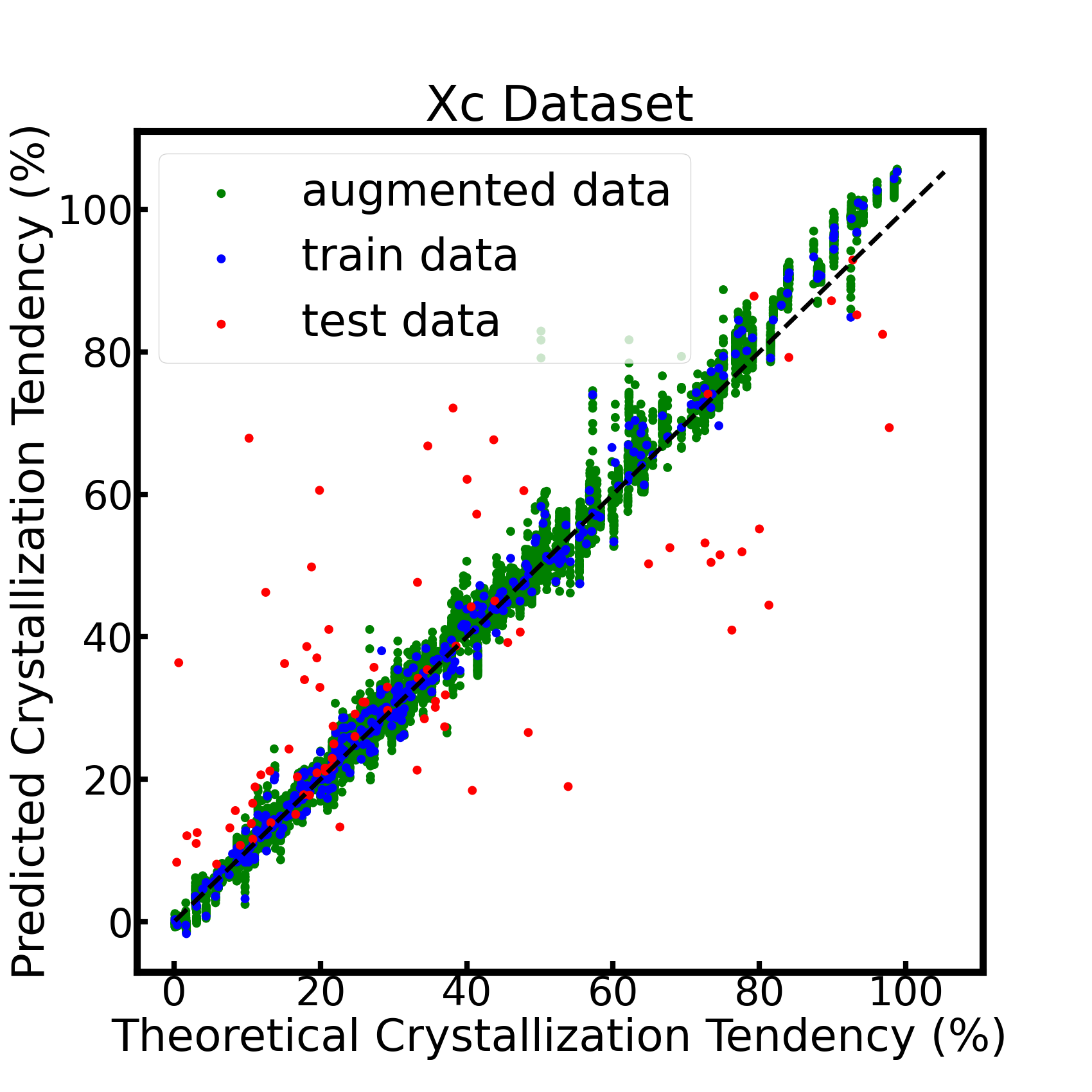}
        \put(-3,95){\textbf{g}}
    \end{overpic}
    \begin{overpic}[scale=0.125]{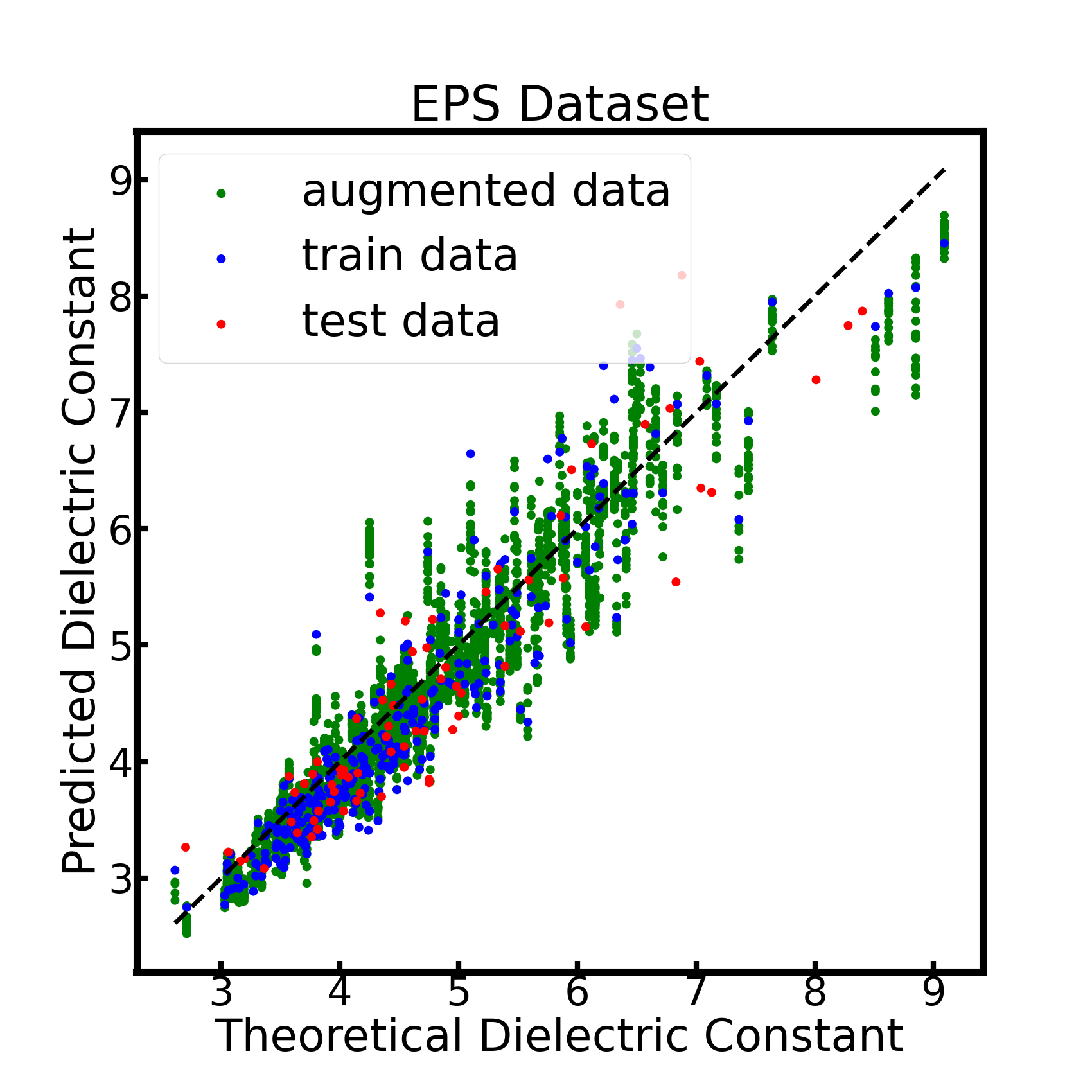}
        \put(-3,95){\textbf{h}}
    \end{overpic}
    \begin{overpic}[scale=0.125]{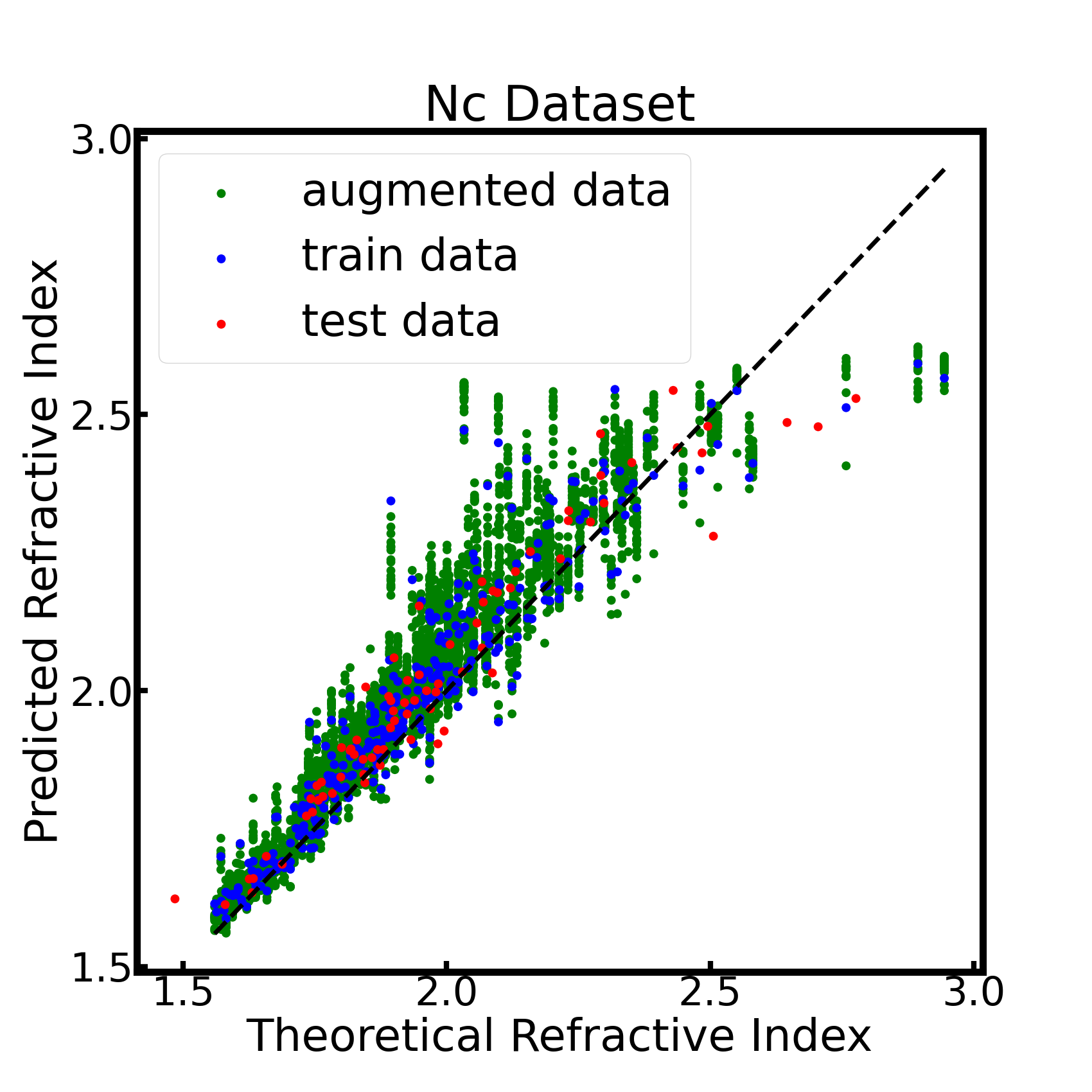}
        \put(-3,95){\textbf{i}}
    \end{overpic}
    
    }
    \makebox[\textwidth][c]{
    \begin{overpic}[scale=0.125]{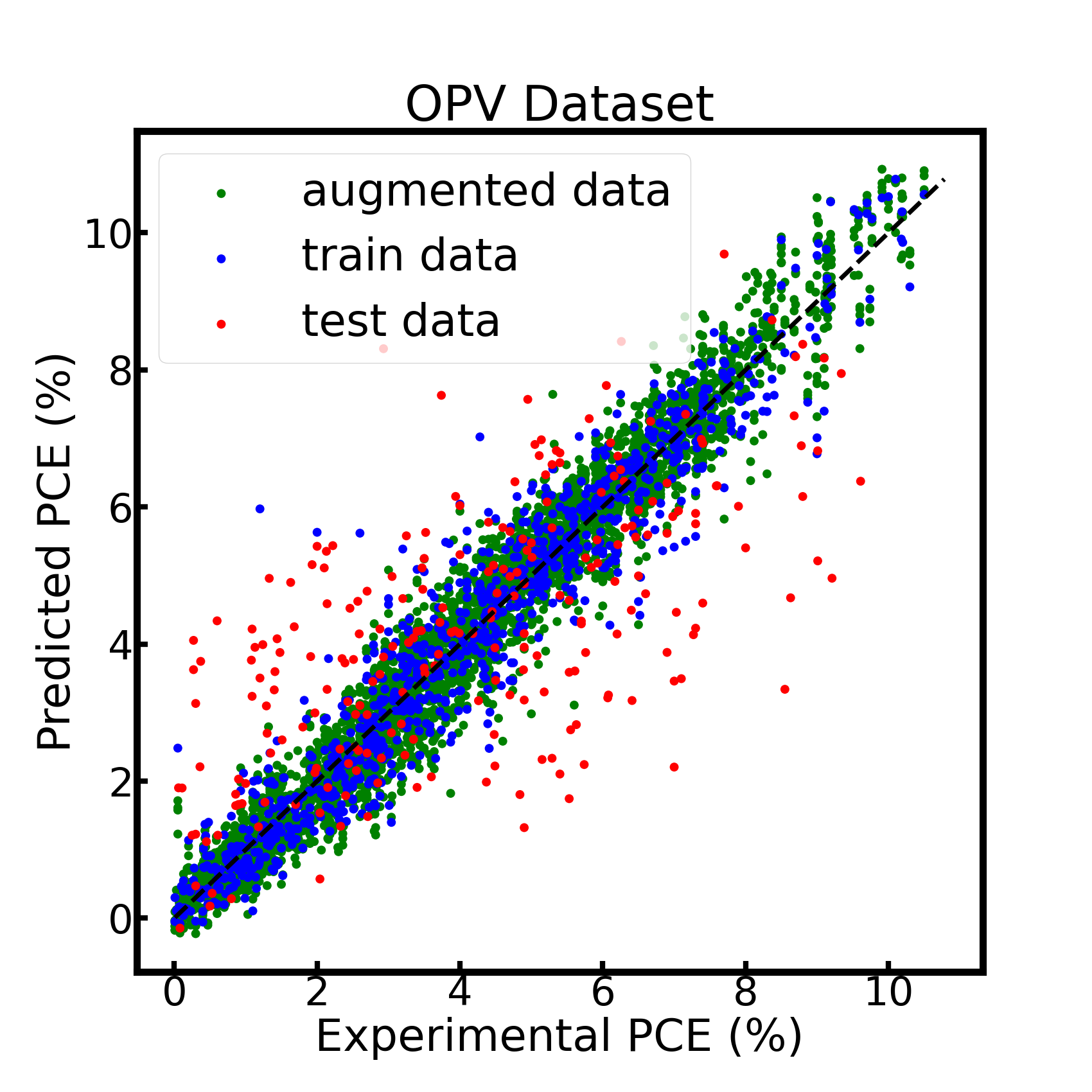}
        \put(-3,95){\textbf{j}}
    \end{overpic}
    }
    \caption{Scatter plots of ground truth vs. predicted values by TransPolymer\textsubscript{pretrained} for downstream datasets: (a) PE-\uppercase\expandafter{\romannumeral1}, (b) PE-\uppercase\expandafter{\romannumeral2}, (c) Egc, (d) Egb, (e) Eea, (f) Ei, (g) Xc, (h) EPS, (i) Nc, and (j) OPV. Augmented data points are also included in the plots. The dashed lines on diagonals stand for perfect regression.}
    \label{Predict_vs_true_aug}
    
\end{figure*}

\subsection*{Model Performance with Varying Pretraining Size}

Supplementary Table 5 summarizes the model performance on downstream tasks where the model is pretrained on data with different sizes. Standard deviation is included when cross-validation is applied in downstream tasks. As is shown in the table, model performance on downstream tasks increases with increasing pretraining data size.

\begin{table}[H]
    \caption{Effects of the size of pretraining dataset on model performance on downstream datasets.}
    \label{pretrain size}
    \centering
    \begin{adjustbox}{width=\textwidth}
        \begin{tabular}{cccccc|ccccc}
        \toprule
         \multirow{2}*{Dataset} & \multicolumn{5}{c}{Test RMSE} & \multicolumn{5}{c}{Test $R^2$} \\
         ~ & 5 K & 50 K & 500 K & 1 M & 5 M & 5 K & 50 K & 500 K & 1 M & 5 M \\ \midrule
         PE-I & 1.18 & 1.15 & 1.08 & 1.1  & 0.67 & 0.05 & 0.10 & 0.21 & 0.18 & 0.69 \\
         PE-II & 0.76 $\pm$ 0.09 & 0.67 $\pm$ 0.06 & 0.67 $\pm$ 0.07 & 0.66 $\pm$ 0.05 & 0.61 $\pm$ 0.07 & 0.57   $\pm$ 0.10  & 0.67 $\pm$ 0.06  & 0.68 $\pm$ 0.04  & 0.68 $\pm$ 0.02  & 0.73 $\pm$ 0.04 \\
         Egc & 0.57 $\pm$ 0.02 & 0.49 $\pm$ 0.02 & 0.46 $\pm$ 0.02 & 0.45 $\pm$ 0.01 & 0.44 $\pm$ 0.01 & 0.87 $\pm$ 0.01 & 0.90 $\pm$ 0.01 & 0.91 $\pm$ 0.01 & 0.92 $\pm$ 0.01 & 0.92 $\pm$ 0.00 \\
         Egb & 0.62 $\pm$ 0.05 & 0.54 $\pm$ 0.07 & 0.53 $\pm$ 0.05 & 0.54 $\pm$ 0.05 & 0.52 $\pm$ 0.05 & 0.90 $\pm$ 0.01 & 0.92 $\pm$ 0.01 & 0.93 $\pm$ 0.01 & 0.92 $\pm$ 0.01 & 0.93 $\pm$ 0.01 \\
         Eea & 0.35 $\pm$ 0.02 & 0.33 $\pm$ 0.04 & 0.31 $\pm$ 0.02 & 0.32 $\pm$ 0.04 & 0.32 $\pm$ 0.02 & 0.89 $\pm$ 0.01 & 0.90 $\pm$ 0.03 & 0.92 $\pm$ 0.02 & 0.90 $\pm$ 0.03 & 0.91 $\pm$ 0.03 \\
         Ei & 0.44 $\pm$ 0.06 & 0.43 $\pm$ 0.07 & 0.43 $\pm$ 0.06 & 0.41 $\pm$ 0.05 & 0.39 $\pm$ 0.07  & 0.79 $\pm$ 0.07 & 0.80 $\pm$ 0.07 & 0.80 $\pm$ 0.07 & 0.82 $\pm$ 0.06 & 0.84 $\pm$ 0.06 \\
         Xc & 19.89 $\pm$ 1.84 & 18.16 $\pm$ 1.09 & 17.43 $\pm$ 0.18 & 17.44 $\pm$ 0.96  & 16.57 $\pm$ 0.68 & 0.29 $\pm$ 0.10  &  0.40 $\pm$ 0.06 & 0.45 $\pm$ 0.07  & 0.45 $\pm$ 0.07  & 0.50 $\pm$ 0.06 \\
         EPS & 0.59 $\pm$ 0.07 & 0.55 $\pm$ 0.07 & 0.58 $\pm$ 0.07 & 0.58 $\pm$ 0.07 & 0.52 $\pm$ 0.07 & 0.71 $\pm$ 0.09  & 0.74 $\pm$ 0.08  & 0.71 $\pm$ 0.09  & 0.71 $\pm$ 0.10 & 0.76 $\pm$ 0.11 \\
          
         Nc & 0.11 $\pm$ 0.02 & 0.10 $\pm$ 0.02 & 0.10 $\pm$ 0.02 & 0.10 $\pm$ 0.02 & 0.10 $\pm$ 0.02  & 0.78 $\pm$ 0.07 & 0.81 $\pm$ 0.07 & 0.81 $\pm$ 0.08 & 0.81 $\pm$ 0.09 & 0.82 $\pm$ 0.07 \\
         OPV & 2.12 $\pm$ 0.07 & 2.07 $\pm$ 0.06 & 1.96 $\pm$ 0.04 & 1.94 $\pm$ 0.05 & 1.92 $\pm$ 0.06 & 0.18 $\pm$ 0.04 & 0.21 $\pm$ 0.04 & 0.30 $\pm$ 0.04 & 0.31 $\pm$ 0.04 & 0.32 $\pm$ 0.05 \\
        \bottomrule 
    		 
        \end{tabular}
    \end{adjustbox}
    
\end{table}

\subsection*{LSTM Gradient Diminishing}

From Table 2-5 in the manuscript, LSTM gives the worst performance on most of the downstream tasks. Plus, we even observe that the model performance is not improved if we reduce the model size from 6 layers, 768 hidden size to 2 layers, 256 hidden size, and add a convolution layer before the LSTM layers. 

In order to investigate the reason for the poor performance of LSTM on the property prediction tasks, we provide the visualization of gradient flow across LSTM layers in Supplementary Figure 3 to check whether the models suffer from the diminishing gradient problem. The models are trained on the Ei dataset for which five-fold cross-validation is applied. We plot the gradient of each layer for every back-propagation step and present them in a single figure for each LSTM architecture. Darker colors in the figures indicate that there are more bars at that place. The mean gradient suffers from the diminishing problem for all three models, while the maximum gradient does not diminish significantly for smaller models (two LSTM layers). Even though LSTM is better at remembering information from past hidden states in long sequences than regular RNNs, it is still confronted with the gradient diminishing problem in this situation. 

\begin{figure*}[htbp]
    \centering
    \makebox[\textwidth][c]{
        \begin{overpic}[scale=0.55]{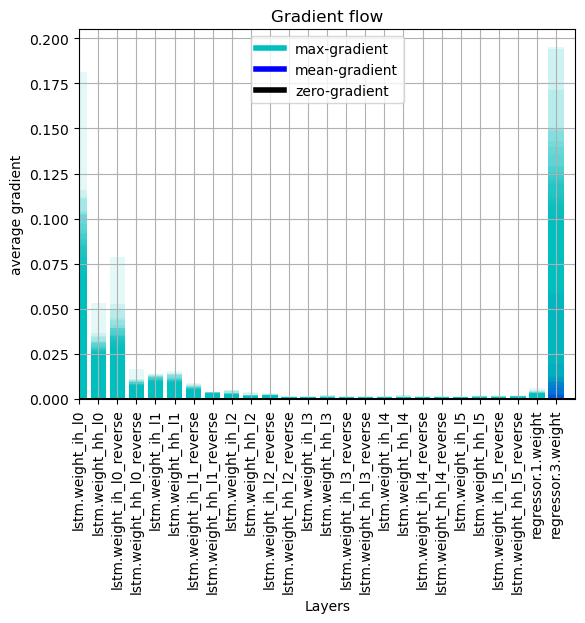}
        \put(1,96){\textbf{a}}
    \end{overpic}
    \begin{overpic}[scale=0.55]{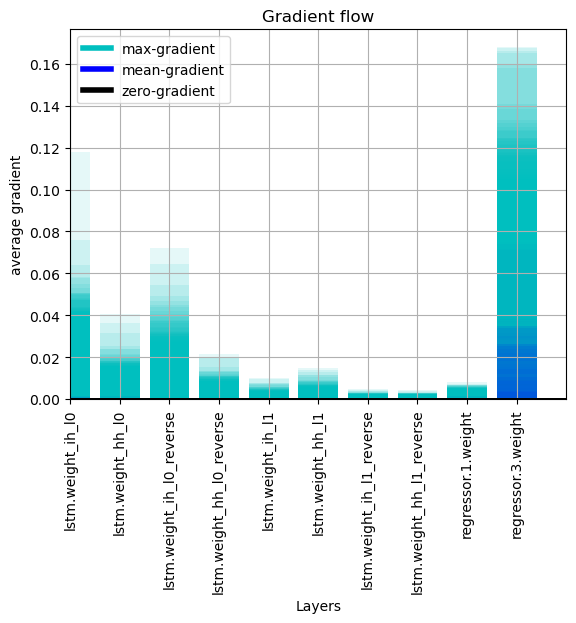}
        \put(1,96){\textbf{b}}
    \end{overpic}
    }
    \makebox[\textwidth][c]{
    \begin{overpic}[scale=0.55]{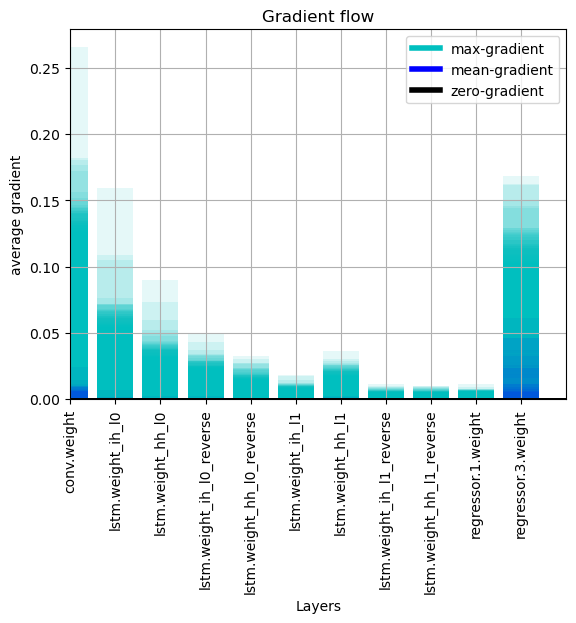}
        \put(1,96){\textbf{c}}
    \end{overpic}
    }
    
    \caption{Visualization of gradient flow across LSTM layers for different model architectures: (a) 6 LSTM layers, 768 hidden size, and no convolution layer; (b) 2 LSTM layers, 256 hidden size, and no convolution layer;(c) 2 LSTM layers, 256 hidden size, and convolution layer.}
    \label{gradient}
    
\end{figure*}

Besides, we include the train and test loss versus epoch plots in Supplementary Figure 4 for different model architectures. We can tell from the loss vs. epoch plots that in many folds the training is early stopped before reaching 20 epochs due to the non-decreasing test loss, which means it is very easy for LSTM to overfit even though dropout and time and frequency masking are used. 

\begin{figure*}[htbp]
    \centering
    
    \begin{overpic}[width=0.9\textwidth]{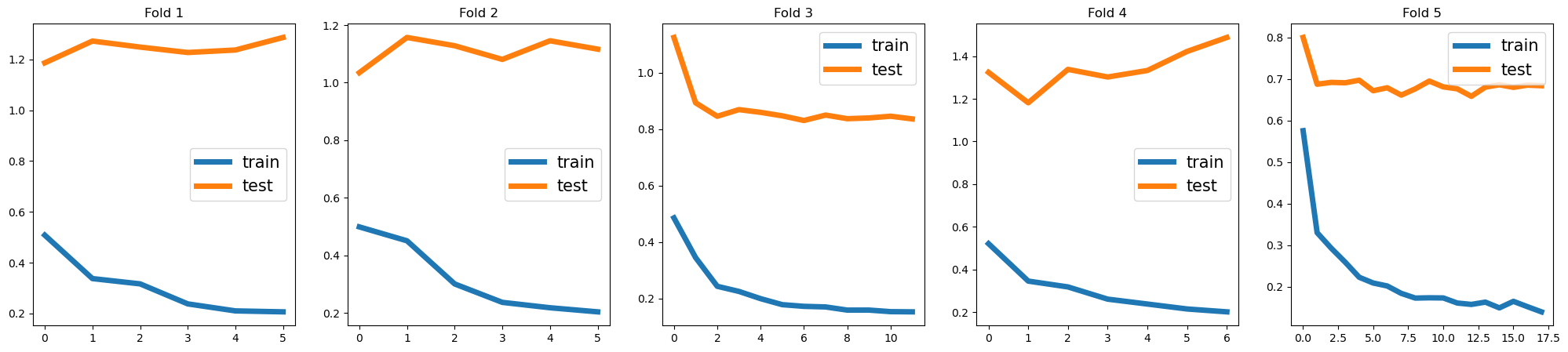}
        \put(-1, 22){\textbf{a}}
    \end{overpic}
    \vspace{10mm}

    \begin{overpic}[width=0.9\textwidth]{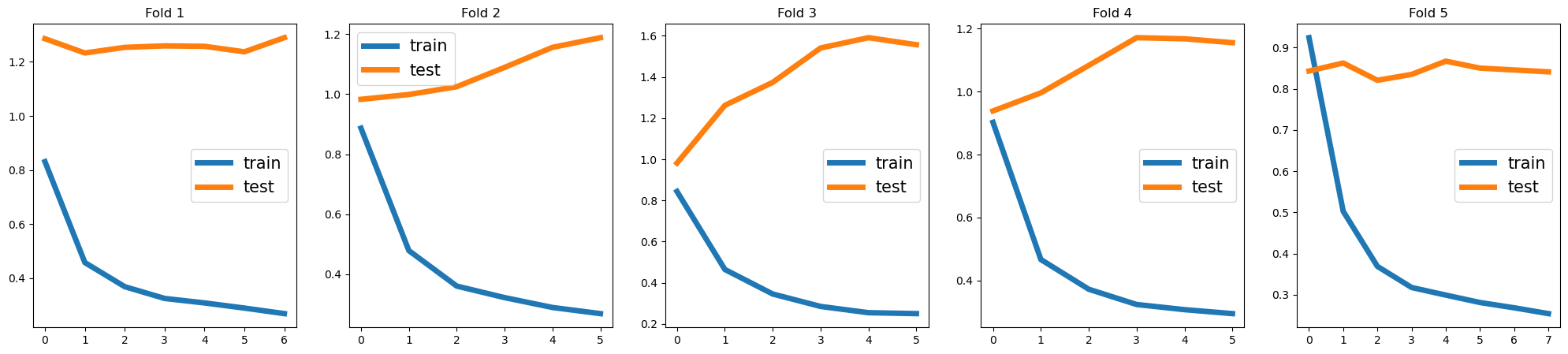}
        \put(-1,22){\textbf{b}}
    \end{overpic}
    \vspace{10mm}

    \begin{overpic}[width=0.9\textwidth]{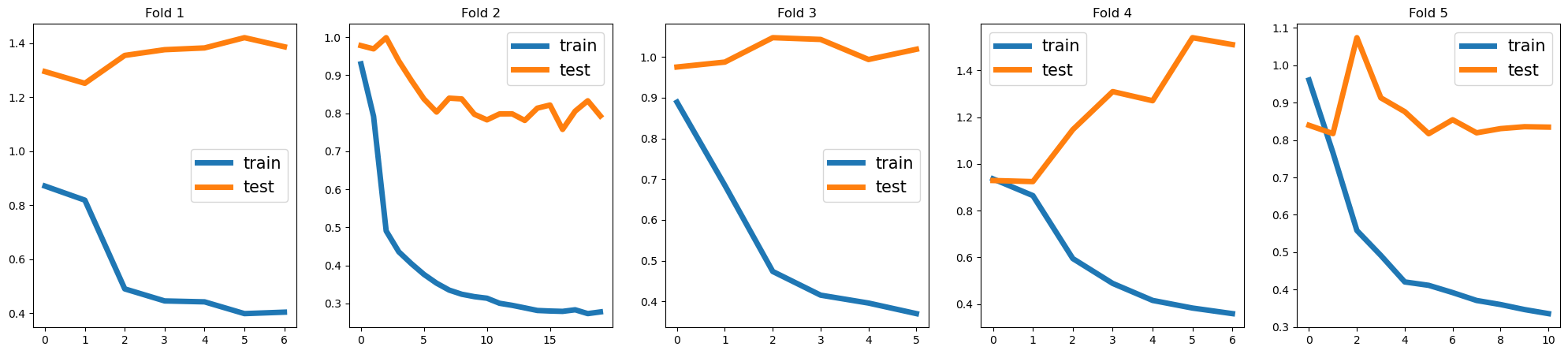}
        \put(-1,22){\textbf{c}}
    \end{overpic}
    
    \caption{Train and test loss versus epoch for different model architectures: (a) 6 LSTM layers, 768 hidden size, and no convolution layer; (b) 2 LSTM layers, 256 hidden size, and no convolution layer; (c) 2 LSTM layers, 256 hidden size, and convolution layer. Five-fold cross-validation is applied for each experiment and the model is trained for 20 epochs for each fold. The training process is early stopped if the test loss is not decreasing for successive 5 epochs.}
    \label{loss}
\end{figure*}

The experiment results strongly suggest that LSTM is not capable of representation learning from polymer sequences. In comparison, the good performance of Transformer on the same dataset highlights the advantage of the attention mechanism in understanding chemical knowledge from polymer sequences.

\clearpage

\bibliography{SI_ref}
